\documentclass{article}

\newcommand{\numbenchmarks}{63 }

 \PassOptionsToPackage{numbers, compress}{natbib}

\usepackage[draft]{changes}

\usepackage[preprint]{neurips_2026}

\usepackage[utf8]{inputenc} %
\usepackage[T1]{fontenc}    %
\usepackage{hyperref}       %
\usepackage{url}            %
\usepackage{booktabs}       %
\usepackage{amsfonts}       %
\usepackage{amsmath}
\usepackage{nicefrac}       %
\usepackage{microtype}      %
\usepackage{subfiles}
\usepackage{xcolor}
\usepackage{multirow}
\usepackage{enumerate}
\usepackage{subfiles}
\usepackage{multirow}
\usepackage{graphicx}
\usepackage{subfiles}
\usepackage{verbatim}
\usepackage{longtable}
\usepackage{float}          %
\usepackage{tabularx}       %
\usepackage[table]{xcolor}  %

\usepackage{etoolbox}       %
\providetoggle{backlog}
\settoggle{backlog}{true}

\usepackage{listings}       %
\usepackage{colortbl}

\newcommand{\unreliable}[1]{\cellcolor{red!15}{#1}}
\newcommand{\insufficient}[1]{\cellcolor{gray!15}{#1}}

\graphicspath{ {./Figures/img/} }

\usepackage{natbib}
\usepackage{pdflscape}  %

\usepackage{placeins}

\newif\ifanonymous

\definechangesauthor[name={Edward}, color=blue]{edward}
\ifanonymous
  \newcommand{\TODO}[1]{}
  \newcommand{\NOTE}[1]{}
\else
  \newcommand{\TODO}[1]{\highlight[id=edward]{TODO | #1}\par}
  \newcommand{\NOTE}[1]{\highlight[id=edward]{NOTE | #1}\par}
\fi

\title{PRIMETIME: Limits of LLMs in Temporal Primitives}

\author{
  Edward Gaere, Florian von Wangenheim\\
  Department of Management, Technology and Economics (D-MTEC)\\
  ETH Zurich\\
  \texttt{egaere@ethz.ch, fwangenheim@ethz.ch}
}

\begin{document}

\maketitle

\begin{abstract}

This paper introduces PRIMETIME, a synthetic generator that supports both benchmarking and fine-tuning of two primitive operations underlying temporal reasoning in Large Language Models (LLMs): parsing and arithmetic on datetimes. Existing temporal benchmarks assume simplified canonical datetime forms, conflate arithmetic, composition, and world knowledge into a single aggregate score, and offer no direct path to remediation. 

The first contribution is methodological: the PRIMETIME synthetic generator delivers non-conflated, uncontaminated, and unlimited datetime exemplars that enable a decompositional evaluation strategy for each primitive in isolation. The generator is extensible to support complex datetime tasks and is publicly released\footnote{\url{https://github.com/LLM-DATETIME/Generator}}, alongside generated benchmarks\footnote{\url{https://github.com/LLM-DATETIME/Datasets}}. The second contribution is diagnostic: under this evaluation strategy, the primitives themselves prove individually unreliable, with per-primitive accuracy ranging from near-zero to perfect across models and prompting conditions. The third contribution is constructive: the same generator used for diagnosis also produces new training exemplars for fine-tuning, and the resulting models show that the primitives are fully learnable and the composed \emph{Event Planning} task reaches frontier-level accuracy using small quantized LoRA transformers. The broader takeaway is that a single synthetic generator can serve both diagnosis and production-ready deployment. This methodological pattern may apply beyond temporal reasoning.

\end{abstract}

\section{Introduction}
  \label{introduction}
  
\subsection{The importance of datetimes}
The ability for machines to accurately process datetimes is important. Datetime reasoning is a core sub-skill in conversational settings, where users ask planning questions that require implicit calendar arithmetic, such as understanding the temporal sequence of events, determining whether a deadline can be met, or assisting with scheduling.

Beyond conversational settings, temporal processing capabilities can streamline and democratize data processing workflows that require datetime standardization, for example in data exchange or analytics. In an increasingly data-driven society, it is systemically beneficial to reduce the economic costs associated with data preparation tasks by increasing the accuracy of automated data processing. In the limit, automated data cleaning and preparation, which requires accurate datetime processing amongst other preconditions, could be pivotal for the automatic generation of new knowledge from unprocessed data sources. It is also important to ensure that such technologies are widely available, remain open source and can run on consumer-grade hardware.

\subsection{Datetimes and their challenges}
A datetime is a date and a time. In PRIMETIME, the same canonical datetime might appear as \texttt{May|6th,2026 11:59:59 PM} — one of the benchmark's messy representations, reflecting datetimes as they could appear in the wild rather than as prior benchmarks present them after normalization.

Datetimes are particularly challenging because they can be represented in millions of ways due to the ordering of the components, component presence or absence, the different representations, locales and time zones. The sheer variety of surface forms means that solutions to tasks cannot be simply memorized and a minimum amount of generic translation and/or reasoning capabilities is required. Furthermore, datetime operations such as adding days are non-trivial due to the variable number of days per month and leap years.

A first primitive task is the \emph{Translation} from a surface form to the ISO-8601 standard\footnote{\url{https://www.iso.org/iso-8601-date-and-time-format.html}} which would yield \emph{2026-05-06T23:59:59} for the above example. A second, deceptively simple task is \emph{Add-250}, the addition of a specific number of days, e.g., 250 days. Although datetime arithmetic is trivial in any programming language, the LLM's ability to perform it internally matters for composed tasks like event planning: handling the arithmetic step inside the model avoids the operational overhead and cost of routing to external tooling for what is, in isolation, a one-line computation.

A third task, \emph{Event Planning}, evaluates both skills jointly in a conversational setting, where users pose planning questions such as ``I need 250 days to prepare for an event on 11th.february.2023 ,1:12:31. If I start on 15th.july.2022 ,1:12:31, will I be ready in time?'' Such queries latently require datetime translation, date addition, and a comparison, composing multiple operations in a single query.

  \section{Contributions}
  \label{contributions}
  This paper presents three contributions to the study of temporal reasoning.

The first is methodological. Recent evaluations report that LLM datetime processing requires improvement~\citep{fatemi2025test, wei2025time, chu-etal-2024-timebench, wang-zhao-2024-tram}. Understanding where datetime reasoning breaks down requires a decompositional evaluation strategy. To this effect, PRIMETIME provides a synthetic data generator that produces non-conflated, uncontaminated, and unlimited datetime exemplars, tests parsing and arithmetic primitives in isolation, and supports composed tasks such as event planning. The generator can be reconfigured (e.g.\ enabling timezones or microseconds; see Appendix~\ref{generator}) or extended in code (e.g.\ adding new operation types) to produce arbitrarily many unseen exemplars and entirely new tasks. The released benchmarks therefore fall in the \emph{Live Ground Truth} category \citep{chiang2024chatbotarenaopenplatform}, with regeneration as a direct response to future saturation.

The second contribution is diagnostic. Under this strategy, proprietary and open-weight models are evaluated. Prior findings are confirmed: temporal reasoning is inconsistent across models, and only state-of-the-art proprietary frontier models perform reliably across all tasks. The new finding is that the primitives themselves, parsing and datetime arithmetic, are individually unreliable. 

The third contribution is constructive. The same generator used for diagnosis produces new training exemplars for fine-tuning. Small quantized LoRA fine-tuned Qwen~2.5 models \citep{unsloth, qwen2025qwen25technicalreport} achieve perfect out-of-sample accuracy on both primitives, and frontier-level accuracy on the composed \emph{Event Planning} task. These results suggest a practical pattern for temporal reasoning and potentially beyond: a single synthetic generator can serve both diagnosis and production-ready deployment, with small specialised models trained on its output acting as co-processors alongside general-purpose LLMs.

\section{Related Work}

\paragraph{Established Benchmarks.} A comprehensive review of \numbenchmarks frequently used LLM benchmarks in recent literature \citep{cobbe2021trainingverifierssolvemath, chen2021evaluatinglargelanguagemodels, hendrycks2021measuringmassivemultitasklanguage, hendrycks2021measuringmathematicalproblemsolving, clark2018thinksolvedquestionanswering, chollet2019measureintelligence, srivastava2023imitationgamequantifyingextrapolating, white2024livebenchchallengingcontaminationfreellm, wang2019glue, wang2020superglue, hu2020xtreme, clark2020tydi, sang2002introduction, adelani2021masakhaner, parrish2022bbqhandbuiltbiasbenchmark, suzgun2022challenging, mishra2022crosstaskgeneralizationnaturallanguage,
zellers2019hellaswagmachinereallyfinish, nie2020adversarialnlinewbenchmark, jin2020diseasedoespatienthave, zhong2023agievalhumancentricbenchmarkevaluating, joshi2017triviaqalargescaledistantly,
bisk2019piqareasoningphysicalcommonsense, sakaguchi2019winograndeadversarialwinogradschema, mihaylov2018suitarmorconductelectricity, clark2019boolqexploringsurprisingdifficulty, talmor2019commonsenseqaquestionansweringchallenge,
lin2022truthfulqameasuringmodelsmimic, austin2021programsynthesislargelanguage, rein2023gpqagraduatelevelgoogleproofqa, zheng2023judgingllmasajudgemtbenchchatbot, zhou2023instructionfollowingevaluationlargelanguage, sprague2024musrtestinglimitschainofthought, patel-etal-2021-nlp, miao2021diversecorpusevaluatingdeveloping, 
kwiatkowski-etal-2019-natural, lu2024mathvistaevaluatingmathematicalreasoning, lu2021intergpsinterpretablegeometryproblem, chen2024breakinglanguagebarriersmultilingual, rajpurkar2016squad100000questionsmachine, rajpurkar2018knowdontknowunanswerable, marcus-etal-1994-penn, reddy2019coqaconversationalquestionanswering, paperno2016lambadadatasetwordprediction,
mostafazadeh2016corpusevaluationframeworkdeeper, Levesque2012TheWS, choi2018quacquestionanswering, lai2017racelargescalereadingcomprehension, dua2019dropreadingcomprehensionbenchmark, warstadt2019neuralnetworkacceptabilityjudgments, williams2018broadcoveragechallengecorpussentence, Cer_2017, 
socher-etal-2013-recursive, dolan-brockett-2005-automatically, qqp, zhang2024naturallanguageembeddedprograms, 2023opencompass}
reveals no systematic evaluation of LLMs on datetime processing. A detailed literature review is provided in Appendix \ref{appendix_benchmark_review}, with examples of existing tasks relating to date and times provided in Appendix \ref{appendix_existing_datetime_tasks}. Specific domains and benchmarks were excluded from the review, such as coding and multimodal domains. 

Only ad hoc and fragmented datetime tasks currently exist, diluted across different benchmarks. They are notably absent from the most popular evaluation frameworks such as SuperGLUE \citep{wang2020superglue}, GPT-3 evaluation suite \citep{brown2020languagemodelsfewshotlearners}, Eleuther AI LM Harness \citep{eval-harness} and BIG-Bench \emph{Date Understanding} contains simple date-related challenges \cite{srivastava2023imitationgamequantifyingextrapolating, suzgun2022challenging}. The HELM framework \citep{liang2023holisticevaluationlanguagemodels}, that aims for a comprehensive assessment across a wide range of tasks (reasoning, knowledge, sentiment analysis) and metrics (accuracy, performance, safety). 

\paragraph{Recent Work In Temporal Reasoning.}
Recent benchmarks have evaluated temporal reasoning in LLMs~\citep{fatemi2025test, wei2025time, chu-etal-2024-timebench, wang-zhao-2024-tram} and identified open challenges that motivate further research. However, these benchmarks evaluate composed reasoning chains, conflate multiple factors, and assume a small set of canonical datetime forms, making it difficult to isolate the exact source of deficiencies. 

\emph{Test of Time}~\citep{fatemi2025test} includes the \emph{ToT-Arithmetic-add-subtract} subset, that evaluates date arithmetic in implicit narrative context with structured JSON output. A single accuracy score conflates four factors: semantic understanding, parsing, arithmetic, and output-format compliance. Operands often appear implicitly, as in ``In a movie, the tower took exactly 4 years to construct. They started construction in 12-15-2018. What year was the tower ready in?''\footnote{\url{https://huggingface.co/datasets/baharef/ToT/viewer/tot_arithmetic/test?p=2}}. PRIMETIME's \emph{Add-250}, by contrast, fixes the operand at 250 days, holds the surface form constant, and states the arithmetic operation explicitly — so only the arithmetic primitive varies across instances.

TIME \citep{wei2025time} contains 10 tasks. Its arithmetic-relevant \emph{Computation} task evaluates only implicit subtraction in complex real-world settings, e.g. ``what was the duration'' or ``how much time has passed.'' Surface forms are also normalised to canonical form during generation, eliminating surface-variation as a difficulty axis: manual inspection of approximately 20 instances per subtask confirms that datetimes appear in only five canonical English surface forms\footnote{The five observed forms are: \emph{April 30, 2017}; \emph{16 August, 2022}; \emph{2017-05-01 01:59:26}; \emph{9:43 am on 28 July, 2022}; \emph{11:16 am, April 08, 2022}. See \url{https://huggingface.co/datasets/SylvainWei/TIME}.}.

TimeBench~\citep{chu-etal-2024-timebench} organises temporal reasoning into three levels (symbolic, commonsense, event) across ten tasks. Its symbolic level includes a date-arithmetic task, \emph{Symbolic Arithmetic}, that differs from PRIMETIME's \emph{Add-250} on three axes: dates appear only in canonical English surface forms, the operation is latent (e.g., ``What is the time 7 year and 11 month before Sep, 1104?'' (sic), where ``before'' implies subtraction), and the operand varies across instances. Day-level addition is not present in the released dataset\footnote{\url{https://github.com/zchuz/TimeBench/blob/main/TimeBench-full-19000.zip}}.

TRAM~\citep{wang-zhao-2024-tram} is a ten-task benchmark of 526K multiple-choice questions. Its date-computation subtask varies operator (addition, subtraction), offset unit (days, weeks, months, years), surface form, and answer format simultaneously, e.g. ``What will be the time 41 year and 11 months after August 1717?'' (sic). Conflating these four axes, an incorrect answer cannot be attributed to any single skill\footnote{\url{https://github.com/EternityYW/TRAM-Benchmark/tree/main/datasets}}.

Across these benchmarks, the closest analogue to PRIMETIME's \emph{Add-250} — date arithmetic — is invariably entangled with parsing, narrative comprehension, or output formatting, and surface forms are restricted to a small set of canonical English variants. None of the four benchmarks evaluates \emph{Translation} from surface forms to ISO-8601

\section{Dataset \& Benchmark Construction}
  \label{construction}
  
This section describes the methodology, design decisions, and constraints for generating both the benchmark instances and the fine-tuning datasets. The fine-tuning datasets and benchmarks are synthetically generated, yielding deterministic input-to-target mappings, ground-truth labels free of annotation noise, frequent low-cost regeneration to reduce contamination, and uniform sampling across days, months, and years. Each decision below is motivated by a constrained objective: maximising the variety of surface forms while ensuring that (i) every input maps to a single unambiguous target, and (ii) smaller models are not penalised for limited multilingual pretraining. All of the following constraints can be trivially relaxed as future work, with instructions provided in Appendix~\ref{generator} or in the accompanying repository\footnote{\url{https://github.com/LLM-DATETIME/Generator}}. Details on benchmark and fine-tuning dataset generation are in Appendix~\ref{reproducibility}.

\paragraph{Component ordering and separators.} The ordering of the six date components year, month, day, hours, minutes and seconds as well as the set of separators (e.g.\ \texttt{' '}, \texttt{/}, \texttt{-}, \texttt{.}) are predetermined as templates, and the generator performs a random selection from these at generation time to render each surface form. The full set of templates, separators, and rendering logic is available in the repository. The two remaining components, microseconds and timezone are discussed hereafter.

\paragraph{Single locale.}
The locale is fixed to \texttt{en\_US} throughout this paper. Multi-locale is supported but the evaluation would disproportionately penalise smaller models that may have had limited or no exposure to multilingual data during pretraining or fine-tuning, conflating datetime reasoning ability with language coverage. 

\paragraph{Unambiguous month representation.}
All benchmark instances use an unambiguous month schema in which months are represented by their English names or standard abbreviations rather than by Arabic numerals alone.
This ensures that the model has a deterministic way to distinguish the month field from the day field. Ambiguous, all-numeric month representations (e.g.\ \texttt{03/04/2024}) are deliberately excluded from the current benchmark.

\paragraph{Four-digit year formats.}
Date schemas are restricted to four-digit years. The schemas used are \texttt{day-month-yyyy}, \texttt{day-month-weekday-yyyy}, \texttt{month-day-yyyy}, and \texttt{month-day-weekday-yyyy}.
Two-digit year formats introduce an additional layer of ambiguity, century inference, that is orthogonal to the parsing and arithmetic tasks evaluated here. 

\paragraph{Time components and formats.}
The time component of each representation must contain at least the hour in a non-ambiguous form, either as a 24-hour value (e.g.\ \texttt{13}) or as a 12-hour value with a meridiem indicator (e.g.\ \texttt{1 PM}). Minutes and seconds may be included but are not required, as the tasks evaluated in this paper do not depend on sub-hour granularity. Timezone and microseconds are omitted for the same reason. These restrictions can be relaxed in future work to support finer-grained parsing and the timezone arithmetic evaluated in \emph{Test of Time}.

\paragraph{Date range.}
Input dates span the years 1000 to 9999. The lower bound follows directly from the four-digit year constraint described above; the upper bound is the maximum representable four-digit year. For the \emph{Event Planning} task, the date range is restricted to the near future (2027--2036). Conversational APIs typically inject the current date into the system prompt. Dates far from this anchor would create a confound between temporal reasoning and the model's handling of contextually implausible inputs.

\paragraph{Target labels.}
For tasks that produce a datetime as output (\emph{Translation}, \emph{Add-250}), the target labels follow ISO~8601, a universally adopted datetime standard whose fixed structure allows trivial evaluation via exact-match comparison or a simple regular expression. Microseconds and timezone offsets are omitted (e.g.\ \texttt{7942-01-22T23:41:06}), as the core parsing and arithmetic tasks are already challenging for smaller models; adding timezone resolution or sub-second precision would increase complexity without additional diagnostic value at this stage.

\paragraph{Month-boundary crossing.}
For all tasks explicitly or latently requiring day-addition, the generator enforces that input and output dates fall in different months (\texttt{same\_month\,=\,0}). This excludes instances solvable by within-month addition (trivial, no calendar knowledge required) and ensures every arithmetic instance requires crossing at least one month boundary — accounting for varying month lengths, year boundaries, and, in some cases, leap-year logic.

\paragraph{Input sequence finalisation.}
Each rendered datetime appearing in an input sequence undergoes a final cleaning step before inclusion in the dataset. Unicode characters are normalised to ASCII via NFKD decomposition to ensure that all input characters map to well-represented tokens across common tokeniser vocabularies. Leading and trailing whitespace are stripped. A final post-generation verification discards any input in which a required component is missing (e.g.\ \texttt{01/\!/ 2010 1PM}), as such inputs are impossible to solve. This occurs silently when \texttt{Babel} \footnote{Babel 2.9.1 \url{https://babel.pocoo.org/en/latest/dates.html}} cannot resolve certain format variations.

\paragraph{Quality review.} Several hundred samples were manually inspected to verify input well-formedness and target correctness.

\paragraph{Implementation.}
The generator is implemented in Python and leverages the \texttt{Babel} library to render datetime objects into arbitrary natural-language representations. Additionally, the \texttt{num2words} \footnote{num2words 0.5.10 \url{https://pypi.org/project/num2words/}} and \texttt{roman} \footnote{roman 3.3 \url{https://pypi.org/project/roman/}} libraries are used to render the numerals in alternative forms, for example \emph{twenty-five} or \emph{IV}, respectively. Quick start and detailed instructions are provided in Appendices~\ref{generator} and~\ref{generator_details}.

\section{Generated Data}
  \label{descriptives}

Each component of a generated datetime admits multiple surface forms: for example, the month may appear as \texttt{March} or \texttt{Mar}, and the day as \texttt{31}, \texttt{31st}, or \texttt{thirty-one}. When requesting $1{,}000{,}000$ representations from a single datetime under the constraints specified in Section~\ref{construction}, the generator produced zero duplicate strings, illustrating the combinatorial diversity due to the interaction of component ordering, separators, optional components, and component-level formatting.

\begin{table}[ht!]
\centering
\caption{Example natural representations of the datetime \texttt{2026-03-31T19:34:23} (Tuesday), illustrating variation in component ordering, separators, granularity, and numeral rendering according to the construction methodology.}
\label{tab:datetime-examples}
\begin{tabular}{l}
\toprule
\textbf{Natural Representation of \texttt{2026-03-31T19:34:23}} \\
\midrule
\texttt{19:34, Tuesday 31\#March\#2026} \\
\texttt{2026/31/Mar, 19:34} \\
\texttt{7 PM 34:23 Tue March|thirty-one|2026} \\
\texttt{Mar.31.2026, 19:34:23} \\
\texttt{Tue 31st\#March-2026 19} \\
\texttt{7 PM 34, 31, March\#2026} \\
\bottomrule
\end{tabular}
\end{table}

The format complexity is justified on two grounds: it generalises the variety of datetime representations that may be encountered in real-world corpora and unprocessed datasets, and as shown in Section~\ref{results}, frontier models as well as small quantized fine-tuned models can parse these formats with near-perfect accuracy, confirming that the representations are within reach of current models. Details on benchmark and fine-tuning dataset generation are in Appendix~\ref{reproducibility}.

\section{Experiments and Results}
  \label{tasks}
  Open-weight and proprietary LLMs are evaluated on three tasks: \emph{Translation}, \emph{Add-250}, and \emph{Event Planning}. Three experimental design choices accommodate the possibility that LLMs were not specifically designed or trained for temporal reasoning, in particular for ISO~8601 generation. First, generation budget is sufficient for scratchpad reasoning~\citep{nye2021workscratchpadsintermediatecomputation, wei2023chainofthoughtpromptingelicitsreasoning}: \texttt{max\_new\_tokens} is set to 3000. Second, lenient regexes — tested for false positives and negatives via unit and property-based tests — extract the answer from each response (Appendix~\ref{regexes}). Third, each task is evaluated under a variety of prompt styles (direct, few-shot with uncontaminated examples, chain-of-thought) to measure prompt sensitivity, on the premise that performance should be prompt-invariant. Some variants (the ``ISO'' suffix in the results tables) include an ISO~8601 definition for models that may not encode this knowledge. Full reproducibility details are in Appendix~\ref{reproducibility}.

  \label{results}

  \subsection{Translation Experiment and Results}
  \label{results_translation}
  \paragraph{Task description.} \emph{Translation} evaluates the parsing primitive: the model must convert a natural-language datetime representation to ISO~8601 format. To establish a baseline, six datetime parsing libraries were evaluated. The best, Python \texttt{dateparser}, achieved 36\%$\pm$3.0 accuracy (see Appendix~\ref{baseline}).

\paragraph{Rendered prompt example (Direct-1).} "Here is a datetime: "Feb/Sun 26th,7111 ,1 AM 45". Translate the datetime to ISO-8601 format." $\to$ "7111-02-26T01:45:00"

\begin{table}[ht!]
\tiny
\caption{Accuracy in percent of proprietary LLMs on the PRIMETIME Translation task, across different models (rows) and different prompt styles (columns). Best result in bold, worst underlined. All ISO 8601 sequences in the output sequence are evaluated; a prediction is correct if any match the gold label. Sorted by descending maximum accuracy across all prompt styles. Minimum across prompt styles is also reported to assess prompt sensitivity. Evaluated on 100 observations from a single run only due to API costs. See Appendix~\ref{datetime_translation_prop_repro} for full reproducibility details.}
\label{datetime_translation_prop}
\begin{tabular}{lrrrrrrrrrr}
\toprule
 & Direct-1 & Direct-2 & Direct-1 ISO & FS-1 & FS-2 & CoT & CoT FS-1 & CoT FS-2 & Min & Max \\
\midrule
OpenAI gpt-5 & $\mathrm{92.0}$ & $\mathrm{89.0}$ & $\mathrm{94.0}$ & $\mathbf{98.0}$ & $\mathrm{95.0}$ & $\mathrm{94.0}$ & $\mathrm{97.0}$ & $\mathbf{98.0}$ & $\mathrm{89.0}$ & $\mathbf{98.0}$ \\
OpenAI gpt-5-mini & $\mathrm{89.0}$ & $\mathrm{91.0}$ & $\mathrm{92.0}$ & $\mathrm{94.0}$ & $\mathrm{89.0}$ & $\mathrm{93.0}$ & $\mathrm{93.0}$ & $\mathrm{92.0}$ & $\mathrm{89.0}$ & $\mathrm{94.0}$ \\
Anthropic claude-4.1-opus & $\mathrm{82.0}$ & $\mathrm{70.0}$ & $\mathrm{85.0}$ & $\mathrm{81.0}$ & $\mathrm{86.0}$ & $\mathrm{88.0}$ & $\mathrm{92.0}$ & $\mathrm{92.0}$ & $\mathrm{70.0}$ & $\mathrm{92.0}$ \\
Gemini gemini-2.5-pro & $\mathrm{82.0}$ & $\mathrm{86.0}$ & $\mathrm{88.0}$ & $\mathrm{85.0}$ & $\mathrm{90.0}$ & $\mathrm{86.0}$ & $\mathrm{88.0}$ & $\mathrm{89.0}$ & $\mathrm{82.0}$ & $\mathrm{90.0}$ \\
Anthropic claude-4.0-sonnet & $\mathrm{83.0}$ & $\mathrm{81.0}$ & $\mathrm{83.0}$ & $\mathrm{85.0}$ & $\mathrm{86.0}$ & $\mathrm{88.0}$ & $\mathrm{89.0}$ & $\mathrm{90.0}$ & $\mathrm{81.0}$ & $\mathrm{90.0}$ \\
Anthropic claude-4.0-opus & $\mathrm{82.0}$ & $\mathrm{77.0}$ & $\mathrm{85.0}$ & $\mathrm{82.0}$ & $\mathrm{85.0}$ & $\mathrm{83.0}$ & $\mathrm{90.0}$ & $\mathrm{86.0}$ & $\mathrm{77.0}$ & $\mathrm{90.0}$ \\
OpenAI gpt-5-nano & $\mathrm{81.0}$ & $\mathrm{78.0}$ & $\mathrm{83.0}$ & $\mathrm{84.0}$ & $\mathrm{83.0}$ & $\mathrm{88.0}$ & $\mathrm{88.0}$ & $\mathrm{87.0}$ & $\mathrm{78.0}$ & $\mathrm{88.0}$ \\
OpenAI gpt-4.0-turbo & $\mathrm{80.0}$ & $\mathrm{82.0}$ & $\mathrm{85.0}$ & $\mathrm{84.0}$ & $\mathrm{86.0}$ & $\mathrm{83.0}$ & $\mathrm{84.0}$ & $\mathrm{84.0}$ & $\mathrm{80.0}$ & $\mathrm{86.0}$ \\
Gemini gemini-2.5-flash & $\mathrm{82.0}$ & $\mathrm{76.0}$ & $\mathrm{84.0}$ & $\mathrm{84.0}$ & $\mathrm{79.0}$ & $\mathrm{77.0}$ & $\mathrm{81.0}$ & $\mathrm{81.0}$ & $\mathrm{76.0}$ & $\mathrm{84.0}$ \\
OpenAI gpt-3.5-turbo & $\mathrm{79.0}$ & $\mathrm{75.0}$ & $\mathrm{74.0}$ & $\mathrm{81.0}$ & $\mathrm{81.0}$ & $\mathrm{76.0}$ & $\mathrm{79.0}$ & $\mathrm{80.0}$ & $\mathrm{74.0}$ & $\mathrm{81.0}$ \\
Gemini gemini-2.5-flash-lite & $\mathrm{65.0}$ & $\mathrm{65.0}$ & $\mathrm{68.0}$ & \underline{$\mathrm{62.0}$} & $\mathrm{76.0}$ & $\mathrm{75.0}$ & $\mathrm{74.0}$ & $\mathrm{81.0}$ & \underline{$\mathrm{62.0}$} & $\mathrm{81.0}$ \\
OpenAI gpt-4o-mini & $\mathrm{63.0}$ & \underline{$\mathrm{62.0}$} & $\mathrm{71.0}$ & $\mathrm{72.0}$ & $\mathrm{75.0}$ & $\mathrm{75.0}$ & $\mathrm{68.0}$ & $\mathrm{74.0}$ & \underline{$\mathrm{62.0}$} & $\mathrm{75.0}$ \\
\bottomrule
\end{tabular}
\end{table}

\begin{table}[ht!]
            \centering
            \caption{Accuracy of small open-weight language models on the PRIMETIME Translation task. We perform 10 runs per (model, prompt) pair. Each run draws a fresh set of 100 datetimes shared across all pairs to enable direct comparison, giving 1{,}000 datetimes per pair in total. Reported values are mean $\pm$ 2 SEM over the 10 runs. Run 1 corresponds to Table~\ref{datetime_translation_prop}. Sorted by descending maximum accuracy. The \emph{Min} and \emph{Max} columns report the mean (across runs) of each run's lowest and highest accuracy across prompt styles respectively, with $\pm$2 SEM. This averaging is preferred to single-cell extrema, which would over-weight tail observations from the noisiest runs. See Appendix~\ref{datetime_translation_vllm_repro} for full reproducibility details.}
            \label{datetime_translation_vllm}
            
            \resizebox{\textwidth}{!}{%
\begin{tabular}{lrrrrrrrrrr}
\toprule
 & Direct-1 & Direct-2 & Direct-1 ISO & FS-1 & FS-2 & CoT & CoT FS-1 & CoT FS-2 & Min & Max \\
\midrule
Qwen3 4B Greedy & $\mathrm{71.6}_{{\pm 3.1}}$ & $\mathrm{74.9}_{{\pm 3.1}}$ & $\mathrm{74.7}_{{\pm 3.1}}$ & $\mathrm{70.0}_{{\pm 3.7}}$ & $\mathrm{70.9}_{{\pm 4.5}}$ & $\mathrm{76.2}_{{\pm 3.5}}$ & $\mathrm{69.6}_{{\pm 3.6}}$ & $\mathrm{69.5}_{{\pm 3.2}}$ & $\mathrm{66.1}_{{\pm 3.5}}$ & $\mathbf{77.0}_{{\pm 2.9}}$ \\
Qwen3 1.7B Greedy & $\mathrm{58.3}_{{\pm 4.1}}$ & $\mathrm{57.1}_{{\pm 4.8}}$ & $\mathrm{60.4}_{{\pm 5.1}}$ & $\mathrm{57.7}_{{\pm 3.6}}$ & $\mathrm{50.7}_{{\pm 2.6}}$ & $\mathrm{63.7}_{{\pm 3.7}}$ & $\mathrm{59.4}_{{\pm 2.8}}$ & $\mathrm{55.0}_{{\pm 2.7}}$ & $\mathrm{49.4}_{{\pm 2.3}}$ & $\mathrm{64.3}_{{\pm 3.3}}$ \\
Llama 3.2 3B Greedy & $\mathrm{3.8}_{{\pm 1.3}}$ & $\mathrm{4.1}_{{\pm 1.2}}$ & $\mathrm{17.2}_{{\pm 1.8}}$ & $\mathrm{20.4}_{{\pm 3.2}}$ & $\mathrm{36.3}_{{\pm 3.3}}$ & $\mathrm{50.9}_{{\pm 3.9}}$ & $\mathrm{32.1}_{{\pm 9.2}}$ & $\mathrm{54.1}_{{\pm 4.7}}$ & $\mathrm{3.3}_{{\pm 0.8}}$ & $\mathrm{54.4}_{{\pm 3.5}}$ \\
Phi 4 Mini Greedy & $\mathrm{1.1}_{{\pm 0.6}}$ & $\mathrm{0.5}_{{\pm 0.3}}$ & $\mathrm{6.7}_{{\pm 1.2}}$ & $\mathrm{7.2}_{{\pm 1.6}}$ & $\mathrm{20.9}_{{\pm 2.5}}$ & $\mathrm{18.3}_{{\pm 2.3}}$ & $\mathrm{27.3}_{{\pm 3.2}}$ & $\mathrm{44.4}_{{\pm 3.9}}$ & $\mathrm{0.4}_{{\pm 0.3}}$ & $\mathrm{43.6}_{{\pm 3.8}}$ \\
Gemma 3 4B Greedy & $\mathrm{8.1}_{{\pm 1.4}}$ & $\mathrm{6.0}_{{\pm 1.8}}$ & $\mathrm{25.7}_{{\pm 3.9}}$ & $\mathrm{27.5}_{{\pm 3.9}}$ & $\mathrm{33.7}_{{\pm 3.3}}$ & $\mathrm{24.2}_{{\pm 3.0}}$ & $\mathrm{30.1}_{{\pm 3.7}}$ & $\mathrm{40.1}_{{\pm 2.8}}$ & $\mathrm{6.0}_{{\pm 1.6}}$ & $\mathrm{40.1}_{{\pm 3.0}}$ \\
\bottomrule
\end{tabular}}
            
            \end{table}

\begin{table}[ht!]
            \centering
            \caption{Accuracy of small quantized LoRA fine-tuned Qwen 2.5 models on the PRIMETIME Translation task. The columns are the number of exemplars for training, single epoch only. Each model is fine-tuned once and then run 10 times, exactly as in Table \ref{datetime_translation_vllm}. Sorted by descending accuracy after 50K training  observations and then by parameter count ascending. See Appendix~\ref{datetime_translation_unsloth_repro} for full reproducibility details.}
            \label{datetime_translation_unsloth}
            
            \resizebox{\textwidth}{!}{%
\begin{tabular}{lrrrrrrrrrr}
\toprule
 & Params (B) & No FT & 250 & 500 & 1K & 2.5K & 5K & 10K & 20K & 50K \\
\midrule
unsloth/Qwen2.5-0.5B & $\mathrm{0.5}$ & $\mathrm{0.0}_{{\pm 0.0}}$ & $\mathrm{3.3}_{{\pm 1.0}}$ & $\mathrm{11.0}_{{\pm 2.0}}$ & $\mathrm{46.9}_{{\pm 2.7}}$ & $\mathrm{91.0}_{{\pm 1.3}}$ & $\mathrm{99.0}_{{\pm 0.5}}$ & $\mathrm{99.8}_{{\pm 0.3}}$ & $\mathrm{99.9}_{{\pm 0.2}}$ & $\mathrm{100.0}_{{\pm 0.0}}$ \\
unsloth/Qwen2.5-3B & $\mathrm{3}$ & $\mathrm{0.6}_{{\pm 0.6}}$ & $\mathrm{10.4}_{{\pm 1.4}}$ & $\mathrm{56.6}_{{\pm 4.0}}$ & $\mathrm{75.9}_{{\pm 3.4}}$ & $\mathrm{98.2}_{{\pm 1.0}}$ & $\mathrm{99.7}_{{\pm 0.3}}$ & $\mathrm{99.7}_{{\pm 0.3}}$ & $\mathrm{99.9}_{{\pm 0.2}}$ & $\mathrm{100.0}_{{\pm 0.0}}$ \\
unsloth/Qwen2.5-7B Instruct & $\mathrm{7}$ & $\mathrm{0.7}_{{\pm 0.4}}$ & $\mathrm{1.5}_{{\pm 0.9}}$ & $\mathrm{21.6}_{{\pm 2.8}}$ & $\mathrm{71.4}_{{\pm 3.6}}$ & $\mathrm{92.7}_{{\pm 2.1}}$ & $\mathrm{98.7}_{{\pm 0.7}}$ & $\mathrm{99.6}_{{\pm 0.3}}$ & $\mathrm{99.5}_{{\pm 0.4}}$ & $\mathrm{100.0}_{{\pm 0.0}}$ \\
unsloth/Qwen2.5-14B Instruct & $\mathrm{14}$ & $\mathrm{0.3}_{{\pm 0.3}}$ & $\mathrm{0.0}_{{\pm 0.0}}$ & $\mathrm{33.3}_{{\pm 3.6}}$ & $\mathrm{79.3}_{{\pm 2.8}}$ & $\mathrm{95.5}_{{\pm 1.8}}$ & $\mathrm{98.9}_{{\pm 0.6}}$ & $\mathrm{99.9}_{{\pm 0.2}}$ & $\mathrm{100.0}_{{\pm 0.0}}$ & $\mathrm{100.0}_{{\pm 0.0}}$ \\
unsloth/Qwen2.5-3B Instruct & $\mathrm{3}$ & $\mathrm{0.6}_{{\pm 0.5}}$ & $\mathrm{0.4}_{{\pm 0.4}}$ & $\mathrm{11.6}_{{\pm 2.0}}$ & $\mathrm{61.6}_{{\pm 3.6}}$ & $\mathrm{88.5}_{{\pm 1.9}}$ & $\mathrm{99.0}_{{\pm 0.8}}$ & $\mathrm{99.6}_{{\pm 0.6}}$ & $\mathrm{100.0}_{{\pm 0.0}}$ & $\mathrm{99.9}_{{\pm 0.2}}$ \\
unsloth/Qwen2.5-1.5B & $\mathrm{1.5}$ & $\mathrm{0.0}_{{\pm 0.0}}$ & $\mathrm{0.7}_{{\pm 0.5}}$ & $\mathrm{10.9}_{{\pm 2.7}}$ & $\mathrm{69.4}_{{\pm 2.4}}$ & $\mathrm{93.7}_{{\pm 2.0}}$ & $\mathrm{95.5}_{{\pm 1.5}}$ & $\mathrm{94.4}_{{\pm 1.4}}$ & $\mathrm{91.7}_{{\pm 1.7}}$ & $\mathrm{95.1}_{{\pm 1.0}}$ \\
\bottomrule
\end{tabular}}
            
            \end{table}

  \subsection{Add-250 Experiment and Results}
  \label{results_computation}
  
\paragraph{Task description.} \emph{Add-250} evaluates date addition, one of several arithmetic primitives that operate on datetimes. The model must add 250 calendar days to an ISO~8601 datetime and produce a new ISO~8601 datetime. The operand is fixed to 250 to isolate arithmetic competence from operand variability. Fine-tuned models are additionally evaluated over learned operand ranges (1--10, 1--100, etc.) to assess generalisation across operands (Tables~\ref{datetime_add_scaling_law_1} and~\ref{datetime_add_scaling_law_2}). Other arithmetic primitives (subtraction, date difference, comparison) are reserved for future work. No specialised-library baseline is reported for this task as the operation is trivially computed in any programming language, but the LLM's ability to perform it internally matters for composed downstream tasks such as \emph{Event Planning}, which latently require date addition or subtraction. As with \emph{Translation}, any well-formed ISO~8601 representation of the correct target datetime is accepted (see Appendix~\ref{regexes}). All results in Appendix~\ref{key_computation_tables}.

\paragraph{Rendered prompt example (Direct-1).} "Here is a datetime in ISO-8601 format: "2024-03-15T14:30:00". Add 250 days to the datetime and generate a new datetime in ISO-8601 format." $\to$ "2024-11-20T14:30:00"

\begin{table}[ht!]
\tiny
\caption{Accuracy in percent of proprietary LLMs on the PRIMETIME Add-250 task, across different models (rows) and different prompt styles (columns). Best result in bold, worst underlined. Evaluated on 100 observations from a single run due to API costs. See Appendix~\ref{datetime_computation_add_250_prop_repro} for full reproducibility details.}
\label{datetime_computation_add_250_prop}
\begin{tabular}{lrrrrrrrrrrr}
\toprule
 & \shortstack{Direct\\1} & \shortstack{Direct\\2} & \shortstack{Direct-1\\ISO} & \shortstack{Direct-2\\ISO} & FS-1 & FS-2 & CoT-1 & \shortstack{CoT-1\\+ FS-1} & \shortstack{CoT-1\\+ FS-2} & Min & Max \\
\midrule
Gemini 2.5-pro & $\mathrm{100.0}$ & $\mathrm{96.0}$ & $\mathrm{98.0}$ & $\mathrm{100.0}$ & $\mathrm{100.0}$ & $\mathrm{98.0}$ & $\mathrm{81.0}$ & $\mathrm{86.0}$ & $\mathrm{71.0}$ & $\mathrm{71.0}$ & $\mathrm{100.0}$ \\
Gemini 3-flash-preview & $\mathrm{100.0}$ & $\mathrm{98.0}$ & $\mathrm{95.0}$ & $\mathrm{95.0}$ & $\mathrm{51.0}$ & $\mathrm{24.0}$ & $\mathrm{80.0}$ & $\mathrm{64.0}$ & $\mathrm{39.0}$ & $\mathrm{24.0}$ & $\mathrm{100.0}$ \\
OpenAI gpt-5 & $\mathrm{95.0}$ & $\mathrm{94.0}$ & $\mathrm{97.0}$ & $\mathrm{96.0}$ & $\mathrm{88.0}$ & $\mathrm{91.0}$ & $\mathrm{94.0}$ & $\mathrm{94.0}$ & $\mathrm{95.0}$ & $\mathrm{88.0}$ & $\mathrm{97.0}$ \\
OpenAI gpt-5-mini & $\mathrm{95.0}$ & $\mathrm{91.0}$ & $\mathrm{92.0}$ & $\mathrm{91.0}$ & $\mathrm{94.0}$ & $\mathrm{92.0}$ & $\mathrm{93.0}$ & $\mathrm{93.0}$ & $\mathrm{91.0}$ & $\mathrm{91.0}$ & $\mathrm{95.0}$ \\
Anthropic claude-4.0-sonnet & $\mathrm{68.0}$ & $\mathrm{71.0}$ & $\mathrm{63.0}$ & $\mathrm{72.0}$ & $\mathrm{66.0}$ & $\mathrm{57.0}$ & $\mathrm{64.0}$ & $\mathrm{84.0}$ & $\mathrm{95.0}$ & $\mathrm{57.0}$ & $\mathrm{95.0}$ \\
Anthropic claude-4.0-opus & $\mathrm{65.0}$ & $\mathrm{68.0}$ & $\mathrm{73.0}$ & $\mathrm{74.0}$ & $\mathrm{67.0}$ & $\mathrm{75.0}$ & $\mathrm{72.0}$ & $\mathrm{75.0}$ & $\mathrm{90.0}$ & $\mathrm{65.0}$ & $\mathrm{90.0}$ \\
Gemini 2.5-flash & $\mathrm{51.0}$ & $\mathrm{85.0}$ & $\mathrm{40.0}$ & $\mathrm{89.0}$ & $\mathrm{47.0}$ & $\mathrm{40.0}$ & $\mathrm{23.0}$ & $\mathrm{28.0}$ & $\mathrm{19.0}$ & $\mathrm{19.0}$ & $\mathrm{89.0}$ \\
\bottomrule
\end{tabular}
\end{table}

\begin{table}[ht!]
            \centering
            \caption{Accuracy of small quantized LoRA fine-tuned Qwen 2.5 models on the PRIMETIME Add-250 task. Sorted by descending maximum accuracy and then by parameter count ascending. See Appendix~\ref{datetime_computation_add_250_unsloth_repro} for full reproducibility details.}
            \label{datetime_computation_add_250_unsloth}
            
            \resizebox{\textwidth}{!}{%
\begin{tabular}{lrrrrrrrrrrrr}
\toprule
 & Params (B) & No FT & 250 & 500 & 1K & 2.5K & 5K & 10K & 20K & 50K & Min & Max \\
\midrule
unsloth/Qwen2.5-0.5B & $\mathrm{0.5}$ & $\mathrm{0.0}_{{\pm 0.0}}$ & $\mathrm{0.0}_{{\pm 0.0}}$ & $\mathrm{1.0}_{{\pm 0.6}}$ & $\mathrm{7.4}_{{\pm 1.9}}$ & $\mathrm{44.3}_{{\pm 4.5}}$ & $\mathrm{75.0}_{{\pm 2.7}}$ & $\mathrm{90.6}_{{\pm 1.5}}$ & $\mathrm{99.8}_{{\pm 0.3}}$ & $\mathrm{100.0}_{{\pm 0.0}}$ & $\mathrm{0.0}_{{\pm 0.0}}$ & $\mathrm{100.0}_{{\pm 0.0}}$ \\
unsloth/Qwen2.5-3B & $\mathrm{3}$ & $\mathrm{0.0}_{{\pm 0.0}}$ & $\mathrm{0.0}_{{\pm 0.0}}$ & $\mathrm{0.4}_{{\pm 0.3}}$ & $\mathrm{25.1}_{{\pm 3.9}}$ & $\mathrm{45.7}_{{\pm 3.5}}$ & $\mathrm{81.7}_{{\pm 2.5}}$ & $\mathrm{97.4}_{{\pm 1.0}}$ & $\mathrm{99.9}_{{\pm 0.2}}$ & $\mathrm{100.0}_{{\pm 0.0}}$ & $\mathrm{0.0}_{{\pm 0.0}}$ & $\mathrm{100.0}_{{\pm 0.0}}$ \\
unsloth/Qwen2.5-14B Instruct & $\mathrm{14}$ & $\mathrm{0.0}_{{\pm 0.0}}$ & $\mathrm{0.0}_{{\pm 0.0}}$ & $\mathrm{5.9}_{{\pm 2.0}}$ & $\mathrm{4.6}_{{\pm 1.1}}$ & $\mathrm{51.6}_{{\pm 2.5}}$ & $\mathrm{70.3}_{{\pm 3.0}}$ & $\mathrm{81.1}_{{\pm 2.3}}$ & $\mathrm{99.9}_{{\pm 0.2}}$ & $\mathrm{N/A}$ & $\mathrm{0.0}_{{\pm 0.0}}$ & $\mathrm{99.9}_{{\pm 0.2}}$ \\
unsloth/Qwen2.5-3B Instruct & $\mathrm{3}$ & $\mathrm{0.0}_{{\pm 0.0}}$ & $\mathrm{0.0}_{{\pm 0.0}}$ & $\mathrm{0.0}_{{\pm 0.0}}$ & $\mathrm{0.6}_{{\pm 0.4}}$ & $\mathrm{29.6}_{{\pm 3.6}}$ & $\mathrm{61.6}_{{\pm 2.8}}$ & $\mathrm{79.8}_{{\pm 1.9}}$ & $\mathrm{99.0}_{{\pm 0.6}}$ & $\mathrm{99.8}_{{\pm 0.3}}$ & $\mathrm{0.0}_{{\pm 0.0}}$ & $\mathrm{99.8}_{{\pm 0.3}}$ \\
unsloth/Qwen2.5-7B Instruct & $\mathrm{7}$ & $\mathrm{0.0}_{{\pm 0.0}}$ & $\mathrm{0.0}_{{\pm 0.0}}$ & $\mathrm{0.3}_{{\pm 0.3}}$ & $\mathrm{3.9}_{{\pm 1.3}}$ & $\mathrm{43.0}_{{\pm 3.0}}$ & $\mathrm{60.9}_{{\pm 3.3}}$ & $\mathrm{81.9}_{{\pm 2.1}}$ & $\mathrm{98.7}_{{\pm 0.8}}$ & $\mathrm{N/A}$ & $\mathrm{0.0}_{{\pm 0.0}}$ & $\mathrm{92.8}_{{\pm 11.8}}$ \\
\bottomrule
    \end{tabular}}
            
            \end{table}

\begin{table}[ht!]
\tiny
\caption{Accuracy of a LoRA fine-tuned quantized Qwen 2.5 0.5B model on PRIMETIME Add tasks with increasing range of days to be added. Each row corresponds to a different range of days added; each column to a different training set size. Single epoch, single run (N=100). Cells are shaded from red (0\%) to green (100\%).}
\label{datetime_add_scaling_law_1}
\begin{tabular}{lrrrrrrrrrr}
\toprule
 & 1K & 2.5K & 5K & 10K & 20K & 50K & 100K & 250K & 500K & 1M \\
Addition range &  &  &  &  &  &  &  &  &  &  \\
\midrule
Add 1-10 days & \cellcolor{red!18!green!11}$\mathrm{39.0}$ & \cellcolor{red!7!green!22}$\mathrm{73.4}$ & \cellcolor{red!4!green!25}$\mathrm{84.8}$ & \cellcolor{red!2!green!27}$\mathrm{90.5}$ & \cellcolor{red!0!green!29}$\mathrm{98.4}$ & \cellcolor{red!0!green!29}$\mathrm{98.7}$ & \cellcolor{red!0!green!29}$\mathrm{99.6}$ & \cellcolor{red!0!green!29}$\mathrm{99.9}$ & \cellcolor{red!0!green!30}$\mathrm{100.0}$ & \cellcolor{red!0!green!30}$\mathrm{100.0}$ \\
Add 1-100 days & \cellcolor{red!29!green!0}$\mathrm{2.0}$ & \cellcolor{red!27!green!2}$\mathrm{6.9}$ & \cellcolor{red!25!green!4}$\mathrm{15.1}$ & \cellcolor{red!19!green!10}$\mathrm{36.2}$ & \cellcolor{red!12!green!18}$\mathrm{60.0}$ & \cellcolor{red!3!green!26}$\mathrm{87.9}$ & \cellcolor{red!2!green!27}$\mathrm{92.4}$ & \cellcolor{red!0!green!29}$\mathrm{98.0}$ & \cellcolor{red!0!green!29}$\mathrm{99.5}$ & \cellcolor{red!0!green!29}$\mathrm{99.9}$ \\
Add 1-250 days & \cellcolor{red!29!green!0}$\mathrm{0.5}$ & \cellcolor{red!29!green!0}$\mathrm{1.8}$ & \cellcolor{red!28!green!1}$\mathrm{4.4}$ & \cellcolor{red!27!green!2}$\mathrm{7.4}$ & \cellcolor{red!23!green!6}$\mathrm{21.4}$ & \cellcolor{red!8!green!21}$\mathrm{70.2}$ & \cellcolor{red!5!green!24}$\mathrm{81.3}$ & \cellcolor{red!2!green!27}$\mathrm{92.2}$ & \cellcolor{red!1!green!28}$\mathrm{95.4}$ & \cellcolor{red!0!green!29}$\mathrm{99.9}$ \\
Add 1-1000 days & \cellcolor{red!30!green!0}$\mathrm{0.0}$ & \cellcolor{red!29!green!0}$\mathrm{0.3}$ & \cellcolor{red!29!green!0}$\mathrm{0.4}$ & \cellcolor{red!29!green!0}$\mathrm{0.7}$ & \cellcolor{red!29!green!0}$\mathrm{1.4}$ & \cellcolor{red!28!green!1}$\mathrm{6.4}$ & \cellcolor{red!21!green!8}$\mathrm{28.2}$ & \cellcolor{red!12!green!17}$\mathrm{58.4}$ & \cellcolor{red!4!green!25}$\mathrm{84.9}$ & \cellcolor{red!1!green!28}$\mathrm{93.5}$ \\
\bottomrule
\end{tabular}
\end{table}

\begin{table}[ht!]
\tiny
\caption{Accuracy of a LoRA fine-tuned quantized Qwen 2.5 0.5B model on PRIMETIME Add tasks with increasing range width, offset by 1{,}001 days. All ranges start at 1{,}001 to isolate the effect of range width from offset magnitude, matching the widths of Table~\ref{datetime_add_scaling_law_1}. Each row corresponds to a different range; each column to a different training set size. Single epoch, single run (N=100).}
\label{datetime_add_scaling_law_2}
\begin{tabular}{lrrrrrrrrrr}
\toprule
 & 1K & 2.5K & 5K & 10K & 20K & 50K & 100K & 250K & 500K & 1M \\
Addition range &  &  &  &  &  &  &  &  &  &  \\
\midrule
Add 1001-1010 days & \cellcolor{red!30!green!0}$\mathrm{0.0}$ & \cellcolor{red!29!green!0}$\mathrm{2.4}$ & \cellcolor{red!26!green!3}$\mathrm{10.3}$ & \cellcolor{red!26!green!3}$\mathrm{11.4}$ & \cellcolor{red!13!green!16}$\mathrm{54.2}$ & \cellcolor{red!3!green!26}$\mathrm{89.4}$ & \cellcolor{red!0!green!29}$\mathrm{99.1}$ & \cellcolor{red!0!green!29}$\mathrm{99.9}$ & \cellcolor{red!0!green!30}$\mathrm{100.0}$ & \cellcolor{red!0!green!30}$\mathrm{100.0}$ \\
Add 1001-1100 days & \cellcolor{red!30!green!0}$\mathrm{0.0}$ & \cellcolor{red!29!green!0}$\mathrm{0.9}$ & \cellcolor{red!29!green!0}$\mathrm{1.9}$ & \cellcolor{red!29!green!0}$\mathrm{1.8}$ & \cellcolor{red!27!green!2}$\mathrm{8.3}$ & \cellcolor{red!24!green!5}$\mathrm{19.4}$ & \cellcolor{red!13!green!16}$\mathrm{56.0}$ & \cellcolor{red!2!green!27}$\mathrm{90.0}$ & \cellcolor{red!0!green!29}$\mathrm{99.7}$ & \cellcolor{red!0!green!30}$\mathrm{100.0}$ \\
Add 1001-1250 days & \cellcolor{red!30!green!0}$\mathrm{0.0}$ & \cellcolor{red!30!green!0}$\mathrm{0.0}$ & \cellcolor{red!29!green!0}$\mathrm{0.3}$ & \cellcolor{red!29!green!0}$\mathrm{0.6}$ & \cellcolor{red!28!green!1}$\mathrm{3.8}$ & \cellcolor{red!27!green!2}$\mathrm{8.7}$ & \cellcolor{red!24!green!5}$\mathrm{18.2}$ & \cellcolor{red!2!green!27}$\mathrm{90.0}$ & \cellcolor{red!0!green!29}$\mathrm{97.3}$ & \cellcolor{red!0!green!29}$\mathrm{99.7}$ \\
Add 1001-2000 days & \cellcolor{red!30!green!0}$\mathrm{0.0}$ & \cellcolor{red!29!green!0}$\mathrm{0.1}$ & \cellcolor{red!29!green!0}$\mathrm{0.4}$ & \cellcolor{red!29!green!0}$\mathrm{0.5}$ & \cellcolor{red!29!green!0}$\mathrm{0.2}$ & \cellcolor{red!29!green!0}$\mathrm{2.1}$ & \cellcolor{red!29!green!0}$\mathrm{2.9}$ & \cellcolor{red!28!green!1}$\mathrm{5.1}$ & \cellcolor{red!26!green!3}$\mathrm{12.8}$ & \cellcolor{red!24!green!6}$\mathrm{20.0}$ \\
\bottomrule
\end{tabular}
\end{table}

  \subsection{Event Planning Experiment and Results}
  \label{results_nc}
  
\paragraph{Task description.} \emph{Event Planning} evaluates the \emph{Translation} and \emph{Add-250} primitives under a unified task in a conversational setting. A user prompts a chat LLM to determine whether a task started on one datetime can be completed before a deadline, with both datetimes shown in surface form rather than pre-normalised. The answer is binary: \texttt{yes} or \texttt{no}. Performance is measured by accuracy, unadjusted for chance: To allow for thinking, the last occurency of yes or no is selected as the final answer. See Appendix~\ref{regexes} for regex details, respectively. All results in Appendix~\ref{key_nc_tables}.

\paragraph{Rendered prompt example (Direct-1).} "I need 250 days to prepare for an event. I can work any day of the week. The event is on 11:13:10 AM,2028-27th/Aug. If I start preparing on 11:13:10 AM,2028-2nd/Jan, will I be ready in time? Answer with yes or no." $\to$ "no"

\begin{table}[ht!]
\tiny
\caption{Accuracy of proprietary LLMs on PRIMETIME Event Planning Task. Sorted by descending maximum accuracy and then by descending minimum accuracy to resolve ties. Best result in bold, worst underlined. The random baseline is 50\% due to the binary outcome. Cells below 75\% accuracy (1 prediction in 4 wrong) are shaded. See Appendix~\ref{datetime_natural_context_prop_repro} for full reproducibility details.}
\label{datetime_natural_context_prop_llm}
\begin{tabular}{lrrrrrrrrrr}
\toprule
 & \shortstack{Direct\\1} & \shortstack{Direct\\2} & \shortstack{Direct\\3} & FS-1 & FS-2 & CoT-1 & \shortstack{CoT-1\\+ FS-1} & \shortstack{CoT-1\\+ FS-2} & Min & Max \\
\midrule
Gemini 3-flash-preview & $\mathbf{100.0}$ & $\mathbf{100.0}$ & $\mathbf{100.0}$ & $\mathbf{100.0}$ & $\mathbf{100.0}$ & $\mathbf{100.0}$ & $\mathrm{99.0}$ & $\mathbf{100.0}$ & $\mathrm{99.0}$ & $\mathrm{100.0}$ \\
Gemini 2.5-pro & $\mathrm{98.0}$ & $\mathrm{99.0}$ & $\mathrm{98.0}$ & $\mathbf{100.0}$ & $\mathbf{100.0}$ & $\mathrm{98.0}$ & $\mathrm{98.0}$ & $\mathrm{99.0}$ & $\mathrm{98.0}$ & $\mathrm{100.0}$ \\
OpenAI gpt-5-mini & $\mathrm{98.0}$ & $\mathrm{99.0}$ & $\mathrm{98.0}$ & $\mathrm{99.0}$ & $\mathrm{100.0}$ & $\mathrm{97.0}$ & $\mathrm{100.0}$ & $\mathrm{99.0}$ & $\mathrm{97.0}$ & $\mathrm{100.0}$ \\
Anthropic claude-4.5-sonnet & $\mathrm{99.0}$ & $\mathrm{99.0}$ & $\mathrm{97.0}$ & $\mathbf{100.0}$ & $\mathbf{100.0}$ & $\mathbf{100.0}$ & $\mathrm{99.0}$ & $\mathrm{99.0}$ & $\mathrm{97.0}$ & $\mathrm{100.0}$ \\
Gemini 2.5-flash & $\mathrm{97.0}$ & $\mathrm{98.0}$ & $\mathrm{98.0}$ & $\mathrm{97.0}$ & $\mathbf{100.0}$ & $\mathbf{100.0}$ & $\mathbf{100.0}$ & $\mathrm{98.0}$ & $\mathrm{97.0}$ & $\mathrm{100.0}$ \\
Anthropic claude-4.5-opus & \cellcolor{red!15}$\mathrm{63.0}$ & \cellcolor{red!15}$\mathrm{58.0}$ & \cellcolor{red!15}$\mathrm{54.0}$ & \cellcolor{red!15}$\mathrm{57.0}$ & \cellcolor{red!15}$\mathrm{68.0}$ & $\mathbf{100.0}$ & $\mathbf{100.0}$ & $\mathbf{100.0}$ & \cellcolor{red!15}$\mathrm{54.0}$ & $\mathrm{100.0}$ \\
Anthropic claude-4.1-opus & \cellcolor{red!15}$\mathrm{57.0}$ & \cellcolor{red!15}$\mathrm{50.0}$ & \cellcolor{red!15}$\mathrm{58.0}$ & \cellcolor{red!15}$\mathrm{53.0}$ & \cellcolor{red!15}$\mathrm{62.0}$ & $\mathrm{99.0}$ & $\mathrm{99.0}$ & $\mathrm{98.0}$ & \cellcolor{red!15}$\mathrm{50.0}$ & $\mathrm{99.0}$ \\
\bottomrule
\end{tabular}
\end{table}

\begin{table}[ht!]
            \centering
            \caption{Accuracy of small quantized LoRA fine-tuned Qwen 2.5 LLMs on the PRIMETIME Event Planning task. Sorted by descending accuracy after 1M training observations and then by parameter count ascending. See Appendix~\ref{datetime_natural_context_unsloth_repro} for full reproducibility details.}
            \label{datetime_event_planning_250_unsloth}
            
            \resizebox{\textwidth}{!}{%
\begin{tabular}{lrrrrrrrrrr}
\toprule
 & Params (B) & No FT & 10K & 20K & 50K & 100K & 250K & 500K & 1M & Max \\
\midrule
unsloth/Qwen2.5-1.5B Instruct & $\mathrm{1.5}$ & $\mathrm{46.90}_{{\pm 3.05}}$ & $\mathrm{49.50}_{{\pm 1.41}}$ & $\mathrm{48.30}_{{\pm 3.13}}$ & $\mathrm{50.80}_{{\pm 2.65}}$ & $\mathrm{98.70}_{{\pm 0.60}}$ & $\mathrm{99.30}_{{\pm 0.52}}$ & $\mathrm{99.30}_{{\pm 0.52}}$ & $\mathrm{99.30}_{{\pm 0.52}}$ & $\mathrm{99.30}_{{\pm 0.52}}$ \\
unsloth/Qwen2.5-3B Instruct & $\mathrm{3}$ & $\mathrm{43.40}_{{\pm 1.69}}$ & $\mathrm{49.10}_{{\pm 2.37}}$ & $\mathrm{51.70}_{{\pm 2.17}}$ & $\mathrm{52.80}_{{\pm 3.58}}$ & $\mathrm{99.00}_{{\pm 0.60}}$ & $\mathrm{99.30}_{{\pm 0.52}}$ & $\mathrm{99.30}_{{\pm 0.52}}$ & $\mathrm{99.30}_{{\pm 0.52}}$ & $\mathrm{99.30}_{{\pm 0.52}}$ \\
unsloth/Qwen2.5-14B Instruct & $\mathrm{14}$ & $\mathrm{33.10}_{{\pm 2.94}}$ & $\mathrm{49.30}_{{\pm 1.91}}$ & $\mathrm{50.80}_{{\pm 1.02}}$ & $\mathrm{97.90}_{{\pm 0.70}}$ & $\mathrm{99.20}_{{\pm 0.50}}$ & $\mathrm{99.30}_{{\pm 0.52}}$ & $\mathrm{99.30}_{{\pm 0.52}}$ & $\mathrm{99.30}_{{\pm 0.52}}$ & $\mathrm{99.30}_{{\pm 0.52}}$ \\
unsloth/Qwen2.5-7B Instruct & $\mathrm{7}$ & $\mathrm{41.30}_{{\pm 1.93}}$ & $\mathrm{50.20}_{{\pm 1.07}}$ & $\mathrm{50.10}_{{\pm 1.25}}$ & $\mathrm{98.40}_{{\pm 0.68}}$ & $\mathrm{99.30}_{{\pm 0.52}}$ & $\mathrm{99.30}_{{\pm 0.52}}$ & $\mathrm{99.30}_{{\pm 0.52}}$ & $\mathrm{99.30}_{{\pm 0.52}}$ & $\mathrm{99.30}_{{\pm 0.52}}$ \\
unsloth/Qwen2.5-3B & $\mathrm{3}$ & $\mathrm{50.00}_{{\pm 0.73}}$ & $\mathrm{50.00}_{{\pm 0.00}}$ & $\mathrm{49.70}_{{\pm 0.31}}$ & $\mathrm{39.60}_{{\pm 1.94}}$ & $\mathrm{50.00}_{{\pm 0.00}}$ & $\mathrm{63.70}_{{\pm 4.06}}$ & $\mathrm{60.90}_{{\pm 3.33}}$ & $\mathrm{49.10}_{{\pm 3.80}}$ & $\mathrm{65.30}_{{\pm 2.53}}$ \\
\bottomrule
\end{tabular}}
            
            \end{table}

\section{Discussion}

  \label{discussion}
  These results confirm prior reports that LLM datetime reasoning is unreliable across models, prompting conditions, and task variants ~\citep{fatemi2025test, wei2025time, chu-etal-2024-timebench, wang-zhao-2024-tram}. Neither a developer building a processing pipeline nor an end user interacting with a chat assistant can predict in advance whether a given temporal query will return a reliable answer without task-specific evaluation.

\paragraph{Primitives are learnable, and so is Event Planning.} The two primitives are fully learnable through fine-tuning, reaching perfect held-out accuracy at every evaluated model size using a training set that covers 1.6\% of the date space\footnote{50K training examples drawn from the 3.2M (year, month, day) combinations across years 1000--9999.} — pointing to learnability rather than memorisation. A scaling-law analysis on the addition primitive shows that learning extends to operand ranges, not just fixed operands: a small model learns to add any value within a bounded range, and shifting that range to higher values does not affect performance. The model therefore acquires the addition operation itself rather than memorising input-output pairs. Composition, in the specific form tested here, is also learnable: a small quantized Instruct model, LoRA fine-tuned on generator output, matches the strongest frontier models on Event Planning.

\paragraph{Failure analysis.} The failure analysis (Appendix~\ref{failure_analysis}) clarifies the nature of the failures in three respects. First, both open-weight and proprietary models reliably produce well-formed ISO-8601 outputs on Translation and Add-250, ruling out format unfamiliarity as a confound. Second, when models fail on Add-250 they fail by a wide margin — typically more than 100 days from the target — indicating categorical breakdowns of the arithmetic procedure rather than near-miss errors. Third, visible scratchpad use varies systematically by model: among proprietary models, OpenAI shows none in the response body across all prompt styles, Anthropic shows one only when CoT is requested, and Gemini shows one even under Direct prompting; among open-weight models, the Qwen family shows one regardless of prompt style, which is consistent with their stronger open-weight performance.

\paragraph{Temporal LLM co-processors.} These results motivate a practical architecture: a lightweight temporal LLM co-processor to which a general-purpose LLM delegates datetime operations. Unlike Toolformer-style delegation to deterministic external functions \citep{schick2023toolformerlanguagemodelsteach}, cost-driven routing across general-purpose models of varying size \citep{dekoninck2025unifiedapproachroutingcascading}, or mixture-of-experts routing within a single model \citep{shazeer2017outrageouslylargeneuralnetworks}, the temporal co-processor is itself an LLM, smaller in scale, with bounded scope and predictable per-task accuracy. The breadth of that scope is itself a question: a scaling-law analysis on date addition reveals a clear trade-off — as the operand range widens (e.g., from 1--10 days to 1--100 days), accuracy degrades and training-data requirements grow sharply (Tables~\ref{datetime_add_scaling_law_1} and~\ref{datetime_add_scaling_law_2}). Full characterisation across operations is reserved for future work.

\section{Limitations}
  \label{limitations}
  Seven limitations apply to the present work within the constraints stated in Section~\ref{construction}. First, training and evaluation share a single generator. Train and test splits are disjoint but drawn from the same closed domain of surface form variations, so the fine-tuning results may not generalise to arbitrary ones. Second, the fine-tuning results cover only one arithmetic operation (date addition) with a fixed operand of 250 days. A scaling-law analysis (Appendix~\ref{scaling_law}) shows that wider operand ranges can be learned but need much more training data. Whether this extends to other operations (subtraction, comparison, etc.) and to fully arbitrary ranges is left for future work. Third, \emph{Event Planning} probes only one compositional structure under a single natural framing. Whether the observed chance-level performance and the improvement through fine-tuning generalise to other subtasks in this space and to alternative natural framings remains an open question. Fourth, the prompt templates are author-designed, see Appendix~\ref{reproducibility}; different formulations could shift model performance. The fine-tuned models in particular are trained and evaluated under a single prompt template per task. Fifth, all fine-tuning experiments use a single model family, Qwen 2.5; the learnability claims may not generalise to other architectures or fine-tuning techniques. Sixth, the comparison primitive that is latently required in the event-planning composition is not evaluated in isolation, although doing so would be straightforward with the extensible generator design. Seventh, ISO~8601 tasks are scored leniently which inflates accuracy relative to production settings requiring a single unambiguous output.

\section{Conclusion}
  \label{conclusion}
  PRIMETIME provides a systematic framework for evaluating and improving LLM temporal reasoning. The benchmark reveals that current models handle datetimes inconsistently, at the level of individual primitives, not just composed tasks, with accuracy ranging from near-zero to perfect across models and prompting conditions. Fine-tuning demonstrates that these primitives are fully learnable by small quantized models that reach perfect accuracy on parsing and arithmetic, and frontier-level accuracy on a composed task. These results support a practical architecture in which a general-purpose LLM delegates datetime operations to a smaller scale LLM with bounded scope and predictable per-task accuracy.

 \clearpage
 \ifanonymous\else
\paragraph{Acknowledgements} I am grateful to Prof. Dr. Florian von Wangenheim for the opportunity to pursue my research; to Sabine Keller, Michaela Diehl and Petra Monsch for their continued support; and to Edoardo Talotti and Philipp Sauter from ETH Zurich IT Services for setting up and maintaining the compute resources.
\fi

\clearpage
\bibliography{../bibliography}

\clearpage
\section{Appendices}
\clearpage
\appendix
\section{Declaration of originality and use of AI tools}
  \label{originality}
  
I hereby declare that I authored the work in question independently. In doing so I only used the authorised aids, which included suggestions from the supervisor regarding language and content and generative artificial intelligence technologies. The use of the latter and the respective source
declarations proceeded in consultation with the supervisor.

The AI tools \emph{Anthropic Claude}, \emph{OpenAI ChatGPT} and \emph{Microsoft Copilot} and were used to improve wording, to correct spelling and grammar, translate Python code and JSON configuration settings to English for describing the methodology, and to write specific code blocks such as regexes, property-tests, and table rendering in LaTeX.

\paragraph{Formulation, grammar and spelling improvement.}
\scriptsize
\begin{verbatim}
OpenAI ChatGPT: "What word can I use to describe the holistic evaluation of an LLM?"

Microsoft Copilot: "rewrite in formal style: comprised of six basic tasks"

OpenAI ChatGPT: "Is it grammatically correct to say: 'evaluate the performance on simple reasoning tasks around datetimes.'"

OpenAI ChatGPT: "Write a sentence in academic style: "easy for humans, difficult for LLMs"

Anthropic Claude: "check spelling: "The closed sourced models were evaluated using each vendor's API,"

\end{verbatim}
\normalfont
\normalsize

\paragraph{Code to English.}
\scriptsize
\begin{verbatim}
Here are my regexes for capturing and parsing ISO-8601


self.regex_iso8601_1_pattern = r'\d{4}-\d{1,2}-\d{1,2}(?:[T ]\d{1,2}(?::\d{1,2})*[^\s]*?)?'
self.regex_iso8601_1 = compile(self.regex_iso8601_1_pattern)

# NOTE: all components are optional
self.regex_iso8601_1_components_pattern = r"(?P<year>\d{4})?(?:-(?P<month>\d{1,2}))?(?:-(?P<day>\d{1,2}))?(?:[T ](?P<hour>\d{1,2}))?(?::(?P<minute>\d{1,2}))?(?::(?P<second>\d{1,2}))?(?P<suffix>.*)"

Can you write up the appendix in latex, short and to the point, describing the two regex expressions and what they aim to capture

\end{verbatim}
\normalfont
\normalsize

\paragraph{English to Code.}
\scriptsize
\begin{verbatim}
shall we fuzz test the iso 8601 capture regex ? this is the regex that finds the instances, not parses

the regex is r'\d{4}-\d{1,2}-\d{1,2}(?:[T ]\d{1,2}(?::\d{1,2})*[^\s]*?)?'

def test_iso8601_regex_1n(self):
    """
    test_iso8601_regex_1 : Find *any* matching ISO-8601 like string using regex and groups to extract components

    Only valid datetimes
    """

    self.assertTrue(self.iso8601_regex_n_comparator('2025-01-01T13:14:15', 'The answer is 2025-01-01T13:14:15'))

\end{verbatim}
\normalfont
\normalsize

  \clearpage

\section{Released Datasets}
  \label{datasets}  
  
This section describes the datasets released with this paper, covering both the evaluation benchmarks and the fine-tuning datasets.

\paragraph{Translation Benchmark \& Dataset.}
\mbox{}\\
\textbf{Dataset ID:} \texttt{datetime/a.3}
\label{dataset_datetime_a3_translate}

The benchmark (\texttt{datetime/a.3}) contains 1{,}000 observations per run across 10 runs. The first 100 observations of each run are used for evaluation; the remainder are reserved for future work. Each observation is a JSON record. 

\textbf{Example observation:}
\begin{verbatim}
{
  "input": "Feb/Sun 26th,7111 ,1 AM 45"
  , "target": "7111-02-26T01:45:00",
}
\end{verbatim}

Each observation pairs a natural-language datetime with its ISO~8601 target. Input datetimes span years 1000--9999, use the \texttt{en\_US} locale, and are generated under four date schemas (\texttt{day-month-yyyy}, \texttt{day-month-weekday-yyyy}, \texttt{month-day-yyyy}, \texttt{month-day-weekday-yyyy}) and three time granularities (hours, hours-minutes, hours-minutes-seconds). Month representations are restricted to unambiguous forms (English names or abbreviations). Surface-form variation across separators, casing, and day representations (e.g.\ \texttt{29th}, \texttt{twenty}) ensures high combinatorial diversity (see Section~\ref{descriptives}).

A representative sample is shown in Table~\ref{tab:datetime-examples}.

\begin{table}[ht!]
\centering
\caption{Representative sample from the datetime/a.3 benchmark, used to evaluate all Translation tasks. Full benchmark contains 1,000 observations.}
\label{tab:datetime_a_3}
\begin{tabularx}{\textwidth}{lXl}
\toprule
Observation ID &                             Input &              Target \\
\midrule
           838 &          9 AM,Fri 29th.March-9354 & 9354-03-29T09:00:00 \\
           191 &       8:47 PM Fri Jun/twenty/3631 & 3631-06-20T20:47:00 \\
           900 &  March.Wednesday 26th/6859 ,17:13 & 6859-03-26T17:13:00 \\
           525 & October,Mon sixth\#9597 ,10 AM 10 & 9597-10-06T10:10:00 \\
           353 &     10:42:16 PM Sat 7th\#May,4895 & 4895-05-07T22:42:16 \\
\bottomrule
\end{tabularx}
\end{table}

The fine-tuning dataset (\texttt{datetime/a.3.ft\_10m}) contains 10{,}000{,}000 observations generated under identical constraints, with zero overlap against all 10 runs of the evaluation benchmark. The overlap criteria are specified in Section~\ref{descriptives}.

\paragraph{Add-250 Benchmark \& Dataset.}
\mbox{}\\
\textbf{Dataset ID:} \texttt{datetime/a3.iso8601.add.day.250.x}
\label{dataset_datetime_a3_add_250}

The benchmark (\texttt{datetime/a3.iso8601.add.day.250.x}) contains 1{,}000 observations per run across 10 runs. The first 100 observations of each run are used for evaluation; the remainder are reserved for future work.

Each observation pairs an ISO~8601 datetime with an ISO~8601 target 250 days later. Each observation is a JSON record. Additional fields exist in the full record for future-work tasks.

\textbf{Example observation:}
\begin{verbatim}
{
  "input": {
    "input_sequence": "1241-07-08T22:56:42",
  },
  "target": "1242-03-15T22:56:42",
}
\end{verbatim}

The fine-tuning dataset (\texttt{datetime/a3.iso8601.add.day.250.x\_ft\_10m}) contains 10{,}000{,}000 observations generated under identical constraints, with zero overlap at the date level (year, month, day) against all 10 runs of the evaluation benchmark. The overlap criteria are specified in Section~\ref{descriptives}.

\paragraph{Event Planning Benchmark \& Dataset.}\mbox{}\\
\textbf{Dataset ID:} \texttt{datetime/nct.1\_250}
\label{dataset_event_planning}

The benchmark (\texttt{datetime/nct.1\_250}) contains 1{,}000 observations per run across 10 runs. The first 100 observations of each run are used for evaluation; the remainder are reserved for future work.

Each observation is a JSON record. The key fields for the \emph{Event Planning} task are \texttt{start\_date} (the date preparation begins), \texttt{prep\_days} (the number of days needed for preparation), and \texttt{event\_date} (the deadline). Additional fields exist in the full record for future-work tasks.

\textbf{Example observation:}
\begin{verbatim}
{
  "input": {
    "start_date": "2034|14|April,21:54:9",
    "prep_days": 250,
    "event_date": "2034|30|November,21:54:9"
  },
  "target": "no"
}
\end{verbatim}

In this example, the start date plus 250 preparation days lands 19 days after the event date, so the target is \texttt{no}.

\textbf{User prompt template:}
\begin{quote}\ttfamily\small
I need \{prep\_days\} days to prepare for an event. I can work any day of the week. The event is on \{event\_date\}. If I start preparing on \{start\_date\}, will I be ready in time? Answer with yes or no.
\end{quote}

The fine-tuning dataset (\texttt{datetime/nct.1\_250\_ft\_1m}) contains 1{,}000{,}000 observations generated under identical constraints, with zero overlap at the date level (year, month, day) against all 10 runs of the evaluation benchmark. The overlap criteria are specified in Section~\ref{descriptives}.

  \clearpage

\section{PRIMETIME Generator Usage}
    \label{generator}
    This section provides instructions for basic use of the generator. Advanced usage is detailed in Appendix~\ref{generator_details}. The generator code and its installation instructions are published on GitHub \footnote{\url{https://github.com/LLM-PRIMETIME/Generator}}.
 
The same generator is used to produce both fine-tuning datasets and evaluation benchmarks. Separate post-processing steps, not included in this release, apply task-specific decontamination criteria and label balancing to ensure that fine-tuning and evaluation instances do not overlap and that evaluation labels are evenly distributed (see Section~\ref{tasks}).

The generator consists of two independent Python 3 classes that can be integrated into a pipeline or invoked via command-line: \texttt{datetime\_natural\_form\_tasks.py} for tasks where the input is a natural datetime representation, and \texttt{iso8601\_tasks.py} for tasks where the input is in ISO-8601 format. Both scripts take a task type and a number of observations as positional arguments, and can be further parameterized with flags and keyword arguments. A full description of all arguments is provided in Appendix~\ref{generator_details}.

Generating 1{,}000 instances for the Event Planning tasks with dates restricted to the range 2027--2036.

{\scriptsize
\begin{verbatim}
python3 datetime_natural_form_tasks.py "event_prep_1(250)" 1000 \
        --start_date 2027-01-01T00:00:00 \
        --end_date 2036-12-31T23:59:59 \
        --locale_schema "en_US" \
        --month_schema "unambiguous" \
        --remove_components 0.0 \
        --date_schemas "day-month-yyyy, day-month-weekday-yyyy, month-day-yyyy, month-day-weekday-yyyy" \
        --time_schemas "hours, hours-minutes, hours-minutes-seconds"
\end{verbatim}
}

Generating 1{,}000 instances for the Add-250 task, where the input and the output are in ISO~8601 format.

{\scriptsize
\begin{verbatim}
python3 iso8601_tasks.py add.day.250 1000 --same_month 0
\end{verbatim}
}

    \clearpage

\section{PRIMETIME Generator Details}
  \label{generator_details}  
  
\subsection{Generator: \texttt{iso8601\_tasks.py}}
\label{appendix:iso8601_generator}

This script generates training and evaluation instances
where both input and output are in ISO~8601 format.
It supports day addition, day subtraction, hour addition,
and minute addition tasks.

\paragraph{Positional arguments.}~\\

\begin{tabular}{lp{9cm}}
\texttt{output} & Task type (see task list below). \\
\texttt{num\_observations} & Number of instances to generate. \\
\end{tabular}

\paragraph{Optional arguments.}~\\

\begin{tabular}{lp{9cm}}
\texttt{-{}-same\_month} & Month-boundary filter.
    \texttt{0}: keep all (default).
    \texttt{1}: keep only same-month pairs.
    \texttt{-1}: keep only month-crossing pairs (same year). \\
\texttt{-{}-start\_date} & Lower bound for input dates (ISO~8601 format). Default: \texttt{1970-01-01T00:00:00}. \\
\texttt{-{}-end\_date} & Upper bound for input dates (ISO~8601 format). Default: \texttt{9999-12-31T00:00:00}. \\
\texttt{-{}-inputs} & Print only input sequences. \\
\texttt{-{}-targets} & Print only target sequences. \\
\texttt{-{}-preview\_rows} & Number of rows to preview from the start and end of the output. \\
\end{tabular}

\paragraph{Supported tasks.}~\\

\begin{tabular}{lp{9cm}}
\multicolumn{2}{l}{\textit{Fixed-offset addition}} \\
\texttt{add.day.1} & Add 1 day. \\
\texttt{add.day.2} & Add 2 days. \\
\texttt{add.day.10} & Add 10 days. \\
\texttt{add.day.20} & Add 20 days. \\
\texttt{add.day.50} & Add 50 days. \\
\texttt{add.day.100} & Add 100 days. \\
\texttt{add.day.250} & Add 250 days. \\
\texttt{add.hours.1000} & Add 1000 hours. \\
\texttt{add.minutes.1000} & Add 1000 minutes. \\
\\
\multicolumn{2}{l}{\textit{Variable-offset addition (random within range)}} \\
\texttt{add.days.1-250} & Add a random number of days in $[1, 250]$. \\
\texttt{add.day.1-25} & Add a random number of days in $[1, 25]$. \\
\texttt{add.day.26-50} & Add a random number of days in $[26, 50]$. \\
\texttt{add.day.51-75} & Add a random number of days in $[51, 75]$. \\
\texttt{add.day.76-100} & Add a random number of days in $[76, 100]$. \\
\texttt{add.day.250-1000} & Add a random number of days in $[250, 1000]$. \\
\texttt{add.day.9001-10000} & Add a random number of days in $[9001, 10000]$. \\
\\
\multicolumn{2}{l}{\textit{Subtraction}} \\
\texttt{subtract.day.1} & Subtract 1 day. \\
\texttt{subtract.day.2} & Subtract 2 days. \\
\end{tabular}

\paragraph{Examples.}~\\

Generate 1{,}000 Add-250 instances with month-boundary crossing enforced:

{\footnotesize
\begin{verbatim}
python3 iso8601_tasks.py add.day.250 1000 --same_month -1
\end{verbatim}
}

\noindent
Generate 500 variable-offset instances in the range $[1, 250]$ days,
restricted to dates between 2027 and 2036:

{\footnotesize
\begin{verbatim}
python3 iso8601_tasks.py add.days.1-250 500 \
        --start_date 2027-01-01T00:00:00 \
        --end_date 2036-12-31T23:59:59
\end{verbatim}
}

\noindent
Preview the first 5 rows of a subtraction task:

{\footnotesize
\begin{verbatim}
python3 iso8601_tasks.py subtract.day.1 100 --preview_rows 5
\end{verbatim}
}

\subsection{Generator: \texttt{datetime\_natural\_form\_tasks.py}}
\label{appendix:natural_form_generator}

This script generates training and evaluation instances
where the input is a natural-form datetime representation
(e.g.\ \texttt{5904|01|22 ,19:19:39 da tarde +05:00}).
Targets are either ISO~8601 datetimes (for translation and addition tasks)
or binary \textit{yes}/\textit{no} labels (for Event Planning tasks).

\paragraph{Positional arguments.}~\\

\begin{tabular}{lp{9cm}}
\texttt{output} & Task type (see task list below).
    For Event Planning tasks, arguments are passed in parentheses,
    e.g.\ \texttt{"event\_prep\_1(250)"} for fixed 250-day preparation
    or \texttt{"event\_prep\_1(200, 300)"} for variable duration. \\
\texttt{num\_observations} & Number of instances to generate. \\
\end{tabular}

\paragraph{Optional arguments.}~\\

\begin{tabular}{lp{9cm}}
\texttt{-{}-same\_month} & Month-boundary filter.
    \texttt{0}: keep all (default).
    \texttt{1}: keep only same-month pairs.
    \texttt{-1}: keep only month-crossing pairs (same year). \\
\texttt{-{}-month\_schema} & Month rendering format.
    \texttt{all}: all formats.
    \texttt{arabic}: numeric only.
    \texttt{roman}: Roman numerals.
    \texttt{unambiguous}: abbreviated and full month names only (e.g.\ Mar, March). \\
\texttt{-{}-locale\_schema} & Locale selection.
    \texttt{en\_US}: English only.
    \texttt{mini.10}: 10 common locales (default).
    \texttt{sap.dominant}: SAP-dominant locales.
    \texttt{babel.all}: all Babel-supported locales.
    \texttt{all}: all available locales. \\
\texttt{-{}-start\_date} & Lower bound for input dates (ISO~8601 format). Default: \texttt{1970-01-01T00:00:00}. \\
\texttt{-{}-end\_date} & Upper bound for input dates (ISO~8601 format). Default: \texttt{9999-12-31T00:00:00}. \\
\texttt{-{}-remove\_components} & Probability of randomly removing a datetime component from the rendered output. Default: \texttt{0.0}. \\
\texttt{-{}-date\_schemas} & Comma-separated list of date component orderings, e.g.\ \texttt{"day-month-yyyy, month-day-weekday-yyyy"}. Default: all available schemas. \\
\texttt{-{}-time\_schemas} & Comma-separated list of time granularities, e.g.\ \texttt{"hours, hours-minutes, hours-minutes-seconds"}. Default: all available schemas. \\
\texttt{-{}-inputs} & Print only input sequences. \\
\texttt{-{}-targets} & Print only target sequences. \\
\texttt{-{}-preview\_rows} & Number of rows to preview from the start and end of the output. \\
\end{tabular}

\paragraph{Supported tasks.}~\\

\begin{tabular}{lp{9cm}}
\multicolumn{2}{l}{\textit{Translation (natural form $\rightarrow$ ISO~8601)}} \\
\texttt{model} & Return the generator model name (used for pipeline testing). \\
\\
\multicolumn{2}{l}{\textit{Fixed-offset addition}} \\
\texttt{add.day.1} & Add 1 day. \\
\texttt{add.day.2} & Add 2 days. \\
\texttt{add.day.10} & Add 10 days. \\
\texttt{add.day.20} & Add 20 days. \\
\texttt{add.day.50} & Add 50 days. \\
\texttt{add.day.100} & Add 100 days. \\
\texttt{add.day.250} & Add 250 days. \\
\texttt{add.day.1000} & Add 1000 days. \\
\texttt{add.day.2500} & Add 2500 days. \\
\\
\multicolumn{2}{l}{\textit{Fixed-offset addition with few-shot exemplars}} \\
\texttt{add.day.250.i} & Add 250 days. Input includes two decontaminated few-shot examples rendered in the same format. \\
\texttt{add.day.1000.i} & Add 1000 days. Input includes two decontaminated few-shot examples. \\
\texttt{add.day.2500.i} & Add 2500 days. Input includes two decontaminated few-shot examples. \\
\\
\multicolumn{2}{l}{\textit{Subtraction}} \\
\texttt{subtract.day.1} & Subtract 1 day. \\
\texttt{subtract.day.2} & Subtract 2 days. \\
\\
\multicolumn{2}{l}{\textit{Event Planning (binary yes/no)}} \\
\texttt{event\_prep\_1(N)} & Single-hop, direct. Fixed preparation of $N$ days. \\
\texttt{event\_prep\_1(N, M)} & Single-hop, direct. Variable preparation sampled from $[N, M]$. \\
\end{tabular}

\paragraph{Examples.}~\\

Generate 1{,}000 Event Planning instances with fixed 250-day preparation,
English locale, unambiguous months, restricted to 2027--2036:

{\footnotesize
\begin{verbatim}
python3 datetime_natural_form_tasks.py "event_prep_1(250)" 1000 \
        --start_date 2027-01-01T00:00:00 \
        --end_date 2036-12-31T23:59:59 \
        --locale_schema "en_US" \
        --month_schema "unambiguous" \
        --remove_components 0.0
\end{verbatim}
}

\noindent
Generate 1{,}000 Event Planning instances with variable preparation (200--300 days):

{\footnotesize
\begin{verbatim}
python3 datetime_natural_form_tasks.py "event_prep_1(200, 300)" 1000 \
        --start_date 2027-01-01T00:00:00 \
        --end_date 2036-12-31T23:59:59
\end{verbatim}
}

\noindent
Generate 500 Add-250 translation instances with all locales
and month-boundary crossing enforced:

{\footnotesize
\begin{verbatim}
python3 datetime_natural_form_tasks.py add.day.250 500 \
        --same_month -1 \
        --locale_schema "babel.all"
\end{verbatim}
}

\noindent
Generate 500 Add-250 instances with few-shot exemplars included
in the output (for prompt construction):

{\footnotesize
\begin{verbatim}
python3 datetime_natural_form_tasks.py add.day.250.i 500 \
        --locale_schema "en_US"
\end{verbatim}
}

  \clearpage

\section{Translation Baseline}
  \label{baseline}
  \subsection{Baseline Methods}
\label{sec:baselines}

We evaluate six widely used datetime parsing libraries across three programming languages.
Each library is invoked in its default, out-of-the-box configuration. For \texttt{dateutil}, we additionally evaluate the \texttt{fuzzy=True} mode.
No pre-cleaning, format hints, or string normalisation are applied to the input.
on syntactically diverse format strings, which serve as the baseline for all translation tasks throughout this paper (\textit{datetime/a.3}).

\subsubsection{Results}
\begin{table}[ht!]
            \centering
            \caption{Baseline. Accuracy in percent of specialized datetime processing libraries on the PRIMETIME Translation task. Each library is run 10 times. Each run draws a fresh set of 100 datetimes; the set is identical for all libraries and all models throughout this paper for direct comparison. Mean $\pm 2$ SEM computed over the 10 runs, 1,000 datetimes in total. Rows sorted by descending accuracy. Best result in bold.}
            \label{datetime_translation_specialised}
            
\small            
\begin{tabular}{lc}
\toprule
 & Accuracy \\
\midrule
Python dateparser & $\textbf{36.0}_{\pm 3.0}$ \\
Python dateutil (fuzzy) & $28.0_{\pm 2.0}$ \\
R lubridate & $16.0_{\pm 2.0}$ \\
Python dateutil & $12.0_{\pm 2.0}$ \\
Python fuzzydate & $2.0_{\pm 1.0}$ \\
Javascript chrono-node & $0.0_{\pm 0.0}$ \\
\bottomrule
\end{tabular}
\end{table}

\subsubsection{Selected Libraries}

\textbf{dateutil} (Python) is the de facto standard datetime parser in the Python ecosystem. It also serves as the fallback parser for \texttt{pandas.to\_datetime} when no explicit format string is supplied.
The function \texttt{dateutil.parser.parse} accepts a string and attempts to resolve it into a \texttt{datetime} object.
Internally, the parser first tries a small set of hardcoded layouts (ISO~8601, \texttt{YYYY-MM-DD}).
If none match, it falls back to a token-based heuristic that identifies numeric and alphabetic fields by position.
We evaluate two configurations: the default mode, which raises an exception on unrecognised tokens, and the \texttt{fuzzy=True} mode, which silently discards tokens the parser cannot interpret.
In fuzzy mode, we additionally sweep over the \texttt{dayfirst} and \texttt{yearfirst} flags and retain the first successful parse.

\textbf{dateparser} (Python) is a higher-level library that layers multiple parsing strategies on top of \texttt{dateutil}.
It iterates over absolute, relative, and locale-aware template sets, attempting each in turn.
A single call to \texttt{dateparser.parse} exercises all internal strategies without additional configuration.

\textbf{lubridate} (R) provides the function \texttt{parse\_date\_time}, which accepts a vector of candidate \emph{orders}---shorthand tokens such as \texttt{Ymd}, \texttt{dbY}, or \texttt{Hp} that specify the expected field sequence.
The parser generates \texttt{strptime}-compatible format strings from each order, tries all of them against the input, and selects the match that consumes the most characters.
We supply a broad set of 25 orders covering common year-month-day, day-month-year, and month-day-year arrangements, each combined with hour, hour-minute, and hour-minute-second time components in both 12-hour and 24-hour notation.
We call lubridate from Python via \texttt{rpy2}.

\textbf{fuzzydate} (Python, Rust backend) converts free-form date strings into \texttt{datetime} objects using a pattern-matching engine implemented in Rust.
It is designed primarily for clean user inputs and relative expressions (e.g.\ ``last week'', ``1 hour ago'').
We call \texttt{fuzzydate.to\_datetime} with no additional arguments.

\textbf{chrono-node} (JavaScript) is a natural language date parser that extracts date references from surrounding prose.
Its extraction pipeline applies a sequence of regex-based parsers to identify date fragments, then passes the candidates through a chain of refiners.
We call \texttt{chrono.parseDate} from Python via a subprocess invocation of Node.js.

\subsubsection{Results}

\begin{table}[ht!]
            \centering
            \caption{Baseline. Accuracy in percent of specialized datetime processing libraries on the PRIMETIME Translation task. Each library is run 10 times. Each run draws a fresh set of 100 datetimes; the set is identical for all libraries and all models throughout this paper for direct comparison. Mean $\pm 2$ SEM computed over the 10 runs, 1,000 datetimes in total. Rows sorted by descending accuracy. Best result in bold.}
            \label{datetime_translation_specialised}
            
\small            
\begin{tabular}{lc}
\toprule
 & Accuracy \\
\midrule
Python dateparser & $\textbf{36.0}_{\pm 3.0}$ \\
Python dateutil (fuzzy) & $28.0_{\pm 2.0}$ \\
R lubridate & $16.0_{\pm 2.0}$ \\
Python dateutil & $12.0_{\pm 2.0}$ \\
Python fuzzydate & $2.0_{\pm 1.0}$ \\
Javascript chrono-node & $0.0_{\pm 0.0}$ \\
\bottomrule
\end{tabular}
\end{table}

The strongest baseline is \texttt{dateparser} at 0.36 accuracy.
No rule-based parser exceeds this threshold.

\subsubsection{Analysis}

All six libraries are instances of the same underlying approach: template enumeration.
Each maintains a finite inventory of \texttt{strptime}-style format skeletons and attempts to match the input against them.
They differ in the size of the inventory, the matching heuristic, and the tolerance for unrecognised tokens, but the parsing mechanism is identical in kind.

Three failure modes recur across all libraries.
First, novel separators such as \texttt{|} are not part of any built-in template and cause immediate parse failure in strict mode.
Second, ordinal suffixes (\texttt{22nd}, \texttt{3rd}, \texttt{1st}) are not stripped by any library except in fuzzy mode, where the suffix is discarded along with the semantic content it decorates.
Third, abbreviated or non-standard time representations (\texttt{1~PM} without \texttt{:00}) are handled only when the library's template set includes a bare-hour variant.

The combination of these three noise types is multiplicative.
A format string that contains a novel separator, an ordinal day, and a bare hour simultaneously falls outside the template space of every library tested.
The number of possible separator, suffix, ordering, and omission variants grows combinatorially, and no finite template set can cover the space.

\texttt{chrono-node} exhibits an additional failure mode: temporal anchoring bias.
When the parser cannot confidently extract a year, it defaults to the current year as a reference date.
Inputs with unusual years (e.g.\ \texttt{3392}) are silently rewritten to the present year, producing a confident but incorrect result.

  \clearpage

  \section{All Translation Results}
  \label{key_translation_tables}  
  
This section reports the full \emph{Translation} results referenced in the main text.

\begin{table}[ht!]
            \centering
            \caption{Baseline. Accuracy in percent of specialized datetime processing libraries on the PRIMETIME Translation task. Each library is run 10 times. Each run draws a fresh set of 100 datetimes; the set is identical for all libraries and all models throughout this paper for direct comparison. Mean $\pm 2$ SEM computed over the 10 runs, 1,000 datetimes in total. Rows sorted by descending accuracy. Best result in bold.}
            \label{datetime_translation_specialised}
            
\small            
\begin{tabular}{lc}
\toprule
 & Accuracy \\
\midrule
Python dateparser & $\textbf{36.0}_{\pm 3.0}$ \\
Python dateutil (fuzzy) & $28.0_{\pm 2.0}$ \\
R lubridate & $16.0_{\pm 2.0}$ \\
Python dateutil & $12.0_{\pm 2.0}$ \\
Python fuzzydate & $2.0_{\pm 1.0}$ \\
Javascript chrono-node & $0.0_{\pm 0.0}$ \\
\bottomrule
\end{tabular}
\end{table}

The proprietary model results are reported in the main paper in Table~\ref{datetime_translation_prop}. 

The open-weight model results are reported in the main paper in Table~\ref{datetime_translation_vllm}.

The fine-tuned model results are reported in the main paper in Table~\ref{datetime_translation_unsloth}.

  \clearpage

\section{All Computation Results}
  \label{key_computation_tables}  
  
This section reports the full \emph{Add-250} results referenced in the main text.

The proprietary model results are reported in the main paper in Table~\ref{datetime_computation_add_250_prop}. 

\begin{table}[ht!]
            \centering
            \caption{Accuracy of small open-weight language models on the PRIMETIME Add-250 task. Sorted by descending maximum accuracy across all prompt styles. See Appendix~\ref{datetime_computation_add_250_vllm_repro} for full reproducibility details.}
            \label{datetime_computation_add_250_vllm}
            
            \resizebox{\textwidth}{!}{%
\begin{tabular}{lrrrrrrrrrrr}
\toprule
 & Direct-1 & Direct-2 & Direct-1 ISO & Direct-2 ISO & FS-1 & FS-2 & CoT-1 & CoT-1 + FS-1 & CoT-1 + FS-2 & Min & Max \\
\midrule
Qwen3 4B Greedy & $\mathrm{66.0}_{{\pm nan}}$ & $\mathrm{59.0}_{{\pm nan}}$ & $\mathrm{76.0}_{{\pm nan}}$ & $\mathrm{63.0}_{{\pm nan}}$ & $\mathrm{59.0}_{{\pm nan}}$ & $\mathrm{55.0}_{{\pm nan}}$ & $\mathrm{60.0}_{{\pm nan}}$ & $\mathrm{60.0}_{{\pm nan}}$ & $\mathrm{65.0}_{{\pm nan}}$ & $\mathrm{55.0}_{{\pm nan}}$ & $\mathrm{76.0}_{{\pm nan}}$ \\
Qwen3 1.7B Greedy & $\mathrm{14.0}_{{\pm nan}}$ & $\mathrm{13.0}_{{\pm nan}}$ & $\mathrm{23.0}_{{\pm nan}}$ & $\mathrm{21.0}_{{\pm nan}}$ & $\mathrm{17.0}_{{\pm nan}}$ & $\mathrm{18.0}_{{\pm nan}}$ & $\mathrm{18.0}_{{\pm nan}}$ & $\mathrm{25.0}_{{\pm nan}}$ & $\mathrm{22.0}_{{\pm nan}}$ & $\mathrm{13.0}_{{\pm nan}}$ & $\mathrm{25.0}_{{\pm nan}}$ \\
Gemma 3 4B Greedy & $\mathrm{0.0}_{{\pm nan}}$ & $\mathrm{1.0}_{{\pm nan}}$ & $\mathrm{0.0}_{{\pm nan}}$ & $\mathrm{0.0}_{{\pm nan}}$ & $\mathrm{0.0}_{{\pm nan}}$ & $\mathrm{0.0}_{{\pm nan}}$ & $\mathrm{0.0}_{{\pm nan}}$ & $\mathrm{0.0}_{{\pm nan}}$ & $\mathrm{0.0}_{{\pm nan}}$ & $\mathrm{0.0}_{{\pm nan}}$ & $\mathrm{1.0}_{{\pm nan}}$ \\
Qwen3 0.6B Greedy & $\mathrm{0.0}_{{\pm nan}}$ & $\mathrm{1.0}_{{\pm nan}}$ & $\mathrm{0.0}_{{\pm nan}}$ & $\mathrm{0.0}_{{\pm nan}}$ & $\mathrm{0.0}_{{\pm nan}}$ & $\mathrm{0.0}_{{\pm nan}}$ & $\mathrm{1.0}_{{\pm nan}}$ & $\mathrm{1.0}_{{\pm nan}}$ & $\mathrm{1.0}_{{\pm nan}}$ & $\mathrm{0.0}_{{\pm nan}}$ & $\mathrm{1.0}_{{\pm nan}}$ \\
Falcon3 3B Greedy & $\mathrm{0.0}_{{\pm nan}}$ & $\mathrm{0.0}_{{\pm nan}}$ & $\mathrm{0.0}_{{\pm nan}}$ & $\mathrm{0.0}_{{\pm nan}}$ & $\mathrm{0.0}_{{\pm nan}}$ & $\mathrm{0.0}_{{\pm nan}}$ & $\mathrm{0.0}_{{\pm nan}}$ & $\mathrm{0.0}_{{\pm nan}}$ & $\mathrm{0.0}_{{\pm nan}}$ & $\mathrm{0.0}_{{\pm nan}}$ & $\mathrm{0.0}_{{\pm nan}}$ \\
\bottomrule
\end{tabular}}
            
\end{table}

The fine-tuned model results are reported in the main paper in Table~\ref{datetime_computation_add_250_unsloth}.

  \clearpage

\section{All Event Planning Results}
  \label{key_nc_tables}

The proprietary model results are reported in the main paper in Table~\ref{datetime_natural_context_prop_llm}. 

\begin{table}[ht!]
            \centering
            \caption{Accuracy of small open-weight language models on the PRIMETIME Event Planning task. Sorted by descending maximum accuracy across all prompt styles. See Appendix~\ref{datetime_natural_context_vllm_repro} for full reproducibility details.}
            \label{datetime_nc_250_vllm}
            
            \resizebox{\textwidth}{!}{%
\begin{tabular}{lrrrrrrrrrr}
\toprule
 & Direct-1 & Direct-2 & Direct-3 & FS-1 & FS-2 & CoT-1 & CoT-1 + FS-1 & CoT-1 + FS-2 & Min & Max \\
\midrule
Qwen3 4B Greedy & $\mathrm{67.0}_{{\pm 3.9}}$ & $\mathrm{81.4}_{{\pm 2.2}}$ & $\mathrm{34.3}_{{\pm 0.9}}$ & $\mathrm{84.5}_{{\pm 2.7}}$ & $\mathrm{82.4}_{{\pm 1.9}}$ & $\mathrm{81.1}_{{\pm 1.5}}$ & $\mathrm{85.8}_{{\pm 1.8}}$ & $\mathrm{89.0}_{{\pm 1.5}}$ & $\mathrm{34.3}_{{\pm 0.9}}$ & $\mathrm{89.1}_{{\pm 1.4}}$ \\
Qwen3 1.7B Greedy & $\mathrm{43.2}_{{\pm 2.5}}$ & $\mathrm{51.5}_{{\pm 4.4}}$ & $\mathrm{54.2}_{{\pm 3.8}}$ & $\mathrm{53.4}_{{\pm 2.3}}$ & $\mathrm{50.4}_{{\pm 0.4}}$ & $\mathrm{63.8}_{{\pm 2.5}}$ & $\mathrm{65.4}_{{\pm 2.6}}$ & $\mathrm{65.9}_{{\pm 2.8}}$ & $\mathrm{43.0}_{{\pm 2.2}}$ & $\mathrm{69.0}_{{\pm 1.5}}$ \\
SmolLM2 1.7B Greedy & $\mathrm{50.0}_{{\pm 0.0}}$ & $\mathrm{50.0}_{{\pm 1.8}}$ & $\mathrm{50.0}_{{\pm 0.0}}$ & $\mathrm{52.2}_{{\pm 3.8}}$ & $\mathrm{50.0}_{{\pm 0.0}}$ & $\mathrm{50.0}_{{\pm 0.0}}$ & $\mathrm{50.0}_{{\pm 0.8}}$ & $\mathrm{50.0}_{{\pm 0.0}}$ & $\mathrm{47.8}_{{\pm 0.9}}$ & $\mathrm{54.3}_{{\pm 2.8}}$ \\
Gemma 3 1B Greedy & $\mathrm{48.4}_{{\pm 2.9}}$ & $\mathrm{49.6}_{{\pm 4.0}}$ & $\mathrm{50.0}_{{\pm 0.0}}$ & $\mathrm{49.8}_{{\pm 3.3}}$ & $\mathrm{50.2}_{{\pm 0.8}}$ & $\mathrm{50.1}_{{\pm 1.2}}$ & $\mathrm{49.9}_{{\pm 0.4}}$ & $\mathrm{50.0}_{{\pm 0.0}}$ & $\mathrm{46.2}_{{\pm 2.0}}$ & $\mathrm{53.4}_{{\pm 2.7}}$ \\
Phi 4 Mini Greedy & $\mathrm{48.1}_{{\pm 1.6}}$ & $\mathrm{49.7}_{{\pm 0.3}}$ & $\mathrm{49.9}_{{\pm 0.2}}$ & $\mathrm{48.4}_{{\pm 2.0}}$ & $\mathrm{49.5}_{{\pm 2.0}}$ & $\mathrm{49.0}_{{\pm 1.1}}$ & $\mathrm{49.7}_{{\pm 3.6}}$ & $\mathrm{50.9}_{{\pm 3.2}}$ & $\mathrm{45.8}_{{\pm 1.7}}$ & $\mathrm{53.0}_{{\pm 2.4}}$ \\
\bottomrule
\end{tabular}}
            
            \end{table}

The fine-tuned model results are reported in the main paper in Table~\ref{datetime_event_planning_250_unsloth}.

  \clearpage

\section{Failure Analysis}
  \label{failure_analysis}
  
\begin{landscape}
The failure analysis framework uses two kinds of evaluators. First, a deterministic Python function is used to compute the ground truth where possible; where this is not feasible, a regex expression is used as a proxy. Second, a Qwen3.5 35B LLM judge \citep{zheng2023judgingllmasajudgemtbenchchatbot}, served via Ollama \footnote{qwen3.5:35b at Q4\_K\_M quantization, see \url{https://ollama.com/library/qwen3.5:35b}}, is used for questions where failure categories are ambiguous, for example if an ISO-8601 was produced in the output sequence, of to evaluate the use of scratchpads. To remove parsing and evaluation ambiguity, the LLM judge is prompted with closed-form questions constrained to binary yes/no outputs, extracted via the same regex parser used for the \emph{Event Planning} tasks in Section~\ref{tasks}.

Each cell reports the per-class F1 of the LLM judge against the ground truth, followed by the judge's prediction distribution in parentheses. The first score is the yes-class F1, the second the no-class F1, and the parenthetical pair gives the counts of items the judge labelled yes and no. Ground truth class balance varies across slices, so per-class F1 is reported rather than a single aggregate score. A slice is treated as reliable, and used for failure-mode claims in the main text, only when both per-class F1 scores meet or exceed the pre-registered threshold of $0.90$.

\paragraph{How to read a cell.}

Each cell is formatted as

\begin{center}
\texttt{\textit{yes-F1} / \textit{no-F1} (\textit{n-yes} / \textit{n-no})}
\end{center}

\noindent and reports four numbers describing how well the evaluator agreed with the ground truth on a single (model, prompt-style) slice of $n=100$ items.

\textbf{Random baseline.} As a noise floor, a random evaluator predicting yes or no with equal probability on a balanced ground truth would yield, in expectation, the cell \texttt{0.5 / 0.5 (50/50)}: per-class F1 of $0.5$ on each class, with the $50/50$ prediction split reflecting random labelling by the LLM judge rather than agreement with the ground truth. Cells substantially above this floor on both classes indicate genuine evaluator signal; cells near it indicate the evaluator is no better than chance. The failure analysis framework was tested against random noise to confirm this expected behaviour.

\textbf{Worked example.} The cell \texttt{1.0 / 0.0 (96/4)} for Claude~4.0~Sonnet on the Direct-1 prompt is read as follows. The evaluator labelled $96$ of the $100$ items as yes and $4$ as no (the parenthetical pair). The yes-class F1 of $1.0$ indicates near-perfect agreement with ground truth on the items the evaluator labelled yes. The no-class F1 of $0.0$ does \emph{not} mean the evaluator failed on the no-class: rather, the underlying ground-truth distribution for this slice is itself heavily skewed --- almost every output produced by Claude~4.0~Sonnet contains an ISO-8601 datetime, so there are essentially no actual-no items against which to score the no-class. Per-class F1 is undefined when one class has zero ground-truth support, and our convention reports such cases as $0.0$. The four no-predictions made by the evaluator are therefore false positives against an all-yes ground truth, dragging the no-class precision (and hence F1) to zero.

\textbf{Reading the parenthetical correctly.} The two counts in parentheses describe the LLM judge's predictions, not the ground-truth class balance. A heavy split such as $98/2$ does not necessarily mean the data is skewed in the same direction; it means the evaluator predicted one class far more often than the other. As the worked example shows, a no-class F1 of $0.0$ can arise either from genuine evaluator failure on the no-class \emph{or} from an absent no-class in the ground truth.

\subsection{Proprietary models on the Translate Task}
\begin{table}[ht!]
\tiny
\caption{Failure analysis of proprietary models on the Translation task. Judge question: was an ISO-8601 date predicted in the model output? LLM-as-Judge: Qwen3.5-35B (Ollama). Ground truth: the same regex used for the task's evaluation, given in Appendix~\ref{regexes}. $n=100$ per cell.}
\label{datetime_v2/translation_tasks_2_regex_n_prop_only_datetime_v2/translation_tasks_2_regex_n_prop_only}
\begin{tabular}{lllllllll}
\toprule
 & Direct-1 & Direct-2 & Direct-1 ISO & FS-1 & FS-2 & CoT & CoT FS-1 & CoT FS-2 \\
Model &  &  &  &  &  &  &  &  \\
\midrule
Anthropic claude-4.0-sonnet & 1.0 / 0.0 (100/0) & 1.0 / 1.0 (96/4) & 1.0 / 0.0 (100/0) & 1.0 / 0.0 (100/0) & 1.0 / 0.0 (100/0) & 1.0 / 0.0 (100/0) & 1.0 / 0.0 (100/0) & 1.0 / 0.0 (100/0) \\
Anthropic claude-4.0-opus & 1.0 / 1.0 (98/2) & 1.0 / 1.0 (91/9) & 1.0 / 1.0 (97/3) & 1.0 / 1.0 (95/5) & 1.0 / 1.0 (97/3) & 1.0 / 1.0 (97/3) & 1.0 / 0.5 (99/1) & 1.0 / 0.0 (100/0) \\
Anthropic claude-4.1-opus & 1.0 / 1.0 (98/2) & 1.0 / 1.0 (81/19) & 1.0 / 1.0 (95/5) & 1.0 / 1.0 (94/6) & 1.0 / 1.0 (97/3) & 1.0 / 0.7 (98/2) & 1.0 / 0.7 (99/1) & 1.0 / 0.0 (100/0) \\
OpenAI gpt-3.5-turbo & 1.0 / 0.0 (99/1) & 1.0 / 1.0 (99/1) & 1.0 / 0.0 (100/0) & 1.0 / 1.0 (99/1) & 1.0 / 0.0 (100/0) & 1.0 / 0.0 (100/0) & 1.0 / 0.0 (100/0) & 1.0 / 0.0 (100/0) \\
OpenAI gpt-4.0-turbo & 1.0 / 0.8 (85/15) & 1.0 / 0.9 (93/7) & 1.0 / 1.0 (94/6) & 1.0 / 1.0 (95/5) & 1.0 / 1.0 (99/1) & 1.0 / 0.8 (97/3) & 1.0 / 0.0 (100/0) & 1.0 / 0.0 (97/3) \\
OpenAI gpt-4o-mini & 1.0 / 0.7 (98/2) & 1.0 / 0.9 (92/8) & 1.0 / 1.0 (98/2) & 1.0 / 1.0 (98/2) & 1.0 / 1.0 (99/1) & 1.0 / 1.0 (99/1) & 1.0 / 1.0 (99/1) & 1.0 / 0.0 (99/1) \\
\bottomrule
\end{tabular}
\end{table}
\begin{table}[ht!]
\tiny
\caption{Failure analysis of proprietary models on the Translation task. Judge question: was an ISO-8601 datetime predicted in the model output? LLM-as-Judge: Qwen3.5-35B (Ollama). Ground truth: the same regex used for the task's evaluation, given in Appendix~\ref{regexes}. $n=100$ per cell.}
\label{datetime_v2/translation_tasks_2_regex_n_prop_only_datetime_v2/translation_tasks_2_regex_n_prop_only}
\begin{tabular}{lllllllll}
\toprule
 & Direct-1 & Direct-2 & Direct-1 ISO & FS-1 & FS-2 & CoT & CoT FS-1 & CoT FS-2 \\
Model &  &  &  &  &  &  &  &  \\
\midrule
Anthropic claude-4.0-sonnet & 1.0 / 0.0 (96/4) & 1.0 / 1.0 (96/4) & 1.0 / 0.0 (100/0) & 1.0 / 0.0 (100/0) & 1.0 / 0.0 (100/0) & 1.0 / 0.0 (100/0) & 1.0 / 0.0 (100/0) & 1.0 / 1.0 (99/1) \\
Anthropic claude-4.0-opus & 1.0 / 0.7 (96/4) & 1.0 / 0.9 (88/12) & 1.0 / 1.0 (97/3) & 1.0 / 1.0 (95/5) & 1.0 / 1.0 (97/3) & 1.0 / 1.0 (97/3) & 1.0 / 1.0 (97/3) & 1.0 / 0.0 (100/0) \\
Anthropic claude-4.1-opus & 1.0 / 0.6 (95/5) & 1.0 / 0.9 (79/21) & 1.0 / 1.0 (95/5) & 1.0 / 1.0 (94/6) & 1.0 / 1.0 (97/3) & 1.0 / 1.0 (99/1) & 1.0 / 0.5 (98/2) & 1.0 / 0.0 (100/0) \\
OpenAI gpt-3.5-turbo & 1.0 / 0.0 (97/3) & 1.0 / 0.3 (95/5) & 1.0 / 0.0 (100/0) & 1.0 / 1.0 (99/1) & 1.0 / 0.0 (100/0) & 1.0 / 0.0 (100/0) & 1.0 / 0.0 (100/0) & 1.0 / 0.0 (99/1) \\
OpenAI gpt-4.0-turbo & \unreliable{0.9 / 0.5 (66/34)} & 0.9 / 0.6 (85/15) & 1.0 / 0.8 (90/10) & 1.0 / 0.7 (91/9) & 1.0 / 0.5 (97/3) & 1.0 / 0.4 (91/9) & 1.0 / 0.0 (94/6) & 1.0 / 0.0 (94/6) \\
OpenAI gpt-4o-mini & 1.0 / 0.7 (98/2) & 1.0 / 0.9 (91/9) & 1.0 / 1.0 (98/2) & 1.0 / 1.0 (98/2) & 1.0 / 1.0 (99/1) & 1.0 / 1.0 (99/1) & 1.0 / 1.0 (99/1) & 1.0 / 0.0 (99/1) \\
\bottomrule
\end{tabular}
\end{table}
\begin{table}[ht!]
\tiny
\caption{Failure analysis of proprietary models on the Translation task. Judge question: was more than one ISO-8601 datetime predicted in the model output? LLM-as-Judge: Qwen3.5-35B (Ollama). Ground truth: the \emph{ISO~8601 capture} regex from Appendix~\ref{regexes}, applied to count matches in the output sequence. $n=100$ per cell.}
\label{datetime_v2/translation_tasks_2_regex_n_prop_only_datetime_v2/translation_tasks_2_regex_n_prop_only}
\begin{tabular}{lllllllll}
\toprule
 & Direct-1 & Direct-2 & Direct-1 ISO & FS-1 & FS-2 & CoT & CoT FS-1 & CoT FS-2 \\
Model &  &  &  &  &  &  &  &  \\
\midrule
Anthropic claude-4.0-sonnet & \insufficient{0.0 / 1.0 (0/100)} & \insufficient{0.0 / 1.0 (0/100)} & \insufficient{0.0 / 1.0 (0/100)} & \insufficient{0.0 / 1.0 (0/100)} & \insufficient{0.0 / 1.0 (0/100)} & \insufficient{0.0 / 1.0 (0/100)} & \insufficient{0.0 / 1.0 (0/100)} & \insufficient{0.0 / 1.0 (0/100)} \\
Anthropic claude-4.0-opus & \insufficient{0.0 / 1.0 (0/100)} & \insufficient{0.0 / 1.0 (0/100)} & \insufficient{0.0 / 1.0 (0/100)} & \insufficient{0.0 / 1.0 (0/100)} & \insufficient{0.0 / 1.0 (0/100)} & \insufficient{0.0 / 1.0 (0/100)} & \insufficient{0.0 / 1.0 (0/100)} & \insufficient{0.0 / 1.0 (0/100)} \\
Anthropic claude-4.1-opus & \insufficient{0.0 / 1.0 (0/100)} & \insufficient{0.0 / 1.0 (0/100)} & \insufficient{0.0 / 1.0 (0/100)} & \insufficient{0.0 / 1.0 (0/100)} & \insufficient{0.0 / 1.0 (0/100)} & \insufficient{0.0 / 1.0 (0/100)} & \insufficient{0.0 / 1.0 (0/100)} & \insufficient{0.0 / 1.0 (0/100)} \\
OpenAI gpt-3.5-turbo & \insufficient{0.0 / 1.0 (0/100)} & \insufficient{0.0 / 1.0 (0/100)} & \insufficient{0.0 / 1.0 (0/100)} & \insufficient{0.0 / 1.0 (0/100)} & \insufficient{0.0 / 1.0 (0/100)} & \insufficient{0.0 / 1.0 (0/100)} & \insufficient{0.0 / 1.0 (0/100)} & \insufficient{0.0 / 1.0 (0/100)} \\
OpenAI gpt-4.0-turbo & \insufficient{1.0 / 1.0 (1/99)} & \insufficient{0.0 / 1.0 (0/100)} & \insufficient{0.0 / 1.0 (0/100)} & \insufficient{0.0 / 1.0 (0/100)} & \insufficient{0.0 / 1.0 (0/100)} & \insufficient{0.0 / 1.0 (0/100)} & \insufficient{0.0 / 1.0 (0/100)} & \insufficient{0.0 / 1.0 (1/99)} \\
OpenAI gpt-4o-mini & \insufficient{0.0 / 1.0 (0/100)} & \insufficient{0.0 / 1.0 (0/100)} & \insufficient{0.0 / 1.0 (0/100)} & \insufficient{0.0 / 1.0 (0/100)} & \insufficient{0.0 / 1.0 (0/100)} & \insufficient{0.0 / 1.0 (0/100)} & \insufficient{0.0 / 1.0 (0/100)} & \insufficient{0.0 / 1.0 (0/100)} \\
\bottomrule
\end{tabular}
\end{table}
\begin{table}[ht!]
\tiny
\caption{Failure analysis of proprietary models on the Translation task. Judge question: did the model use a scratchpad before producing its answer? LLM-as-Judge: Qwen3.5-35B (Ollama). Ground truth: output length exceeds 50 characters, used as a proxy for the presence of intermediate reasoning. $n=100$ per cell.}
\label{datetime_v2/translation_tasks_2_regex_n_prop_only_datetime_v2/translation_tasks_2_regex_n_prop_only}
\begin{tabular}{lllllllll}
\toprule
 & Direct-1 & Direct-2 & Direct-1 ISO & FS-1 & FS-2 & CoT & CoT FS-1 & CoT FS-2 \\
Model &  &  &  &  &  &  &  &  \\
\midrule
Anthropic claude-4.0-sonnet & \insufficient{0.0 / 1.0 (0/100)} & \insufficient{1.0 / 1.0 (4/96)} & \insufficient{0.0 / 1.0 (0/100)} & \insufficient{0.0 / 1.0 (0/100)} & \insufficient{0.0 / 1.0 (0/100)} & 1.0 / 0.0 (98/2) & 1.0 / 0.0 (99/1) & 1.0 / 0.0 (100/0) \\
Anthropic claude-4.0-opus & \insufficient{1.0 / 1.0 (2/98)} & 1.0 / 1.0 (9/91) & \insufficient{1.0 / 1.0 (3/97)} & 1.0 / 1.0 (5/95) & \insufficient{1.0 / 1.0 (3/97)} & 1.0 / 0.0 (94/6) & 1.0 / 1.0 (86/14) & 1.0 / 1.0 (43/57) \\
Anthropic claude-4.1-opus & \insufficient{1.0 / 1.0 (2/98)} & 1.0 / 1.0 (19/81) & 1.0 / 1.0 (5/95) & 1.0 / 1.0 (6/94) & \insufficient{1.0 / 1.0 (4/96)} & 1.0 / 0.0 (98/2) & 1.0 / 0.0 (100/0) & 1.0 / 1.0 (91/9) \\
OpenAI gpt-3.5-turbo & \insufficient{0.0 / 1.0 (0/100)} & \insufficient{1.0 / 1.0 (1/99)} & \insufficient{0.0 / 1.0 (0/100)} & \insufficient{0.0 / 1.0 (0/100)} & \insufficient{0.0 / 1.0 (0/100)} & \insufficient{0.0 / 1.0 (0/100)} & \insufficient{0.0 / 1.0 (0/100)} & \insufficient{0.0 / 1.0 (0/100)} \\
OpenAI gpt-4.0-turbo & \unreliable{0.3 / 0.5 (14/86)} & \unreliable{0.3 / 0.7 (8/92)} & \unreliable{0.5 / 0.9 (7/93)} & \unreliable{0.5 / 0.9 (6/94)} & \unreliable{0.3 / 1.0 (1/99)} & \unreliable{0.3 / 0.9 (5/95)} & \unreliable{0.0 / 0.8 (0/100)} & \unreliable{0.1 / 0.8 (2/98)} \\
OpenAI gpt-4o-mini & \insufficient{0.7 / 1.0 (1/99)} & \unreliable{0.8 / 1.0 (7/93)} & \insufficient{0.0 / 1.0 (1/99)} & \insufficient{0.8 / 1.0 (2/98)} & \insufficient{0.0 / 1.0 (1/99)} & \unreliable{0.6 / 0.9 (10/90)} & \insufficient{0.8 / 1.0 (2/98)} & \insufficient{0.7 / 1.0 (1/99)} \\
\bottomrule
\end{tabular}
\end{table}
\begin{table}[ht!]
\tiny
\caption{Failure analysis of proprietary models on the Translation task. Question: was the predicted ISO-8601 datetime within one day of the target? Evaluation is fully deterministic: the predicted datetime is extracted via the \emph{ISO~8601 capture} regex from Appendix~\ref{regexes} and compared to the target with a one-day tolerance. $n=100$ per cell.}
\label{datetime_v2/translation_tasks_2_regex_n_prop_only_datetime_v2/translation_tasks_2_regex_n_prop_only}
\begin{tabular}{lllllllll}
\toprule
 & Direct-1 & Direct-2 & Direct-1 ISO & FS-1 & FS-2 & CoT & CoT FS-1 & CoT FS-2 \\
Model &  &  &  &  &  &  &  &  \\
\midrule
Anthropic claude-4.0-sonnet & 87/13 & 82/18 & 89/11 & 89/11 & 89/11 & 91/9 & 93/7 & 93/7 \\
Anthropic claude-4.0-opus & 90/10 & 82/18 & 89/11 & 90/10 & 91/9 & 89/11 & 92/8 & 94/6 \\
Anthropic claude-4.1-opus & 87/13 & 75/25 & 89/11 & 89/11 & 93/7 & 92/8 & 94/6 & 94/6 \\
OpenAI gpt-3.5-turbo & 90/10 & 89/11 & 93/7 & 93/7 & 93/7 & 92/8 & 94/6 & 93/7 \\
OpenAI gpt-4.0-turbo & 84/16 & 87/13 & 90/10 & 89/11 & 94/6 & 91/9 & 92/8 & 95/5 \\
OpenAI gpt-4o-mini & 71/29 & 74/26 & 82/18 & 82/18 & 88/12 & 84/16 & 83/17 & 89/11 \\
OpenAI gpt-5 & 96/4 & 94/6 & 96/4 & 99/1 & 98/2 & 94/6 & 97/3 & 99/1 \\
OpenAI gpt-5-mini & 94/6 & 93/7 & 95/5 & 96/4 & 95/5 & 96/4 & 95/5 & 94/6 \\
OpenAI gpt-5-nano & 88/12 & 84/16 & 87/13 & 90/10 & 93/7 & 89/11 & 92/8 & 94/6 \\
Gemini gemini-2.5-flash-lite & 73/27 & 75/25 & 79/21 & 69/31 & 87/13 & 86/14 & 85/15 & 90/10 \\
Gemini gemini-2.5-flash & 88/12 & 85/15 & 90/10 & 89/11 & 90/10 & 89/11 & 89/11 & 91/9 \\
Gemini gemini-2.5-pro & 86/14 & 90/10 & 91/9 & 91/9 & 94/6 & 91/9 & 92/8 & 93/7 \\
\bottomrule
\end{tabular}
\end{table}
\begin{table}[ht!]
\tiny
\caption{Failure analysis of proprietary models on the Translation task. Question: was the predicted ISO-8601 datetime within ten days day of the target? Evaluation is fully deterministic: the predicted datetime is extracted via the \emph{ISO~8601 capture} regex from Appendix~\ref{regexes} and compared to the target with a ten-day tolerance. $n=100$ per cell.}
\label{datetime_v2/translation_tasks_2_regex_n_prop_only_datetime_v2/translation_tasks_2_regex_n_prop_only}
\begin{tabular}{lllllllll}
\toprule
 & Direct-1 & Direct-2 & Direct-1 ISO & FS-1 & FS-2 & CoT & CoT FS-1 & CoT FS-2 \\
Model &  &  &  &  &  &  &  &  \\
\midrule
Anthropic claude-4.0-sonnet & 92/8 & 86/14 & 94/6 & 94/6 & 94/6 & 95/5 & 98/2 & 96/4 \\
Anthropic claude-4.0-opus & 95/5 & 87/13 & 94/6 & 94/6 & 95/5 & 94/6 & 95/5 & 99/1 \\
Anthropic claude-4.1-opus & 93/7 & 78/22 & 94/6 & 93/7 & 96/4 & 96/4 & 97/3 & 99/1 \\
OpenAI gpt-3.5-turbo & 97/3 & 92/8 & 99/1 & 98/2 & 98/2 & 98/2 & 98/2 & 99/1 \\
OpenAI gpt-4.0-turbo & 86/14 & 90/10 & 92/8 & 92/8 & 98/2 & 94/6 & 95/5 & 98/2 \\
OpenAI gpt-4o-mini & 76/24 & 77/23 & 87/13 & 87/13 & 93/7 & 90/10 & 88/12 & 95/5 \\
OpenAI gpt-5 & 97/3 & 96/4 & 99/1 & 99/1 & 99/1 & 97/3 & 100/0 & 99/1 \\
OpenAI gpt-5-mini & 97/3 & 96/4 & 98/2 & 99/1 & 99/1 & 99/1 & 99/1 & 98/2 \\
OpenAI gpt-5-nano & 92/8 & 87/13 & 91/9 & 94/6 & 97/3 & 92/8 & 96/4 & 98/2 \\
Gemini gemini-2.5-flash-lite & 77/23 & 78/22 & 82/18 & 72/28 & 93/7 & 90/10 & 88/12 & 94/6 \\
Gemini gemini-2.5-flash & 93/7 & 90/10 & 94/6 & 94/6 & 96/4 & 94/6 & 94/6 & 97/3 \\
Gemini gemini-2.5-pro & 91/9 & 94/6 & 94/6 & 95/5 & 98/2 & 95/5 & 96/4 & 98/2 \\
\bottomrule
\end{tabular}
\end{table}
\begin{table}[ht!]
\tiny
\caption{Failure analysis of proprietary models on the Translation task. Question: was the predicted ISO-8601 datetime within hundred days day of the target? Evaluation is fully deterministic: the predicted datetime is extracted via the \emph{ISO~8601 capture} regex from Appendix~\ref{regexes} and compared to the target with a hundred-day tolerance. $n=100$ per cell.}
\label{datetime_v2/translation_tasks_2_regex_n_prop_only_datetime_v2/translation_tasks_2_regex_n_prop_only}
\begin{tabular}{lllllllll}
\toprule
 & Direct-1 & Direct-2 & Direct-1 ISO & FS-1 & FS-2 & CoT & CoT FS-1 & CoT FS-2 \\
Model &  &  &  &  &  &  &  &  \\
\midrule
Anthropic claude-4.0-sonnet & 94/6 & 88/12 & 96/4 & 96/4 & 96/4 & 97/3 & 98/2 & 96/4 \\
Anthropic claude-4.0-opus & 95/5 & 88/12 & 94/6 & 94/6 & 95/5 & 96/4 & 96/4 & 99/1 \\
Anthropic claude-4.1-opus & 94/6 & 79/21 & 94/6 & 93/7 & 96/4 & 97/3 & 97/3 & 99/1 \\
OpenAI gpt-3.5-turbo & 98/2 & 93/7 & 100/0 & 99/1 & 99/1 & 99/1 & 99/1 & 100/0 \\
OpenAI gpt-4.0-turbo & 86/14 & 90/10 & 92/8 & 92/8 & 98/2 & 94/6 & 95/5 & 98/2 \\
OpenAI gpt-4o-mini & 78/22 & 79/21 & 88/12 & 88/12 & 94/6 & 91/9 & 88/12 & 97/3 \\
OpenAI gpt-5 & 97/3 & 96/4 & 99/1 & 99/1 & 99/1 & 97/3 & 100/0 & 99/1 \\
OpenAI gpt-5-mini & 98/2 & 97/3 & 98/2 & 99/1 & 99/1 & 99/1 & 99/1 & 99/1 \\
OpenAI gpt-5-nano & 93/7 & 88/12 & 92/8 & 95/5 & 98/2 & 93/7 & 98/2 & 99/1 \\
Gemini gemini-2.5-flash-lite & 78/22 & 80/20 & 84/16 & 72/28 & 96/4 & 93/7 & 90/10 & 97/3 \\
Gemini gemini-2.5-flash & 94/6 & 91/9 & 95/5 & 95/5 & 96/4 & 94/6 & 95/5 & 98/2 \\
Gemini gemini-2.5-pro & 92/8 & 95/5 & 95/5 & 95/5 & 98/2 & 95/5 & 97/3 & 98/2 \\
\bottomrule
\end{tabular}
\end{table}
\FloatBarrier

\subsection{Open-weight models on the Translate Task}
\begin{table}[ht!]
\tiny
\caption{Failure analysis of open-weight models on the Translation task. Judge question: was an ISO-8601 date predicted in the model output? LLM-as-Judge: Qwen3.5-35B (Ollama). Ground truth: the same regex used for the task's evaluation, given in Appendix~\ref{regexes}. $n=100$ per cell.}
\label{datetime_v2/translation_tasks_2_regex_n_vllm_1_datetime_v2/translation_tasks_2_regex_n_vllm_1}
\begin{tabular}{lllllllll}
\toprule
 & Direct-1 & Direct-2 & Direct-1 ISO & FS-1 & FS-2 & CoT & CoT FS-1 & CoT FS-2 \\
Model &  &  &  &  &  &  &  &  \\
\midrule
Llama 3.2 1B Greedy & 1.0 / 1.0 (56/44) & 1.0 / 1.0 (65/35) & 1.0 / 1.0 (48/52) & 1.0 / 1.0 (51/49) & 1.0 / 1.0 (68/32) & \unreliable{0.6 / 0.9 (34/66)} & 0.9 / 0.9 (54/46) & 0.9 / 0.9 (57/43) \\
Llama 3.2 3B Greedy & 1.0 / 1.0 (88/12) & 1.0 / 1.0 (80/20) & 1.0 / 1.0 (97/3) & 1.0 / 1.0 (99/1) & 1.0 / 0.8 (98/2) & 0.9 / 0.4 (78/22) & 1.0 / 0.9 (83/17) & 1.0 / 0.8 (90/10) \\
Gemma 3 1B Greedy & 1.0 / 0.9 (83/17) & 1.0 / 0.9 (92/8) & 1.0 / 0.6 (95/5) & 1.0 / 0.9 (96/4) & 1.0 / 1.0 (94/6) & 1.0 / 0.7 (90/10) & 1.0 / 0.7 (90/10) & 1.0 / 0.9 (90/10) \\
Gemma 3 4B Greedy & 1.0 / 0.9 (93/7) & 1.0 / 0.8 (90/10) & 1.0 / 0.0 (100/0) & 1.0 / 1.0 (99/1) & 1.0 / 0.0 (100/0) & 1.0 / 1.0 (98/2) & 1.0 / 1.0 (99/1) & 1.0 / 1.0 (99/1) \\
Phi 4 Mini Greedy & 1.0 / 0.0 (100/0) & 1.0 / 0.0 (100/0) & 1.0 / 0.0 (100/0) & 1.0 / 0.0 (100/0) & 1.0 / 0.8 (97/3) & 1.0 / 0.8 (98/2) & 1.0 / 1.0 (96/4) & 1.0 / 1.0 (95/5) \\
SmolLM2 1.7B Greedy & 1.0 / 0.9 (94/6) & 1.0 / 0.9 (93/7) & 1.0 / 0.9 (80/20) & 1.0 / 1.0 (81/19) & 1.0 / 0.9 (79/21) & 0.9 / 0.8 (81/19) & 1.0 / 1.0 (45/55) & 1.0 / 1.0 (33/67) \\
\bottomrule
\end{tabular}
\end{table}
\begin{table}[ht!]
\tiny
\caption{Failure analysis of open-weight models on the Translation task. Judge question: was an ISO-8601 datetime predicted in the model output? LLM-as-Judge: Qwen3.5-35B (Ollama). Ground truth: the same regex used for the task's evaluation, given in Appendix~\ref{regexes}. $n=100$ per cell.}
\label{datetime_v2/translation_tasks_2_regex_n_vllm_1_datetime_v2/translation_tasks_2_regex_n_vllm_1}
\begin{tabular}{lllllllll}
\toprule
 & Direct-1 & Direct-2 & Direct-1 ISO & FS-1 & FS-2 & CoT & CoT FS-1 & CoT FS-2 \\
Model &  &  &  &  &  &  &  &  \\
\midrule
Llama 3.2 1B Greedy & 1.0 / 0.9 (49/51) & 1.0 / 1.0 (62/38) & 1.0 / 1.0 (45/55) & 0.9 / 0.9 (44/56) & 0.9 / 0.9 (59/41) & \unreliable{0.6 / 0.9 (24/76)} & \unreliable{0.7 / 0.8 (35/65)} & \unreliable{0.8 / 0.8 (41/59)} \\
Llama 3.2 3B Greedy & 1.0 / 0.8 (81/19) & 1.0 / 0.9 (76/24) & 1.0 / 0.8 (95/5) & 1.0 / 0.3 (94/6) & 0.9 / 0.3 (89/11) & 0.9 / 0.4 (78/22) & 0.9 / 0.8 (76/24) & 1.0 / 0.7 (86/14) \\
Gemma 3 1B Greedy & 0.9 / 0.7 (74/26) & 1.0 / 0.7 (86/14) & 1.0 / 0.5 (94/6) & 1.0 / 0.7 (94/6) & 1.0 / 0.8 (90/10) & 0.9 / 0.5 (81/19) & \unreliable{0.9 / 0.5 (76/24)} & 1.0 / 0.7 (86/14) \\
Gemma 3 4B Greedy & 1.0 / 0.9 (91/9) & 1.0 / 0.8 (86/14) & 1.0 / 0.0 (99/1) & 1.0 / 1.0 (99/1) & 1.0 / 0.0 (100/0) & 1.0 / 1.0 (98/2) & 1.0 / 1.0 (99/1) & 1.0 / 1.0 (99/1) \\
Phi 4 Mini Greedy & 1.0 / 0.0 (96/4) & 1.0 / 0.0 (100/0) & 1.0 / 0.0 (99/1) & 1.0 / 0.0 (99/1) & 1.0 / 0.8 (97/3) & 1.0 / 1.0 (97/3) & 1.0 / 1.0 (96/4) & 1.0 / 1.0 (95/5) \\
SmolLM2 1.7B Greedy & 1.0 / 0.7 (91/9) & 1.0 / 0.7 (87/13) & 1.0 / 0.9 (76/24) & 1.0 / 0.9 (78/22) & 1.0 / 0.9 (77/23) & \unreliable{0.8 / 0.7 (56/44)} & \unreliable{0.7 / 0.8 (24/76)} & 1.0 / 1.0 (31/69) \\
\bottomrule
\end{tabular}
\end{table}
\begin{table}[ht!]
\tiny
\caption{Failure analysis of open-weight models on the Translation task. Judge question: was more than one ISO-8601 datetime predicted in the model output? LLM-as-Judge: Qwen3.5-35B (Ollama). Ground truth: the \emph{ISO~8601 capture} regex from Appendix~\ref{regexes}, applied to count matches in the output sequence. $n=100$ per cell.}
\label{datetime_v2/translation_tasks_2_regex_n_vllm_1_datetime_v2/translation_tasks_2_regex_n_vllm_1}
\begin{tabular}{lllllllll}
\toprule
 & Direct-1 & Direct-2 & Direct-1 ISO & FS-1 & FS-2 & CoT & CoT FS-1 & CoT FS-2 \\
Model &  &  &  &  &  &  &  &  \\
\midrule
Llama 3.2 1B Greedy & \insufficient{0.0 / 1.0 (1/99)} & \insufficient{0.0 / 1.0 (1/99)} & \insufficient{0.5 / 1.0 (3/97)} & \insufficient{0.0 / 0.9 (15/85)} & \insufficient{0.1 / 0.8 (34/66)} & \insufficient{0.0 / 1.0 (0/100)} & \insufficient{0.0 / 1.0 (5/95)} & \insufficient{0.5 / 1.0 (7/93)} \\
Llama 3.2 3B Greedy & \insufficient{0.0 / 1.0 (0/100)} & \insufficient{0.0 / 1.0 (0/100)} & \insufficient{0.0 / 1.0 (0/100)} & \insufficient{0.0 / 1.0 (0/100)} & \insufficient{0.0 / 1.0 (0/100)} & \insufficient{0.0 / 1.0 (2/98)} & \insufficient{0.0 / 1.0 (0/100)} & \insufficient{0.0 / 1.0 (0/100)} \\
Gemma 3 1B Greedy & \insufficient{0.0 / 1.0 (0/100)} & \insufficient{0.0 / 1.0 (1/99)} & \insufficient{0.0 / 1.0 (0/100)} & \insufficient{0.0 / 1.0 (0/100)} & \insufficient{0.0 / 1.0 (0/100)} & \insufficient{0.0 / 1.0 (0/100)} & \insufficient{0.0 / 1.0 (0/100)} & \insufficient{0.0 / 1.0 (0/100)} \\
Gemma 3 4B Greedy & \insufficient{0.0 / 1.0 (0/100)} & \insufficient{0.0 / 1.0 (0/100)} & \insufficient{0.0 / 1.0 (0/100)} & \insufficient{0.0 / 1.0 (0/100)} & \insufficient{0.0 / 1.0 (0/100)} & \insufficient{0.0 / 1.0 (0/100)} & \insufficient{0.0 / 1.0 (1/99)} & \insufficient{0.0 / 1.0 (0/100)} \\
Phi 4 Mini Greedy & \insufficient{0.0 / 1.0 (0/100)} & \insufficient{0.0 / 1.0 (0/100)} & \insufficient{0.0 / 1.0 (1/99)} & \insufficient{0.0 / 1.0 (2/98)} & \insufficient{0.0 / 1.0 (6/94)} & \insufficient{0.0 / 1.0 (0/100)} & \insufficient{0.0 / 1.0 (4/96)} & \insufficient{0.0 / 1.0 (7/93)} \\
SmolLM2 1.7B Greedy & \insufficient{0.0 / 1.0 (0/100)} & \insufficient{0.0 / 1.0 (0/100)} & \insufficient{0.0 / 1.0 (0/100)} & \insufficient{0.0 / 1.0 (0/100)} & \insufficient{0.0 / 1.0 (0/100)} & \insufficient{0.0 / 1.0 (0/100)} & \insufficient{0.0 / 1.0 (0/100)} & \insufficient{0.0 / 1.0 (0/100)} \\
\bottomrule
\end{tabular}
\end{table}
\input{Tables/translation_tasks_2_regex_n_vllm_1_scratchpad_1.tex}
\begin{table}[ht!]
\tiny
\caption{Failure analysis of open-weight models on the Translation task. Question: was the predicted ISO-8601 datetime within one day of the target? Evaluation is fully deterministic: the predicted datetime is extracted via the \emph{ISO~8601 capture} regex from Appendix~\ref{regexes} and compared to the target with a one-day tolerance. $n=100$ per cell.}
\label{datetime_v2/translation_tasks_2_regex_n_vllm_1_datetime_v2/translation_tasks_2_regex_n_vllm_1}
\begin{tabular}{lllllllll}
\toprule
 & Direct-1 & Direct-2 & Direct-1 ISO & FS-1 & FS-2 & CoT & CoT FS-1 & CoT FS-2 \\
Model &  &  &  &  &  &  &  &  \\
\midrule
Llama 3.2 1B Greedy & 4/96 & 0/100 & 3/97 & 4/96 & 6/94 & 3/97 & 8/92 & 23/77 \\
Llama 3.2 3B Greedy & 3/97 & 3/97 & 22/78 & 30/70 & 51/49 & 48/52 & 44/56 & 69/31 \\
Gemma 3 1B Greedy & 1/99 & 2/98 & 2/98 & 2/98 & 3/97 & 2/98 & 2/98 & 4/96 \\
Gemma 3 4B Greedy & 8/92 & 8/92 & 29/71 & 31/69 & 38/62 & 30/70 & 37/63 & 51/49 \\
Phi 4 Mini Greedy & 2/98 & 1/99 & 10/90 & 11/89 & 27/73 & 24/76 & 38/62 & 59/41 \\
SmolLM2 1.7B Greedy & 2/98 & 2/98 & 1/99 & 2/98 & 2/98 & 14/86 & 4/96 & 4/96 \\
Qwen3 0.6B Greedy & 15/85 & 7/93 & 23/77 & 23/77 & 28/72 & 38/62 & 31/69 & 45/55 \\
Qwen3 1.7B Greedy & 78/22 & 70/30 & 76/24 & 67/33 & 64/36 & 77/23 & 74/26 & 68/32 \\
Qwen3 4B Greedy & 85/15 & 87/13 & 85/15 & 82/18 & 80/20 & 81/19 & 74/26 & 74/26 \\
OLMo2 1B Greedy & 4/96 & 5/95 & 21/79 & 0/100 & 0/100 & 28/72 & 0/100 & 2/98 \\
Falcon3 3B Greedy & 2/98 & 2/98 & 8/92 & 15/85 & 26/74 & 11/89 & 16/84 & 38/62 \\
\bottomrule
\end{tabular}
\end{table}
\input{Tables/translation_tasks_2_regex_n_vllm_1_error_days_10.tex}
\input{Tables/translation_tasks_2_regex_n_vllm_1_error_days_100.tex}
\FloatBarrier

\subsection{Proprietary models on the Add-250 Task}
\input{Tables/computation_tasks_2_regex_n_prop_llm_iso_8601_predicted_1}
\input{Tables/computation_tasks_2_regex_n_prop_llm_iso_8601_predicted_2}
\input{Tables/computation_tasks_2_regex_n_prop_llm_iso_8601_multiple_predicted_1}
\input{Tables/computation_tasks_2_regex_n_prop_llm_scratchpad_1}
\input{Tables/computation_tasks_2_regex_n_prop_llm_error_days_1.tex}
\input{Tables/computation_tasks_2_regex_n_prop_llm_error_days_10.tex}
\input{Tables/computation_tasks_2_regex_n_prop_llm_error_days_100.tex}
\FloatBarrier

\subsection{Open-weight models on the Add-250 Task}
\input{Tables/computation_tasks_2_regex_250_vllm_1_iso_8601_predicted_1}
\input{Tables/computation_tasks_2_regex_250_vllm_1_iso_8601_predicted_2}
\input{Tables/computation_tasks_2_regex_250_vllm_1_iso_8601_multiple_predicted_1}
\input{Tables/computation_tasks_2_regex_250_vllm_1_scratchpad_1}
\input{Tables/computation_tasks_2_regex_250_vllm_1_error_days_1}
\input{Tables/computation_tasks_2_regex_250_vllm_1_error_days_10}
\input{Tables/computation_tasks_2_regex_250_vllm_1_error_days_100}
\FloatBarrier

\subsection{Event Planning Task}
\input{Tables/natural_context_tasks_1_prop_llm_scratchpad_2}
\input{Tables/datetime_v2_natural_context_tasks_1_vllm_1_scratchpad_2}
\FloatBarrier

\end{landscape}

  \clearpage

\section{Scaling Analysis}
  \label{scaling_law}
  
To investigate how accuracy scales with task range, a 4-bit quantized Qwen 2.5 0.5B model is LoRA fine-tuned on addition tasks of progressively wider range, from adding at most 10 days up to adding at most 10{,}000 days. Training set sizes range from 1{,}000 to 1{,}000{,}000 exemplars. All training sets are decontaminated against the evaluation benchmark, ensuring that accuracy reflects generalisation to unseen dates rather than memorisation. Each cell is a single run, single epoch, evaluated on 1{,}000 observations. Task range confounds two factors: the number of distinct operations to learn and the arithmetic complexity of larger offsets. To separate these effects, a second experiment holds the offset magnitude constant at 1{,}000 days while varying the range width (Table~\ref{datetime_add_scaling_law_2}).

The fine-tuned model results on ranges of operands are reported in the main paper in Table~\ref{datetime_add_scaling_law_1} and Table~\ref{datetime_add_scaling_law_2}.

The results reveal a clear trade-off between addition range and data requirement (Table~\ref{datetime_add_scaling_law_1}). Accuracy degrades sharply as the addition range widens, and learning a range of tasks is fundamentally harder than learning a single one. Extending this analysis to larger models, full fine-tuning, and additional task configurations is left to future work.

  \clearpage

\section{Reproducibility Details}
  \label{reproducibility}
  
This section contains reproducibility details for all tables in the main sections of the paper. Release information for each dataset is in Appendix~\ref{datasets}, details on the regex comparators are in Appendix~\ref{regexes} and details on the models are in Appendix~\ref{models}.

\subsection{Dataset Generation}
\paragraph{Benchmark dataset generation.}
Each benchmark is generated in 10 independent \emph{runs} with different random seeds. Error bars reported in the evaluation tables are computed across these runs, using $\pm 2$ standard errors of the mean (hereafter \emph{SEM}). The catalog of published datasets and representative samples can be found in Appendix~\ref{datasets}.

\paragraph{Fine-tuning dataset generation.}
\label{descriptives_ft}
Fine-tuning datasets are generated from the same generator but are constructed to have zero overlap with their respective benchmark. 
For the \emph{Translation} task, exclusion operates on the ISO~8601 representation of each input datetime (without microseconds or timezone), so that any variation in the representation of the same underlying value is discarded. Fr the \emph{Add-250} task, the exclusion key is the date component (year, month, and day) of the ISO~8601 representation since hours, minutes, and seconds are irrelevant to the arithmetic in the context of adding days; excluding on the full timestamp would allow samples with the same date but different times to pass the exclusion filter. For the \emph{Event Planning} task, the target labels (\textit{yes} and \textit{no}) are balanced by overgeneration followed by equal sampling from each label. The exclusions and balancing are implemented after the data generation step.

\subsection{Table~\ref{datetime_translation_prop}:  Translation Accuracy (Proprietary Models)}
\label{datetime_translation_prop_repro}
\subsubsection{Datasets \& Comparators}

Release information for each dataset is in Appendix~\ref{datasets}; details on the regex comparators are in Appendix~\ref{regexes}.

\begin{tabular}{lrll}
\textbf{Benchmark} & \textbf{N} & \textbf{Randomised} & \textbf{Comparator} \\
\hline
\texttt{datetime/a.3} & 100 & no & \texttt{iso8601-regex-n} \\
\end{tabular}

\subsubsection{Models}

Information and parameters for each model are in Appendix~\ref{models}. Where applicable, compute resources are detailed in Appendix~\ref{compute_resources}.

\begin{tabular}{ll}
\textbf{Model} & \textbf{Configuration} \\
\hline
\texttt{Anthropic claude-4.0-sonnet} & claude-sonnet-4-20250514 \\
\texttt{Anthropic claude-4.0-opus} & claude-opus-4-20250514 \\
\texttt{Anthropic claude-4.1-opus} & claude-opus-4-1-20250805 \\
\texttt{OpenAI gpt-3.5-turbo} & gpt-3.5-turbo-0125 \\
\texttt{OpenAI gpt-4.0-turbo} & gpt-4-turbo-2024-04-09 \\
\texttt{OpenAI gpt-4o-mini} & gpt-4o-mini-2024-07-18 \\
\texttt{OpenAI gpt-5} & gpt-5-2025-08-07 \\
\texttt{OpenAI gpt-5-mini} & gpt-5-mini-2025-08-07 \\
\texttt{OpenAI gpt-5-nano} & gpt-5-nano-2025-08-07 \\
\texttt{Gemini gemini-2.5-flash-lite} & gemini-2.5-flash-lite \\
\texttt{Gemini gemini-2.5-flash} & gemini-2.5-flash \\
\texttt{Gemini gemini-2.5-pro} & gemini-2.5-pro \\
\end{tabular}

\subsubsection{Prompts}

\subsubsection{Prompt: Direct-1}

\textbf{System prompt:}\\
{\ttfamily\small Never provide a preamble like 'sure', 'of course', 'here is the answer'. Answer the question directly. Do not provide an explanation.}

\textbf{Prompt template:}\\
{\ttfamily\small Here is a datetime : 

"\{input\_sequence\}"

. Translate the datetime to ISO-8601 format.}

\textbf{Rendered example:}\\
{\ttfamily\small Here is a datetime : 

"Feb/Sun 26th,7111 ,1 AM 45"

. Translate the datetime to ISO-8601 format.}

\textbf{Target:} \texttt{7111-02-26T01:45:00}

\subsubsection{Prompt: Direct-2}

\textbf{System prompt:}\\
{\ttfamily\small Never provide a preamble like 'sure', 'of course', 'here is the answer'. Answer the question directly. Do not provide an explanation.}

\textbf{Prompt template:}\\
{\ttfamily\small Convert the following datetime to ISO-8601 format: "\{input\_sequence\}"}

\textbf{Rendered example:}\\
{\ttfamily\small Convert the following datetime to ISO-8601 format: "Feb/Sun 26th,7111 ,1 AM 45"}

\textbf{Target:} \texttt{7111-02-26T01:45:00}

\subsubsection{Prompt: Direct-1 ISO}

\textbf{System prompt:}\\
{\ttfamily\small Never provide a preamble like 'sure', 'of course', 'here is the answer'. Answer the question directly. Do not provide an explanation.}

\textbf{Prompt template:}\\
{\ttfamily\small Here is a datetime : 

"\{input\_sequence\}"

. Translate the datetime to ISO-8601 format. ISO-8601 is the international standard for representing dates and times. A DateTime must have the format 'YYYY-MM-DDTHH:MM:SS' (e.g., '2023-03-15T14:30:25')}

\textbf{Rendered example:}\\
{\ttfamily\small Here is a datetime : 

"Feb/Sun 26th,7111 ,1 AM 45"

. Translate the datetime to ISO-8601 format. ISO-8601 is the international standard for representing dates and times. A DateTime must have the format 'YYYY-MM-DDTHH:MM:SS' (e.g., '2023-03-15T14:30:25')}

\textbf{Target:} \texttt{7111-02-26T01:45:00}

\subsubsection{Prompt: FS-1}

\textbf{System prompt:}\\
{\ttfamily\small Never provide a preamble like 'sure', 'of course', 'here is the answer'. Answer the question directly. Do not provide an explanation.}

\textbf{Prompt template:}\\
{\ttfamily\small Here is a datetime : 

"\{input\_sequence\}"

. Translate the datetime to ISO-8601 format. For example, the datetime '13 pm 58:5 thu 19|2|1999' should be translated to '1999-02-19T13:58:05'. ISO-8601 is the international standard for representing dates and times. A datetime must have the format 'YYYY-MM-DDTHH:MM:SS' (e.g., 2023-03-15T14:30:25)}

\textbf{Rendered example:}\\
{\ttfamily\small Here is a datetime : 

"Feb/Sun 26th,7111 ,1 AM 45"

. Translate the datetime to ISO-8601 format. For example, the datetime '13 pm 58:5 thu 19|2|1999' should be translated to '1999-02-19T13:58:05'. ISO-8601 is the international standard for representing dates and times. A datetime must have the format 'YYYY-MM-DDTHH:MM:SS' (e.g., 2023-03-15T14:30:25)}

\textbf{Target:} \texttt{7111-02-26T01:45:00}

\subsubsection{Prompt: FS-2}

\textbf{System prompt:}\\
{\ttfamily\small Never provide a preamble like 'sure', 'of course', 'here is the answer'. Answer the question directly. Do not provide an explanation.}

\textbf{Prompt template:}\\
{\ttfamily\small Here is a datetime : 

"\{input\_sequence\}"

. Translate the datetime to ISO-8601 format. For example, the datetime '13 pm 58:5 thu 19|2|1999' should be translated to '1999-02-19T13:58:05'. If we take another example, the datetime '5:48:05 pm mon 09-9-6480' should be written as '6480-09-09T17:48:05'. ISO-8601 is the international standard for representing dates and times. A datetime must have the format 'YYYY-MM-DDTHH:MM:SS' (e.g., 2023-03-15T14:30:25)}

\textbf{Rendered example:}\\
{\ttfamily\small Here is a datetime : 

"Feb/Sun 26th,7111 ,1 AM 45"

. Translate the datetime to ISO-8601 format. For example, the datetime '13 pm 58:5 thu 19|2|1999' should be translated to '1999-02-19T13:58:05'. If we take another example, the datetime '5:48:05 pm mon 09-9-6480' should be written as '6480-09-09T17:48:05'. ISO-8601 is the international standard for representing dates and times. A datetime must have the format 'YYYY-MM-DDTHH:MM:SS' (e.g., 2023-03-15T14:30:25)}

\textbf{Target:} \texttt{7111-02-26T01:45:00}

\subsubsection{Prompt: CoT}

\textbf{System prompt:}\\
{\ttfamily\small Never provide a preamble like 'sure', 'of course', 'here is the answer'. Answer the question directly. Do not provide an explanation.}

\textbf{Prompt template:}\\
{\ttfamily\small Here is a datetime : 

"\{input\_sequence\}"

. Translate the datetime to ISO-8601 format. Think step by step. First identify the year, the month, the day, the hours, the minutes and then the seconds. Finally you can construct an ISO-8601 string in the format '\{year\}-\{month\}-\{day\}T\{hour\}:\{minute\}:\{second\}'. ISO-8601 is the international standard for representing dates and times. A datetime must have the format 'YYYY-MM-DDTHH:MM:SS' (e.g., 2023-03-15T14:30:25)}

\textbf{Rendered example:}\\
{\ttfamily\small Here is a datetime : 

"Feb/Sun 26th,7111 ,1 AM 45"

. Translate the datetime to ISO-8601 format. Think step by step. First identify the year, the month, the day, the hours, the minutes and then the seconds. Finally you can construct an ISO-8601 string in the format '\{year\}-\{month\}-\{day\}T\{hour\}:\{minute\}:\{second\}'. ISO-8601 is the international standard for representing dates and times. A datetime must have the format 'YYYY-MM-DDTHH:MM:SS' (e.g., 2023-03-15T14:30:25)}

\textbf{Target:} \texttt{7111-02-26T01:45:00}

\subsubsection{Prompt: CoT FS-1}

\textbf{System prompt:}\\
{\ttfamily\small Never provide a preamble like 'sure', 'of course', 'here is the answer'. Answer the question directly. Do not provide an explanation.}

\textbf{Prompt template:}\\
{\ttfamily\small Here is a datetime : 

"\{input\_sequence\}"

. Translate the datetime to ISO-8601. For example, let's take the datetime '13 pm 58:5 thu 19|2|1999'. Think step by step. First identify the year, then the month, the day, the hours, the minutes and the seconds. Finally you can construct an ISO-8601 string in the format '\{year\}-\{month\}-\{day\}T\{hour\}:\{minute\}:\{second\}'. So the answer in this example would be '1999-02-19T13:58:05'. ISO-8601 is the international standard for representing dates and times. A datetime must have the format 'YYYY-MM-DDTHH:MM:SS' (e.g., 2023-03-15T14:30:25)}

\textbf{Rendered example:}\\
{\ttfamily\small Here is a datetime : 

"Feb/Sun 26th,7111 ,1 AM 45"

. Translate the datetime to ISO-8601. For example, let's take the datetime '13 pm 58:5 thu 19|2|1999'. Think step by step. First identify the year, then the month, the day, the hours, the minutes and the seconds. Finally you can construct an ISO-8601 string in the format '\{year\}-\{month\}-\{day\}T\{hour\}:\{minute\}:\{second\}'. So the answer in this example would be '1999-02-19T13:58:05'. ISO-8601 is the international standard for representing dates and times. A datetime must have the format 'YYYY-MM-DDTHH:MM:SS' (e.g., 2023-03-15T14:30:25)}

\textbf{Target:} \texttt{7111-02-26T01:45:00}

\subsubsection{Prompt: CoT FS-2}

\textbf{System prompt:}\\
{\ttfamily\small Never provide a preamble like 'sure', 'of course', 'here is the answer'. Answer the question directly. Do not provide an explanation.}

\textbf{Prompt template:}\\
{\ttfamily\small Here is a datetime : 

"\{input\_sequence\}"

. Translate the datetime to ISO-8601 format. For example, let's take the datetime '13 pm 58:5 thu 19|2|1999'. Think step by step. First identify the year, then the month, the day, the hours, the minutes and the seconds. Finally you can construct an ISO-8601 string in the format '\{year\}-\{month\}-\{day\}T\{hour\}:\{minute\}:\{second\}'. So the answer in this example would be '1999-02-19T13:58:05'. If we take another example, the datetime '5:48:05 pm mon 09-9-6480' should be written as '6480-09-09T17:48:05'. ISO-8601 is the international standard for representing dates and times. A datetime must have the format 'YYYY-MM-DDTHH:MM:SS' (e.g., 2023-03-15T14:30:25)}

\textbf{Rendered example:}\\
{\ttfamily\small Here is a datetime : 

"Feb/Sun 26th,7111 ,1 AM 45"

. Translate the datetime to ISO-8601 format. For example, let's take the datetime '13 pm 58:5 thu 19|2|1999'. Think step by step. First identify the year, then the month, the day, the hours, the minutes and the seconds. Finally you can construct an ISO-8601 string in the format '\{year\}-\{month\}-\{day\}T\{hour\}:\{minute\}:\{second\}'. So the answer in this example would be '1999-02-19T13:58:05'. If we take another example, the datetime '5:48:05 pm mon 09-9-6480' should be written as '6480-09-09T17:48:05'. ISO-8601 is the international standard for representing dates and times. A datetime must have the format 'YYYY-MM-DDTHH:MM:SS' (e.g., 2023-03-15T14:30:25)}

\textbf{Target:} \texttt{7111-02-26T01:45:00}

\subsection{Table~\ref{datetime_translation_vllm}:  Translation Accuracy (Open-weight Models)}
\label{datetime_translation_vllm_repro}
\subsubsection{Datasets \& Comparators}

Release information for each dataset is in Appendix~\ref{datasets}; details on the regex comparators are in Appendix~\ref{regexes}.

\begin{tabular}{lrll}
\textbf{Benchmark} & \textbf{N} & \textbf{Randomised} & \textbf{Comparator} \\
\hline
\texttt{datetime/a.3\_1} & 100 & no & \texttt{iso8601-regex-n} \\
\end{tabular}

\subsubsection{Models}

Information and parameters for each model are in Appendix~\ref{models}. Where applicable, compute resources are detailed in Appendix~\ref{compute_resources}.

\begin{tabular}{ll}
\textbf{Model} & \textbf{Configuration} \\
\hline
\texttt{Llama 3.2 1B Greedy} & meta-llama/Llama-3.2-1B-Instruct \\
\texttt{Llama 3.2 3B Greedy} & meta-llama/Llama-3.2-3B-Instruct \\
\texttt{Gemma 3 1B Greedy} & google/gemma-3-1b-it \\
\texttt{Gemma 3 4B Greedy} & google/gemma-3-4b-it \\
\texttt{Phi 4 Mini Greedy} & microsoft/Phi-4-mini-instruct \\
\texttt{SmolLM2 1.7B Greedy} & HuggingFaceTB/SmolLM2-1.7B-Instruct \\
\texttt{Qwen3 0.6B Greedy} & Qwen/Qwen3-0.6B \\
\texttt{Qwen3 1.7B Greedy} & Qwen/Qwen3-1.7B \\
\texttt{Qwen3 4B Greedy} & Qwen/Qwen3-4B \\
\texttt{OLMo2 1B Greedy} & allenai/OLMo-2-0425-1B-Instruct \\
\texttt{Falcon3 3B Greedy} & tiiuae/Falcon3-3B-Instruct \\
\end{tabular}

\subsubsection{Prompts}

\subsubsection{Prompt: Direct-1}

\textbf{System prompt:}\\
{\ttfamily\small Never provide a preamble like 'sure', 'of course', 'here is the answer'. Answer the question directly. Do not provide an explanation.}

\textbf{Prompt template:}\\
{\ttfamily\small Here is a datetime : 

"\{input\_sequence\}"

. Translate the datetime to ISO-8601 format.}

\textbf{Rendered example:}\\
{\ttfamily\small Here is a datetime : 

"Feb/Sun 26th,7111 ,1 AM 45"

. Translate the datetime to ISO-8601 format.}

\textbf{Target:} \texttt{7111-02-26T01:45:00}

\subsubsection{Prompt: Direct-2}

\textbf{System prompt:}\\
{\ttfamily\small Never provide a preamble like 'sure', 'of course', 'here is the answer'. Answer the question directly. Do not provide an explanation.}

\textbf{Prompt template:}\\
{\ttfamily\small Convert the following datetime to ISO-8601 format: "\{input\_sequence\}"}

\textbf{Rendered example:}\\
{\ttfamily\small Convert the following datetime to ISO-8601 format: "Feb/Sun 26th,7111 ,1 AM 45"}

\textbf{Target:} \texttt{7111-02-26T01:45:00}

\subsubsection{Prompt: Direct-1 ISO}

\textbf{System prompt:}\\
{\ttfamily\small Never provide a preamble like 'sure', 'of course', 'here is the answer'. Answer the question directly. Do not provide an explanation.}

\textbf{Prompt template:}\\
{\ttfamily\small Here is a datetime : 

"\{input\_sequence\}"

. Translate the datetime to ISO-8601 format. ISO-8601 is the international standard for representing dates and times. A DateTime must have the format 'YYYY-MM-DDTHH:MM:SS' (e.g., '2023-03-15T14:30:25')}

\textbf{Rendered example:}\\
{\ttfamily\small Here is a datetime : 

"Feb/Sun 26th,7111 ,1 AM 45"

. Translate the datetime to ISO-8601 format. ISO-8601 is the international standard for representing dates and times. A DateTime must have the format 'YYYY-MM-DDTHH:MM:SS' (e.g., '2023-03-15T14:30:25')}

\textbf{Target:} \texttt{7111-02-26T01:45:00}

\subsubsection{Prompt: FS-1}

\textbf{System prompt:}\\
{\ttfamily\small Never provide a preamble like 'sure', 'of course', 'here is the answer'. Answer the question directly. Do not provide an explanation.}

\textbf{Prompt template:}\\
{\ttfamily\small Here is a datetime : 

"\{input\_sequence\}"

. Translate the datetime to ISO-8601 format. For example, the datetime '13 pm 58:5 thu 19|2|1999' should be translated to '1999-02-19T13:58:05'. ISO-8601 is the international standard for representing dates and times. A datetime must have the format 'YYYY-MM-DDTHH:MM:SS' (e.g., 2023-03-15T14:30:25)}

\textbf{Rendered example:}\\
{\ttfamily\small Here is a datetime : 

"Feb/Sun 26th,7111 ,1 AM 45"

. Translate the datetime to ISO-8601 format. For example, the datetime '13 pm 58:5 thu 19|2|1999' should be translated to '1999-02-19T13:58:05'. ISO-8601 is the international standard for representing dates and times. A datetime must have the format 'YYYY-MM-DDTHH:MM:SS' (e.g., 2023-03-15T14:30:25)}

\textbf{Target:} \texttt{7111-02-26T01:45:00}

\subsubsection{Prompt: FS-2}

\textbf{System prompt:}\\
{\ttfamily\small Never provide a preamble like 'sure', 'of course', 'here is the answer'. Answer the question directly. Do not provide an explanation.}

\textbf{Prompt template:}\\
{\ttfamily\small Here is a datetime : 

"\{input\_sequence\}"

. Translate the datetime to ISO-8601 format. For example, the datetime '13 pm 58:5 thu 19|2|1999' should be translated to '1999-02-19T13:58:05'. If we take another example, the datetime '5:48:05 pm mon 09-9-6480' should be written as '6480-09-09T17:48:05'. ISO-8601 is the international standard for representing dates and times. A datetime must have the format 'YYYY-MM-DDTHH:MM:SS' (e.g., 2023-03-15T14:30:25)}

\textbf{Rendered example:}\\
{\ttfamily\small Here is a datetime : 

"Feb/Sun 26th,7111 ,1 AM 45"

. Translate the datetime to ISO-8601 format. For example, the datetime '13 pm 58:5 thu 19|2|1999' should be translated to '1999-02-19T13:58:05'. If we take another example, the datetime '5:48:05 pm mon 09-9-6480' should be written as '6480-09-09T17:48:05'. ISO-8601 is the international standard for representing dates and times. A datetime must have the format 'YYYY-MM-DDTHH:MM:SS' (e.g., 2023-03-15T14:30:25)}

\textbf{Target:} \texttt{7111-02-26T01:45:00}

\subsubsection{Prompt: CoT}

\textbf{System prompt:}\\
{\ttfamily\small Never provide a preamble like 'sure', 'of course', 'here is the answer'. Answer the question directly. Do not provide an explanation.}

\textbf{Prompt template:}\\
{\ttfamily\small Here is a datetime : 

"\{input\_sequence\}"

. Translate the datetime to ISO-8601 format. Think step by step. First identify the year, the month, the day, the hours, the minutes and then the seconds. Finally you can construct an ISO-8601 string in the format '\{year\}-\{month\}-\{day\}T\{hour\}:\{minute\}:\{second\}'. ISO-8601 is the international standard for representing dates and times. A datetime must have the format 'YYYY-MM-DDTHH:MM:SS' (e.g., 2023-03-15T14:30:25)}

\textbf{Rendered example:}\\
{\ttfamily\small Here is a datetime : 

"Feb/Sun 26th,7111 ,1 AM 45"

. Translate the datetime to ISO-8601 format. Think step by step. First identify the year, the month, the day, the hours, the minutes and then the seconds. Finally you can construct an ISO-8601 string in the format '\{year\}-\{month\}-\{day\}T\{hour\}:\{minute\}:\{second\}'. ISO-8601 is the international standard for representing dates and times. A datetime must have the format 'YYYY-MM-DDTHH:MM:SS' (e.g., 2023-03-15T14:30:25)}

\textbf{Target:} \texttt{7111-02-26T01:45:00}

\subsubsection{Prompt: CoT FS-1}

\textbf{System prompt:}\\
{\ttfamily\small Never provide a preamble like 'sure', 'of course', 'here is the answer'. Answer the question directly. Do not provide an explanation.}

\textbf{Prompt template:}\\
{\ttfamily\small Here is a datetime : 

"\{input\_sequence\}"

. Translate the datetime to ISO-8601. For example, let's take the datetime '13 pm 58:5 thu 19|2|1999'. Think step by step. First identify the year, then the month, the day, the hours, the minutes and the seconds. Finally you can construct an ISO-8601 string in the format '\{year\}-\{month\}-\{day\}T\{hour\}:\{minute\}:\{second\}'. So the answer in this example would be '1999-02-19T13:58:05'. ISO-8601 is the international standard for representing dates and times. A datetime must have the format 'YYYY-MM-DDTHH:MM:SS' (e.g., 2023-03-15T14:30:25)}

\textbf{Rendered example:}\\
{\ttfamily\small Here is a datetime : 

"Feb/Sun 26th,7111 ,1 AM 45"

. Translate the datetime to ISO-8601. For example, let's take the datetime '13 pm 58:5 thu 19|2|1999'. Think step by step. First identify the year, then the month, the day, the hours, the minutes and the seconds. Finally you can construct an ISO-8601 string in the format '\{year\}-\{month\}-\{day\}T\{hour\}:\{minute\}:\{second\}'. So the answer in this example would be '1999-02-19T13:58:05'. ISO-8601 is the international standard for representing dates and times. A datetime must have the format 'YYYY-MM-DDTHH:MM:SS' (e.g., 2023-03-15T14:30:25)}

\textbf{Target:} \texttt{7111-02-26T01:45:00}

\subsubsection{Prompt: CoT FS-2}

\textbf{System prompt:}\\
{\ttfamily\small Never provide a preamble like 'sure', 'of course', 'here is the answer'. Answer the question directly. Do not provide an explanation.}

\textbf{Prompt template:}\\
{\ttfamily\small Here is a datetime : 

"\{input\_sequence\}"

. Translate the datetime to ISO-8601 format. For example, let's take the datetime '13 pm 58:5 thu 19|2|1999'. Think step by step. First identify the year, then the month, the day, the hours, the minutes and the seconds. Finally you can construct an ISO-8601 string in the format '\{year\}-\{month\}-\{day\}T\{hour\}:\{minute\}:\{second\}'. So the answer in this example would be '1999-02-19T13:58:05'. If we take another example, the datetime '5:48:05 pm mon 09-9-6480' should be written as '6480-09-09T17:48:05'. ISO-8601 is the international standard for representing dates and times. A datetime must have the format 'YYYY-MM-DDTHH:MM:SS' (e.g., 2023-03-15T14:30:25)}

\textbf{Rendered example:}\\
{\ttfamily\small Here is a datetime : 

"Feb/Sun 26th,7111 ,1 AM 45"

. Translate the datetime to ISO-8601 format. For example, let's take the datetime '13 pm 58:5 thu 19|2|1999'. Think step by step. First identify the year, then the month, the day, the hours, the minutes and the seconds. Finally you can construct an ISO-8601 string in the format '\{year\}-\{month\}-\{day\}T\{hour\}:\{minute\}:\{second\}'. So the answer in this example would be '1999-02-19T13:58:05'. If we take another example, the datetime '5:48:05 pm mon 09-9-6480' should be written as '6480-09-09T17:48:05'. ISO-8601 is the international standard for representing dates and times. A datetime must have the format 'YYYY-MM-DDTHH:MM:SS' (e.g., 2023-03-15T14:30:25)}

\textbf{Target:} \texttt{7111-02-26T01:45:00}

\subsection{Table~\ref{datetime_translation_unsloth}:  Translation Accuracy (Fine-Tuned Models)}
\label{datetime_translation_unsloth_repro}
\subsubsection{Datasets \& Comparators}

Release information for each dataset is in Appendix~\ref{datasets}; details on the regex comparators are in Appendix~\ref{regexes}.

\begin{tabular}{lrll}
\textbf{Benchmark} & \textbf{N} & \textbf{Randomised} & \textbf{Comparator} \\
\hline
\texttt{datetime/a.3} & 100 & no & \texttt{iso8601-regex-n} \\
\end{tabular}

Fine-tuning dataset is \texttt{datetime/a.3.ft\_10m}, with no overlapping input sequences with the benchmark; see Section~\ref{descriptives_ft}.

\subsubsection{Models}

Information and parameters for each model are in Appendix~\ref{models}. Where applicable, compute resources are detailed in Appendix~\ref{compute_resources}.

\begin{tabular}{ll}
\textbf{Model} & \textbf{Configuration} \\
\hline
\texttt{unsloth/Qwen2.5-0.5B} & unsloth/Qwen2.5-0.5B \\
\texttt{unsloth/Qwen2.5-1.5B} & unsloth/Qwen2.5-1.5B \\
\texttt{unsloth/Qwen2.5-1.5B Instruct} & unsloth/Qwen2.5-1.5B-Instruct \\
\texttt{unsloth/Qwen2.5-3B} & unsloth/Qwen2.5-3B \\
\texttt{unsloth/Qwen2.5-3B Instruct} & unsloth/Qwen2.5-3B-Instruct \\
\texttt{unsloth/Qwen2.5-7B Instruct} & unsloth/Qwen2.5-7B-Instruct \\
\texttt{unsloth/Qwen2.5-14B Instruct} & unsloth/Qwen2.5-14B-Instruct \\
\end{tabular}

\subsubsection{Prompts}

A single prompt template is used for both fine-tuning and inference; the limitation of evaluating fine-tuned models under a single prompt is discussed in Section~\ref{limitations}.

\textbf{System prompt:}\\
{\ttfamily\small You are a datetime processor}

\textbf{Prompt template:}\\
{\ttfamily\small translate to ISO-8601 : \{input\_sequence\}}

\textbf{Rendered example:}\\
{\ttfamily\small translate to ISO-8601 : Feb/Sun 26th,7111 ,1 AM 45}

\textbf{Target:} \texttt{7111-02-26T01:45:00}

\subsection{Table~\ref{datetime_computation_add_250_prop}:  Add-250 Accuracy (Proprietary Models)}
\label{datetime_computation_add_250_prop_repro}
\subsubsection{Datasets \& Comparators}

Release information for each dataset is in Appendix~\ref{datasets}; details on the regex comparators are in Appendix~\ref{regexes}.

\begin{tabular}{lrll}
\textbf{Benchmark} & \textbf{N} & \textbf{Randomised} & \textbf{Comparator} \\
\hline
\texttt{datetime/a3.iso8601.add.day.250.x} & 100 & no & \texttt{iso8601-regex-n} \\
\end{tabular}

\subsubsection{Models}

Information and parameters for each model are in Appendix~\ref{models}. Where applicable, compute resources are detailed in Appendix~\ref{compute_resources}.

\begin{tabular}{ll}
\textbf{Model} & \textbf{Configuration} \\
\hline
\texttt{OpenAI gpt-3.5-turbo} & gpt-3.5-turbo-0125 \\
\texttt{OpenAI gpt-4.0-turbo} & gpt-4-turbo-2024-04-09 \\
\texttt{OpenAI gpt-4o-mini} & gpt-4o-mini-2024-07-18 \\
\texttt{OpenAI gpt-5} & gpt-5-2025-08-07 \\
\texttt{OpenAI gpt-5-mini} & gpt-5-mini-2025-08-07 \\
\texttt{OpenAI gpt-5-nano} & gpt-5-nano-2025-08-07 \\
\texttt{Gemini 2.5-flash-lite} & gemini-2.5-flash-lite \\
\texttt{Gemini 2.5-flash} & gemini-2.5-flash \\
\texttt{Gemini 2.5-pro} & gemini-2.5-pro \\
\texttt{Gemini 3-flash-preview} & gemini-3-flash-preview \\
\texttt{Anthropic claude-4.0-sonnet} & claude-sonnet-4-20250514 \\
\texttt{Anthropic claude-4.0-opus} & claude-opus-4-20250514 \\
\texttt{Anthropic claude-4.1-opus} & claude-opus-4-1-20250805 \\
\end{tabular}

\subsubsection{Prompts}

\subsubsubsection{Prompt:Direct-1}

\textbf{System prompt:}\\
{\ttfamily\small Never provide a preamble like 'sure', 'of course', 'here is the answer'. Answer the question directly. Do not provide an explanation.}

\textbf{Prompt template:}\\
{\ttfamily\small Here is a datetime in ISO-8601 format : 

"\{input\_sequence\}"

. Add \{num\_days\} days to the datetime and generate a new datetime in ISO-8601 format.}

\textbf{Rendered example:}\\
{\ttfamily\small Here is a datetime in ISO-8601 format : 

"1241-07-08T22:56:42"

. Add 250 days to the datetime and generate a new datetime in ISO-8601 format.}

\textbf{Target:} \texttt{1242-03-15T22:56:42}

\subsubsubsection{Prompt:Direct-2}

\textbf{System prompt:}\\
{\ttfamily\small Never provide a preamble like 'sure', 'of course', 'here is the answer'. Answer the question directly. Do not provide an explanation.}

\textbf{Prompt template:}\\
{\ttfamily\small Add \{num\_days\} days to the following datetime "\{input\_sequence\}". The answer must be in ISO-8601 format.}

\textbf{Rendered example:}\\
{\ttfamily\small Add 250 days to the following datetime "1241-07-08T22:56:42". The answer must be in ISO-8601 format.}

\textbf{Target:} \texttt{1242-03-15T22:56:42}

\subsubsubsection{Prompt:Direct-1 ISO}

\textbf{System prompt:}\\
{\ttfamily\small Never provide a preamble like 'sure', 'of course', 'here is the answer'. Answer the question directly. Do not provide an explanation.}

\textbf{Prompt template:}\\
{\ttfamily\small Here is a datetime in ISO-8601 format : 

"\{input\_sequence\}"

. Add \{num\_days\} days to the datetime and generate a new datetime in ISO-8601 format. ISO-8601 is the international standard for representing dates and times. A datetime must have the format 'YYYY-MM-DDTHH:MM:SS' (e.g., '2023-03-15T14:30:25')}

\textbf{Rendered example:}\\
{\ttfamily\small Here is a datetime in ISO-8601 format : 

"1241-07-08T22:56:42"

. Add 250 days to the datetime and generate a new datetime in ISO-8601 format. ISO-8601 is the international standard for representing dates and times. A datetime must have the format 'YYYY-MM-DDTHH:MM:SS' (e.g., '2023-03-15T14:30:25')}

\textbf{Target:} \texttt{1242-03-15T22:56:42}

\subsubsubsection{Prompt:Direct-2 ISO}

\textbf{System prompt:}\\
{\ttfamily\small Never provide a preamble like 'sure', 'of course', 'here is the answer'. Answer the question directly. Do not provide an explanation.}

\textbf{Prompt template:}\\
{\ttfamily\small Add \{num\_days\} days to the following datetime "\{input\_sequence\}". The answer must be in ISO-8601 format. ISO-8601 is the international standard for representing dates and times. A datetime must have the format 'YYYY-MM-DDTHH:MM:SS' (e.g., '2023-03-15T14:30:25')}

\textbf{Rendered example:}\\
{\ttfamily\small Add 250 days to the following datetime "1241-07-08T22:56:42". The answer must be in ISO-8601 format. ISO-8601 is the international standard for representing dates and times. A datetime must have the format 'YYYY-MM-DDTHH:MM:SS' (e.g., '2023-03-15T14:30:25')}

\textbf{Target:} \texttt{1242-03-15T22:56:42}

\subsubsubsection{Prompt:FS-1}

\textbf{System prompt:}\\
{\ttfamily\small Never provide a preamble like 'sure', 'of course', 'here is the answer'. Answer the question directly. Do not provide an explanation.}

\textbf{Prompt template:}\\
{\ttfamily\small Here is a datetime in ISO-8601 format : 

"\{input\_sequence\}"

. Add \{num\_days\} days to the datetime and generate a new datetime in ISO-8601 format. For example, if the input datetime is '\{input\_fs\_1\}', then your answer should be '\{target\_fs\_1\}'}

\textbf{Rendered example:}\\
{\ttfamily\small Here is a datetime in ISO-8601 format : 

"1241-07-08T22:56:42"

. Add 250 days to the datetime and generate a new datetime in ISO-8601 format. For example, if the input datetime is '4047-11-01T12:35:06', then your answer should be '4048-07-08T12:35:06'}

\textbf{Target:} \texttt{1242-03-15T22:56:42}

\subsubsubsection{Prompt:FS-2}

\textbf{System prompt:}\\
{\ttfamily\small Never provide a preamble like 'sure', 'of course', 'here is the answer'. Answer the question directly. Do not provide an explanation.}

\textbf{Prompt template:}\\
{\ttfamily\small Here is a datetime in ISO-8601 format : 

"\{input\_sequence\}"

. Add \{num\_days\} days to the datetime and generate a new datetime in ISO-8601 format. For example, if the input datetime is '\{input\_fs\_1\}', then your answer should be '\{target\_fs\_1\}'. As another example, if the input datetime is '\{input\_fs\_2\}', then your answer should be '\{target\_fs\_2\}'}

\textbf{Rendered example:}\\
{\ttfamily\small Here is a datetime in ISO-8601 format : 

"1241-07-08T22:56:42"

. Add 250 days to the datetime and generate a new datetime in ISO-8601 format. For example, if the input datetime is '4047-11-01T12:35:06', then your answer should be '4048-07-08T12:35:06'. As another example, if the input datetime is '8897-06-25T17:00:20', then your answer should be '8898-03-02T17:00:20'}

\textbf{Target:} \texttt{1242-03-15T22:56:42}

\subsubsubsection{Prompt:CoT-1}

\textbf{System prompt:}\\
{\ttfamily\small Never provide a preamble like 'sure', 'of course', 'here is the answer'. Answer the question directly. Do not provide an explanation.}

\textbf{Prompt template:}\\
{\ttfamily\small Here is a datetime in ISO-8601 format : 

"\{input\_sequence\}"

. Add \{num\_days\} days to the datetime and generate a new datetime in ISO-8601 format. Let's think step-by-step. Parse the input date components (year, month, day). Add \{num\_days\} days, accounting for the number of days in each month and leap years.
.}

\textbf{Rendered example:}\\
{\ttfamily\small Here is a datetime in ISO-8601 format : 

"1241-07-08T22:56:42"

. Add 250 days to the datetime and generate a new datetime in ISO-8601 format. Let's think step-by-step. Parse the input date components (year, month, day). Add 250 days, accounting for the number of days in each month and leap years.
.}

\textbf{Target:} \texttt{1242-03-15T22:56:42}

\subsubsubsection{Prompt:CoT-1 + FS-1}

\textbf{System prompt:}\\
{\ttfamily\small Never provide a preamble like 'sure', 'of course', 'here is the answer'. Answer the question directly. Do not provide an explanation.}

\textbf{Prompt template:}\\
{\ttfamily\small Here is a datetime in ISO-8601 format : 

"\{input\_sequence\}"

. Add \{num\_days\} days to the datetime and generate a new datetime in ISO-8601 format. Let's think step-by-step. Parse the input date components (year, month, day). Add \{num\_days\} days, accounting for the number of days in each month and leap years. For example, if the datetime is '4939-01-31T12:30:10' and you need to add 250 days, your answer should be '4939-10-08T12:30:10'}

\textbf{Rendered example:}\\
{\ttfamily\small Here is a datetime in ISO-8601 format : 

"1241-07-08T22:56:42"

. Add 250 days to the datetime and generate a new datetime in ISO-8601 format. Let's think step-by-step. Parse the input date components (year, month, day). Add 250 days, accounting for the number of days in each month and leap years. For example, if the datetime is '4939-01-31T12:30:10' and you need to add 250 days, your answer should be '4939-10-08T12:30:10'}

\textbf{Target:} \texttt{1242-03-15T22:56:42}

\subsubsubsection{Prompt:CoT-1 + FS-2}

\textbf{System prompt:}\\
{\ttfamily\small Never provide a preamble like 'sure', 'of course', 'here is the answer'. Answer the question directly. Do not provide an explanation.}

\textbf{Prompt template:}\\
{\ttfamily\small Here is a datetime in ISO-8601 format : 

"\{input\_sequence\}"

. Add \{num\_days\} days to the datetime and generate a new datetime in ISO-8601 format. Let's think step-by-step. Parse the input date components (year, month, day). Add \{num\_days\} days, accounting for the number of days in each month and leap years. For example, if the datetime is '4939-01-31T12:30:10' and you need to add 250 days, your answer should be '4939-10-08T12:30:10'. As another example, if the datetime is '6553-05-22T03:47:11' and you also need to add 250 days, your answer should be '6554-01-27T03:47:11'}

\textbf{Rendered example:}\\
{\ttfamily\small Here is a datetime in ISO-8601 format : 

"1241-07-08T22:56:42"

. Add 250 days to the datetime and generate a new datetime in ISO-8601 format. Let's think step-by-step. Parse the input date components (year, month, day). Add 250 days, accounting for the number of days in each month and leap years. For example, if the datetime is '4939-01-31T12:30:10' and you need to add 250 days, your answer should be '4939-10-08T12:30:10'. As another example, if the datetime is '6553-05-22T03:47:11' and you also need to add 250 days, your answer should be '6554-01-27T03:47:11'}

\textbf{Target:} \texttt{1242-03-15T22:56:42}

\subsection{Table~\ref{datetime_computation_add_250_vllm}:  Add-250 Accuracy (Open-weight)}
\label{datetime_computation_add_250_vllm_repro}
\subsubsection{Datasets \& Comparators}

Release information for each dataset is in Appendix~\ref{datasets}; details on the regex comparators are in Appendix~\ref{regexes}.

\begin{tabular}{lrll}
\textbf{Benchmark} & \textbf{N} & \textbf{Randomised} & \textbf{Comparator} \\
\hline
\texttt{datetime/a3.iso8601.add.day.250.x} & 100 & no & \texttt{iso8601-regex-n} \\
\end{tabular}

\subsubsection{Models}

Information and parameters for each model are in Appendix~\ref{models}. Where applicable, compute resources are detailed in Appendix~\ref{compute_resources}.

\begin{tabular}{ll}
\textbf{Model} & \textbf{Configuration} \\
\hline
\texttt{Llama 3.2 1B Greedy} & meta-llama/Llama-3.2-1B-Instruct \\
\texttt{Llama 3.2 3B Greedy} & meta-llama/Llama-3.2-3B-Instruct \\
\texttt{Gemma 3 1B Greedy} & google/gemma-3-1b-it \\
\texttt{Gemma 3 4B Greedy} & google/gemma-3-4b-it \\
\texttt{Phi 4 Mini Greedy} & microsoft/Phi-4-mini-instruct \\
\texttt{SmolLM2 1.7B Greedy} & HuggingFaceTB/SmolLM2-1.7B-Instruct \\
\texttt{Qwen3 0.6B Greedy} & Qwen/Qwen3-0.6B \\
\texttt{Qwen3 1.7B Greedy} & Qwen/Qwen3-1.7B \\
\texttt{Qwen3 4B Greedy} & Qwen/Qwen3-4B \\
\texttt{OLMo2 1B Greedy} & allenai/OLMo-2-0425-1B-Instruct \\
\texttt{Falcon3 3B Greedy} & tiiuae/Falcon3-3B-Instruct \\
\end{tabular}

\subsubsection{Prompts}

\subsubsubsection{Prompt:Direct-1}

\textbf{System prompt:}\\
{\ttfamily\small Never provide a preamble like 'sure', 'of course', 'here is the answer'. Answer the question directly. Do not provide an explanation.}

\textbf{Prompt template:}\\
{\ttfamily\small Here is a datetime in ISO-8601 format : 

"\{input\_sequence\}"

. Add \{num\_days\} days to the datetime and generate a new datetime in ISO-8601 format.}

\textbf{Rendered example:}\\
{\ttfamily\small Here is a datetime in ISO-8601 format : 

"1241-07-08T22:56:42"

. Add 250 days to the datetime and generate a new datetime in ISO-8601 format.}

\textbf{Target:} \texttt{1242-03-15T22:56:42}

\subsubsubsection{Prompt:Direct-2}

\textbf{System prompt:}\\
{\ttfamily\small Never provide a preamble like 'sure', 'of course', 'here is the answer'. Answer the question directly. Do not provide an explanation.}

\textbf{Prompt template:}\\
{\ttfamily\small Add \{num\_days\} days to the following datetime "\{input\_sequence\}". The answer must be in ISO-8601 format.}

\textbf{Rendered example:}\\
{\ttfamily\small Add 250 days to the following datetime "1241-07-08T22:56:42". The answer must be in ISO-8601 format.}

\textbf{Target:} \texttt{1242-03-15T22:56:42}

\subsubsubsection{Prompt:Direct-1 ISO}

\textbf{System prompt:}\\
{\ttfamily\small Never provide a preamble like 'sure', 'of course', 'here is the answer'. Answer the question directly. Do not provide an explanation.}

\textbf{Prompt template:}\\
{\ttfamily\small Here is a datetime in ISO-8601 format : 

"\{input\_sequence\}"

. Add \{num\_days\} days to the datetime and generate a new datetime in ISO-8601 format. ISO-8601 is the international standard for representing dates and times. A datetime must have the format 'YYYY-MM-DDTHH:MM:SS' (e.g., '2023-03-15T14:30:25')}

\textbf{Rendered example:}\\
{\ttfamily\small Here is a datetime in ISO-8601 format : 

"1241-07-08T22:56:42"

. Add 250 days to the datetime and generate a new datetime in ISO-8601 format. ISO-8601 is the international standard for representing dates and times. A datetime must have the format 'YYYY-MM-DDTHH:MM:SS' (e.g., '2023-03-15T14:30:25')}

\textbf{Target:} \texttt{1242-03-15T22:56:42}

\subsubsubsection{Prompt:Direct-2 ISO}

\textbf{System prompt:}\\
{\ttfamily\small Never provide a preamble like 'sure', 'of course', 'here is the answer'. Answer the question directly. Do not provide an explanation.}

\textbf{Prompt template:}\\
{\ttfamily\small Add \{num\_days\} days to the following datetime "\{input\_sequence\}". The answer must be in ISO-8601 format. ISO-8601 is the international standard for representing dates and times. A datetime must have the format 'YYYY-MM-DDTHH:MM:SS' (e.g., '2023-03-15T14:30:25')}

\textbf{Rendered example:}\\
{\ttfamily\small Add 250 days to the following datetime "1241-07-08T22:56:42". The answer must be in ISO-8601 format. ISO-8601 is the international standard for representing dates and times. A datetime must have the format 'YYYY-MM-DDTHH:MM:SS' (e.g., '2023-03-15T14:30:25')}

\textbf{Target:} \texttt{1242-03-15T22:56:42}

\subsubsubsection{Prompt:FS-1}

\textbf{System prompt:}\\
{\ttfamily\small Never provide a preamble like 'sure', 'of course', 'here is the answer'. Answer the question directly. Do not provide an explanation.}

\textbf{Prompt template:}\\
{\ttfamily\small Here is a datetime in ISO-8601 format : 

"\{input\_sequence\}"

. Add \{num\_days\} days to the datetime and generate a new datetime in ISO-8601 format. For example, if the input datetime is '\{input\_fs\_1\}', then your answer should be '\{target\_fs\_1\}'}

\textbf{Rendered example:}\\
{\ttfamily\small Here is a datetime in ISO-8601 format : 

"1241-07-08T22:56:42"

. Add 250 days to the datetime and generate a new datetime in ISO-8601 format. For example, if the input datetime is '4047-11-01T12:35:06', then your answer should be '4048-07-08T12:35:06'}

\textbf{Target:} \texttt{1242-03-15T22:56:42}

\subsubsubsection{Prompt:FS-2}

\textbf{System prompt:}\\
{\ttfamily\small Never provide a preamble like 'sure', 'of course', 'here is the answer'. Answer the question directly. Do not provide an explanation.}

\textbf{Prompt template:}\\
{\ttfamily\small Here is a datetime in ISO-8601 format : 

"\{input\_sequence\}"

. Add \{num\_days\} days to the datetime and generate a new datetime in ISO-8601 format. For example, if the input datetime is '\{input\_fs\_1\}', then your answer should be '\{target\_fs\_1\}'. As another example, if the input datetime is '\{input\_fs\_2\}', then your answer should be '\{target\_fs\_2\}'}

\textbf{Rendered example:}\\
{\ttfamily\small Here is a datetime in ISO-8601 format : 

"1241-07-08T22:56:42"

. Add 250 days to the datetime and generate a new datetime in ISO-8601 format. For example, if the input datetime is '4047-11-01T12:35:06', then your answer should be '4048-07-08T12:35:06'. As another example, if the input datetime is '8897-06-25T17:00:20', then your answer should be '8898-03-02T17:00:20'}

\textbf{Target:} \texttt{1242-03-15T22:56:42}

\subsubsubsection{Prompt:CoT-1}

\textbf{System prompt:}\\
{\ttfamily\small Never provide a preamble like 'sure', 'of course', 'here is the answer'. Answer the question directly. Do not provide an explanation.}

\textbf{Prompt template:}\\
{\ttfamily\small Here is a datetime in ISO-8601 format : 

"\{input\_sequence\}"

. Add \{num\_days\} days to the datetime and generate a new datetime in ISO-8601 format. Let's think step-by-step. Parse the input date components (year, month, day). Add \{num\_days\} days, accounting for the number of days in each month and leap years.
.}

\textbf{Rendered example:}\\
{\ttfamily\small Here is a datetime in ISO-8601 format : 

"1241-07-08T22:56:42"

. Add 250 days to the datetime and generate a new datetime in ISO-8601 format. Let's think step-by-step. Parse the input date components (year, month, day). Add 250 days, accounting for the number of days in each month and leap years.
.}

\textbf{Target:} \texttt{1242-03-15T22:56:42}

\subsubsubsection{Prompt:CoT-1 + FS-1}

\textbf{System prompt:}\\
{\ttfamily\small Never provide a preamble like 'sure', 'of course', 'here is the answer'. Answer the question directly. Do not provide an explanation.}

\textbf{Prompt template:}\\
{\ttfamily\small Here is a datetime in ISO-8601 format : 

"\{input\_sequence\}"

. Add \{num\_days\} days to the datetime and generate a new datetime in ISO-8601 format. Let's think step-by-step. Parse the input date components (year, month, day). Add \{num\_days\} days, accounting for the number of days in each month and leap years. For example, if the datetime is '4939-01-31T12:30:10' and you need to add 250 days, your answer should be '4939-10-08T12:30:10'}

\textbf{Rendered example:}\\
{\ttfamily\small Here is a datetime in ISO-8601 format : 

"1241-07-08T22:56:42"

. Add 250 days to the datetime and generate a new datetime in ISO-8601 format. Let's think step-by-step. Parse the input date components (year, month, day). Add 250 days, accounting for the number of days in each month and leap years. For example, if the datetime is '4939-01-31T12:30:10' and you need to add 250 days, your answer should be '4939-10-08T12:30:10'}

\textbf{Target:} \texttt{1242-03-15T22:56:42}

\subsubsubsection{Prompt:CoT-1 + FS-2}

\textbf{System prompt:}\\
{\ttfamily\small Never provide a preamble like 'sure', 'of course', 'here is the answer'. Answer the question directly. Do not provide an explanation.}

\textbf{Prompt template:}\\
{\ttfamily\small Here is a datetime in ISO-8601 format : 

"\{input\_sequence\}"

. Add \{num\_days\} days to the datetime and generate a new datetime in ISO-8601 format. Let's think step-by-step. Parse the input date components (year, month, day). Add \{num\_days\} days, accounting for the number of days in each month and leap years. For example, if the datetime is '4939-01-31T12:30:10' and you need to add 250 days, your answer should be '4939-10-08T12:30:10'. As another example, if the datetime is '6553-05-22T03:47:11' and you also need to add 250 days, your answer should be '6554-01-27T03:47:11'}

\textbf{Rendered example:}\\
{\ttfamily\small Here is a datetime in ISO-8601 format : 

"1241-07-08T22:56:42"

. Add 250 days to the datetime and generate a new datetime in ISO-8601 format. Let's think step-by-step. Parse the input date components (year, month, day). Add 250 days, accounting for the number of days in each month and leap years. For example, if the datetime is '4939-01-31T12:30:10' and you need to add 250 days, your answer should be '4939-10-08T12:30:10'. As another example, if the datetime is '6553-05-22T03:47:11' and you also need to add 250 days, your answer should be '6554-01-27T03:47:11'}

\textbf{Target:} \texttt{1242-03-15T22:56:42}

\subsection{Table~\ref{datetime_computation_add_250_unsloth}:  Add-250 Accuracy (Fine-Tuned Models)}
\label{datetime_computation_add_250_unsloth_repro}

\subsection{Table~\ref{datetime_natural_context_prop_llm}:  Event Planning Accuracy (Proprietary Models)}
\label{datetime_natural_context_prop_repro}
\subsubsection{Datasets \& Comparators}

Release information for each dataset is in Appendix~\ref{datasets}; details on the regex comparators are in Appendix~\ref{regexes}.

\begin{tabular}{lrll}
\textbf{Benchmark} & \textbf{N} & \textbf{Randomised} & \textbf{Comparator} \\
\hline
\texttt{datetime/nct.1\_250} & 100 & no & \texttt{boolean-end-1} \\
\end{tabular}

\subsubsection{Models}

Information and parameters for each model are in Appendix~\ref{models}. Where applicable, compute resources are detailed in Appendix~\ref{compute_resources}.

\begin{tabular}{ll}
\textbf{Model} & \textbf{Configuration} \\
\hline
\texttt{OpenAI gpt-3.5-turbo} & gpt-3.5-turbo-0125 \\
\texttt{OpenAI gpt-4.0-turbo} & gpt-4-turbo-2024-04-09 \\
\texttt{OpenAI gpt-4o-mini} & gpt-4o-mini-2024-07-18 \\
\texttt{OpenAI gpt-5} & gpt-5-2025-08-07 \\
\texttt{OpenAI gpt-5-mini} & gpt-5-mini-2025-08-07 \\
\texttt{OpenAI gpt-5-nano} & gpt-5-nano-2025-08-07 \\
\texttt{Anthropic claude-4.5-haiku} & claude-haiku-4-5-20251001 \\
\texttt{Anthropic claude-4.5-sonnet} & claude-sonnet-4-5-20250929 \\
\texttt{Anthropic claude-4.5-opus} & claude-opus-4-5-20251101 \\
\texttt{Anthropic claude-4.0-sonnet} & claude-sonnet-4-20250514 \\
\texttt{Anthropic claude-4.0-opus} & claude-opus-4-20250514 \\
\texttt{Anthropic claude-4.1-opus} & claude-opus-4-1-20250805 \\
\texttt{Gemini 2.5-flash-lite} & gemini-2.5-flash-lite \\
\texttt{Gemini 2.5-flash} & gemini-2.5-flash \\
\texttt{Gemini 2.5-pro} & gemini-2.5-pro \\
\texttt{Gemini 3-flash-preview} & gemini-3-flash-preview \\
\end{tabular}

\subsubsection{Prompts}

\subsubsubsection{Prompt:Direct-1}

\textbf{System prompt:}\\
{\ttfamily\small Never provide a preamble like sure, of course. Do not provide an explanation. Always answer with "Yes" or "No", nothing else.}

\textbf{Prompt template:}\\
{\ttfamily\small I need \{prep\_days\} days to prepare for an event. I can work any day of the week. The event is on \{event\_date\}. If I start preparing on \{start\_date\}, will I be ready in time? Answer with yes or no.}

\textbf{Rendered example:}\\
{\ttfamily\small I need 250 days to prepare for an event. I can work any day of the week. The event is on 2034|30|November,21:54:9. If I start preparing on 2034|14|April,21:54:9, will I be ready in time? Answer with yes or no.}

\textbf{Target:} \texttt{no}

\subsubsubsection{Prompt:Direct-2}

\textbf{System prompt:}\\
{\ttfamily\small Never provide a preamble like sure, of course. Do not provide an explanation. Always answer with "Yes" or "No", nothing else.}

\textbf{Prompt template:}\\
{\ttfamily\small I have an event on \{event\_date\}. I need \{prep\_days\} days to get everything ready, and I can use any day including weekends. Can I make it if I start on \{start\_date\}? Answer with yes or no.}

\textbf{Rendered example:}\\
{\ttfamily\small I have an event on 2034|30|November,21:54:9. I need 250 days to get everything ready, and I can use any day including weekends. Can I make it if I start on 2034|14|April,21:54:9? Answer with yes or no.}

\textbf{Target:} \texttt{no}

\subsubsubsection{Prompt:Direct-3}

\textbf{System prompt:}\\
{\ttfamily\small Never provide a preamble like sure, of course. Do not provide an explanation. Always answer with "Yes" or "No", nothing else.}

\textbf{Prompt template:}\\
{\ttfamily\small Starting on \{start\_date\}, I have \{prep\_days\} days of work to finish before an event on \{event\_date\}. I can work every day of the week. Will I finish on time? Answer with yes or no.}

\textbf{Rendered example:}\\
{\ttfamily\small Starting on 2034|14|April,21:54:9, I have 250 days of work to finish before an event on 2034|30|November,21:54:9. I can work every day of the week. Will I finish on time? Answer with yes or no.}

\textbf{Target:} \texttt{no}

\subsubsubsection{Prompt:FS-1}

\textbf{System prompt:}\\
{\ttfamily\small Never provide a preamble like sure, of course. Do not provide an explanation. Always answer with "Yes" or "No", nothing else.}

\textbf{Prompt template:}\\
{\ttfamily\small I need \{prep\_days\} days to prepare for an event. I can work any day of the week. The event is on \{event\_date\}. If I start preparing on \{start\_date\}, will I be ready in time? Answer with yes or no. For example, if I need 250 days to prepare, the event is on 3:46:13 PM 2027-Dec,thirteenth, and I start on 3:46:13 PM 2027-Apr,fifth, the answer is Yes.}

\textbf{Rendered example:}\\
{\ttfamily\small I need 250 days to prepare for an event. I can work any day of the week. The event is on 2034|30|November,21:54:9. If I start preparing on 2034|14|April,21:54:9, will I be ready in time? Answer with yes or no. For example, if I need 250 days to prepare, the event is on 3:46:13 PM 2027-Dec,thirteenth, and I start on 3:46:13 PM 2027-Apr,fifth, the answer is Yes.}

\textbf{Target:} \texttt{no}

\subsubsubsection{Prompt:FS-2}

\textbf{System prompt:}\\
{\ttfamily\small Never provide a preamble like sure, of course. Do not provide an explanation. Always answer with "Yes" or "No", nothing else.}

\textbf{Prompt template:}\\
{\ttfamily\small I need \{prep\_days\} days to prepare for an event. I can work any day of the week. The event is on \{event\_date\}. If I start preparing on \{start\_date\}, will I be ready in time? Answer with yes or no. For example, if I need 250 days to prepare, the event is on 3:46:13 PM 2027-Dec,thirteenth, and I start on 3:46:13 PM 2027-Apr,fifth, the answer is Yes. As a second example, if I need 250 days to prepare, the event is on 6:39:07 AM,Sat sixth September-2036, and I start on 6:39:07 AM,Thu third January-2036, the answer is No.}

\textbf{Rendered example:}\\
{\ttfamily\small I need 250 days to prepare for an event. I can work any day of the week. The event is on 2034|30|November,21:54:9. If I start preparing on 2034|14|April,21:54:9, will I be ready in time? Answer with yes or no. For example, if I need 250 days to prepare, the event is on 3:46:13 PM 2027-Dec,thirteenth, and I start on 3:46:13 PM 2027-Apr,fifth, the answer is Yes. As a second example, if I need 250 days to prepare, the event is on 6:39:07 AM,Sat sixth September-2036, and I start on 6:39:07 AM,Thu third January-2036, the answer is No.}

\textbf{Target:} \texttt{no}

\subsubsubsection{Prompt:CoT-1}

\textbf{System prompt:}\\
{\ttfamily\small Never provide a preamble like sure, of course. Do not provide an explanation. Always answer with "Yes" or "No", nothing else.}

\textbf{Prompt template:}\\
{\ttfamily\small I need \{prep\_days\} days to prepare for an event. I can work any day of the week. The event is on \{event\_date\}. If I start preparing on \{start\_date\}, will I be ready in time? Answer with yes or no. Think step by step. First, parse the start date and the event date. Second, add the number of preparation days to the start date to find the completion date. Third, compare the completion date to the event date. If the completion date is on or before the event date, the answer is Yes. Otherwise, the answer is No.}

\textbf{Rendered example:}\\
{\ttfamily\small I need 250 days to prepare for an event. I can work any day of the week. The event is on 2034|30|November,21:54:9. If I start preparing on 2034|14|April,21:54:9, will I be ready in time? Answer with yes or no. Think step by step. First, parse the start date and the event date. Second, add the number of preparation days to the start date to find the completion date. Third, compare the completion date to the event date. If the completion date is on or before the event date, the answer is Yes. Otherwise, the answer is No.}

\textbf{Target:} \texttt{no}

\subsubsubsection{Prompt:CoT-1 + FS-1}

\textbf{System prompt:}\\
{\ttfamily\small Never provide a preamble like sure, of course. Do not provide an explanation. Always answer with "Yes" or "No", nothing else.}

\textbf{Prompt template:}\\
{\ttfamily\small I need \{prep\_days\} days to prepare for an event. I can work any day of the week. The event is on \{event\_date\}. If I start preparing on \{start\_date\}, will I be ready in time? Answer with yes or no. Think step by step. First, parse the start date and the event date. Second, add the number of preparation days to the start date to find the completion date. Third, compare the completion date to the event date. If the completion date is on or before the event date, the answer is Yes. Otherwise, the answer is No. For example, if I need 250 days to prepare, the event is on 3:46:13 PM 2027-Dec,thirteenth, and I start on 3:46:13 PM 2027-Apr,fifth, the answer is Yes.}

\textbf{Rendered example:}\\
{\ttfamily\small I need 250 days to prepare for an event. I can work any day of the week. The event is on 2034|30|November,21:54:9. If I start preparing on 2034|14|April,21:54:9, will I be ready in time? Answer with yes or no. Think step by step. First, parse the start date and the event date. Second, add the number of preparation days to the start date to find the completion date. Third, compare the completion date to the event date. If the completion date is on or before the event date, the answer is Yes. Otherwise, the answer is No. For example, if I need 250 days to prepare, the event is on 3:46:13 PM 2027-Dec,thirteenth, and I start on 3:46:13 PM 2027-Apr,fifth, the answer is Yes.}

\textbf{Target:} \texttt{no}

\subsubsubsection{Prompt:CoT-1 + FS-2}

\textbf{System prompt:}\\
{\ttfamily\small Never provide a preamble like sure, of course. Do not provide an explanation. Always answer with "Yes" or "No", nothing else.}

\textbf{Prompt template:}\\
{\ttfamily\small I need \{prep\_days\} days to prepare for an event. I can work any day of the week. The event is on \{event\_date\}. If I start preparing on \{start\_date\}, will I be ready in time? Answer with yes or no. Think step by step. First, parse the start date and the event date. Second, add the number of preparation days to the start date to find the completion date. Third, compare the completion date to the event date. If the completion date is on or before the event date, the answer is Yes. Otherwise, the answer is No. For example, if I need 250 days to prepare, the event is on 3:46:13 PM 2027-Dec,thirteenth, and I start on 3:46:13 PM 2027-Apr,fifth, the answer is Yes. As a second example, if I need 250 days to prepare, the event is on 6:39:07 AM,Sat sixth September-2036, and I start on 6:39:07 AM,Thu third January-2036, the answer is No.}

\textbf{Rendered example:}\\
{\ttfamily\small I need 250 days to prepare for an event. I can work any day of the week. The event is on 2034|30|November,21:54:9. If I start preparing on 2034|14|April,21:54:9, will I be ready in time? Answer with yes or no. Think step by step. First, parse the start date and the event date. Second, add the number of preparation days to the start date to find the completion date. Third, compare the completion date to the event date. If the completion date is on or before the event date, the answer is Yes. Otherwise, the answer is No. For example, if I need 250 days to prepare, the event is on 3:46:13 PM 2027-Dec,thirteenth, and I start on 3:46:13 PM 2027-Apr,fifth, the answer is Yes. As a second example, if I need 250 days to prepare, the event is on 6:39:07 AM,Sat sixth September-2036, and I start on 6:39:07 AM,Thu third January-2036, the answer is No.}

\textbf{Target:} \texttt{no}

\subsection{Table~\ref{datetime_nc_250_vllm}:  Event Planning Accuracy (Open-weight)}
\label{datetime_natural_context_vllm_repro}

\subsection{Table~\ref{datetime_event_planning_250_unsloth}:  Event Planning Accuracy (Fine-Tuned Models)}
\label{datetime_natural_context_unsloth_repro}
\subsubsection{Datasets \& Comparators}

Release information for each dataset is in Appendix~\ref{datasets}; details on the regex comparators are in Appendix~\ref{regexes}.

\begin{tabular}{lrll}
\textbf{Benchmark} & \textbf{N} & \textbf{Randomised} & \textbf{Comparator} \\
\hline
\texttt{datetime/nct.1\_250} & 100 & no & \texttt{boolean-end-1} \\
\end{tabular}

Fine-tuning dataset is \texttt{datetime/nct.1\_250\_ft\_1m}, with no overlapping start date with the benchmark; see Section~\ref{descriptives_ft}.

\subsubsection{Models}

Information and parameters for each model are in Appendix~\ref{models}. Where applicable, compute resources are detailed in Appendix~\ref{compute_resources}.

\begin{tabular}{ll}
\textbf{Model} & \textbf{Configuration} \\
\hline
\texttt{unsloth/Qwen2.5-0.5B} & unsloth/Qwen2.5-0.5B \\
\texttt{unsloth/Qwen2.5-1.5B} & unsloth/Qwen2.5-1.5B \\
\texttt{unsloth/Qwen2.5-1.5B Instruct} & unsloth/Qwen2.5-1.5B-Instruct \\
\texttt{unsloth/Qwen2.5-3B} & unsloth/Qwen2.5-3B \\
\texttt{unsloth/Qwen2.5-3B Instruct} & unsloth/Qwen2.5-3B-Instruct \\
\texttt{unsloth/Qwen2.5-7B Instruct} & unsloth/Qwen2.5-7B-Instruct \\
\texttt{unsloth/Qwen2.5-14B Instruct} & unsloth/Qwen2.5-14B-Instruct \\
\end{tabular}

\subsubsection{Prompts}

A single prompt template is used for both fine-tuning and inference; the limitation of evaluating fine-tuned models under a single prompt is discussed in Section~\ref{limitations}.

\textbf{System prompt:}\\
{\ttfamily\small Never provide a preamble like sure, of course. Do not provide an explanation. Always answer with "Yes" or "No", nothing else.}

\textbf{Prompt template:}\\
{\ttfamily\small I need \{prep\_days\} days to prepare for an event. I can work any day of the week. The event is on \{event\_date\}. If I start preparing on \{start\_date\}, will I be ready in time? Answer with yes or no.}

\textbf{Rendered example:}\\
{\ttfamily\small I need 250 days to prepare for an event. I can work any day of the week. The event is on 2034|30|November,21:54:9. If I start preparing on 2034|14|April,21:54:9, will I be ready in time? Answer with yes or no.}

\textbf{Target:} \texttt{no}

  \clearpage

\section{Reasoning steps required to solve PRIMETIME Computation tasks}   
  
\begin{figure}[H]
  \includegraphics[trim={0 0 0 0}, clip, width=1.0\textwidth, page=1]{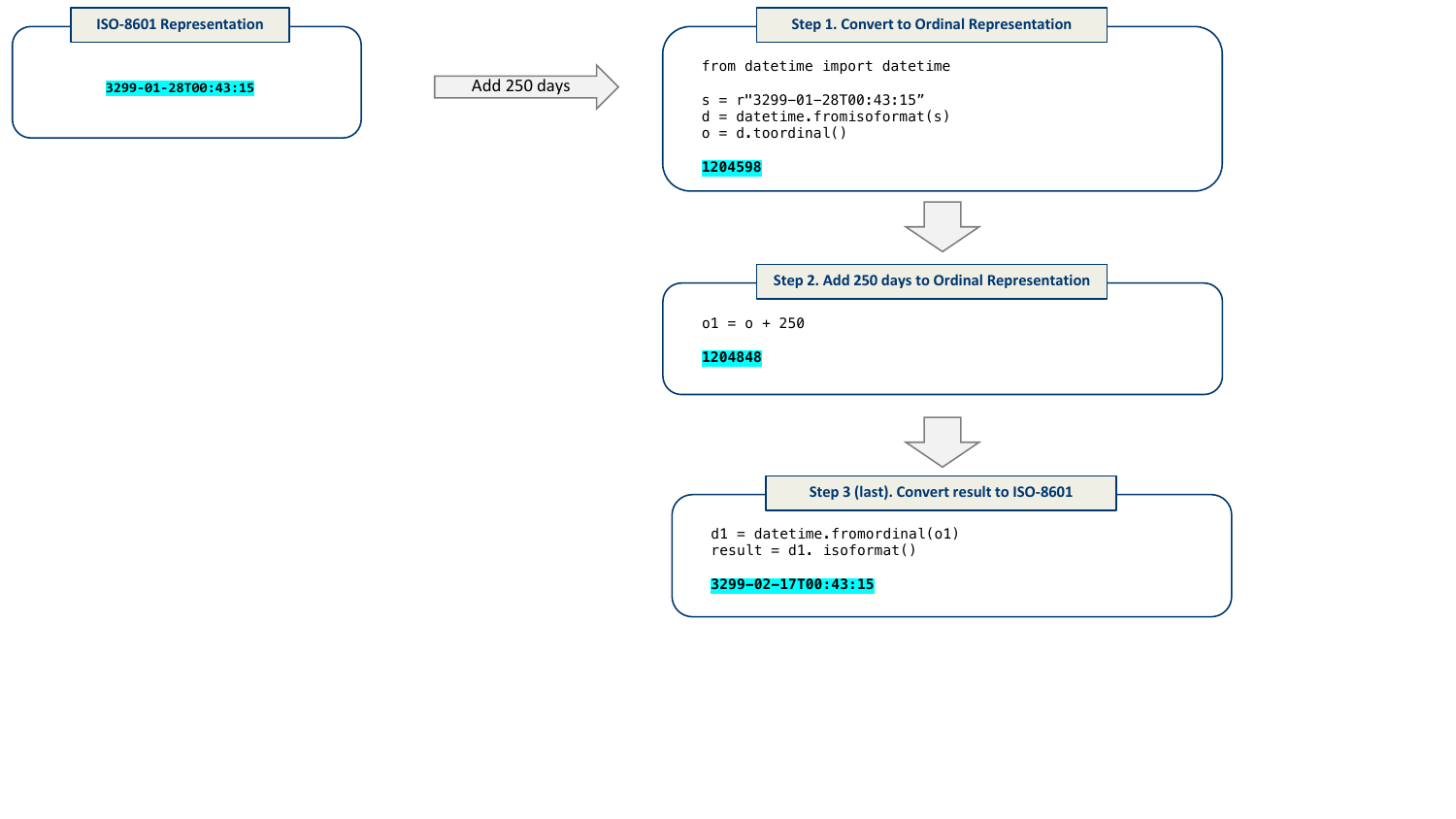}
  \caption{Required Python steps to resolve the Add-250 task. A solution requires converting the date to an ordinal, i.e. the number of days since 1st Jan in year 1. From there, 250 days can easily be added using a plain addition.}
  \label{fig:add-250-reasonong-steps-python}
\end{figure}

\begin{figure}[H]
  \includegraphics[trim={0 0 0 0}, clip, width=1.0\textwidth, page=2]{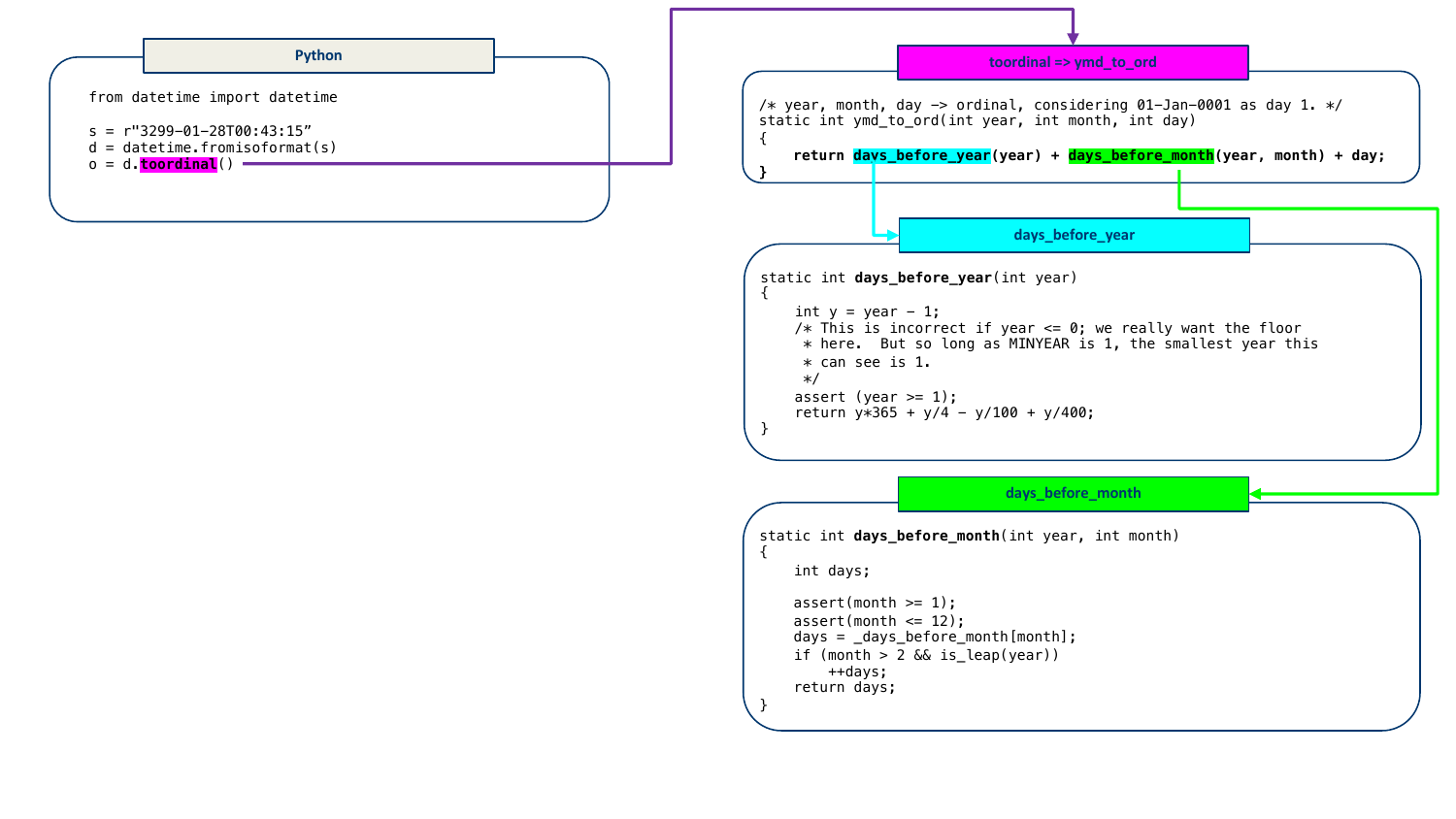}
  \caption{Underlying CPython function calls. Source available code on GitHub \url{https://github.com/python/cpython/_datetimemodule.c} }
  \label{fig:add-250-reasonong-steps-cpython}
\end{figure}

This section illustrates the required reasoning steps to solve the \emph{Computation} tasks, in Python (Figure \ref{fig:add-250-reasonong-steps-python}) and in the underlying CPython (Figure \ref{fig:add-250-reasonong-steps-cpython}).

\medbreak
\medbreak

\newpage

  \label{appendix_reasoning_steps}
  \clearpage

\section{Regex Expressions for output sequence evaluation} 
  \label{regexes}
  This section describes the regexes used to extract predictions from model outputs. 

Because LLMs may not have been designed or trained for temporal reasoning, in particular working with or generating datetimes in ISO-8601 format, the output sequences are parsed using lenient regexes. Several regexes are employed. 

For tasks requiring an ISO~8601 output, a first regex captures \emph{all} extended-format ISO~8601 sequences in the response, tolerating omitted zero-padding and either \texttt{T} or a space as the date-time separator\footnote{\url{https://en.wikipedia.org/wiki/ISO_8601}}. Date-only outputs (e.g.\ \texttt{2023-12-19}) are also captured, in case a model omits the time component. However, evaluation always compares against the full year-month-day-hour-minute-second target. A second regex then parses the captured regexes into date and time components. All captured sequences are compared against the gold label; if any match, the prediction is scored as correct. This lenient policy credits the model for arriving at the correct answer even if it also produces intermediate or alternative candidates. We note that this would be unsuitable for production use, where a single unambiguous output is required; however, the goal here is to assess core temporal reasoning ability, not output formatting.
For \emph{Event Planning}, a third regex parses the output for \textit{Yes} or \textit{No}, case-insensitive, and considers only the last match as the model's final answer. All regexes were unit-tested and property-tested using the Hypothesis library, and adjusted based on inspection of preliminary results. 

\subsection{ISO-8601 Regex Extraction Patterns}
\mbox{}\\
\textbf{Pattern ID:} \texttt{iso8601-regex-n}
\label{regex_iso8601}

Two regex patterns are used to extract predictions from model outputs for \emph{Translation}, \emph{Add-250} and \emph{Mixed-Add-250} tasks.

\paragraph{Step 1. ISO~8601 capture.}
The first regex captures datetime strings in ISO~8601 extended format. It requires a four-digit year, followed by month and day separated by hyphens. An optional time component, introduced by \texttt{T} or a space, captures hours, minutes, and seconds separated by colons. Zero-padding is not required for any component. Only the extended ISO~8601 variant with hyphen and colon separators is matched.
\begin{verbatim}
\d{4}-\d{1,2}-\d{1,2}(?:[T ]\d{1,2}(?::\d{1,2})*[^\s]*?)?
\end{verbatim}

\paragraph{Step 2. ISO~8601 parsing.}
A second regex decomposes a captured string into named groups (\texttt{year}, \texttt{month}, \texttt{day}, \texttt{hour}, \texttt{minute}, \texttt{second}) for evaluation against the gold label.
All named groups are optional, allowing partial matches at varying levels of temporal granularity. A catch-all group captures any remaining characters after the last matched component; these are discarded during evaluation.

\begin{verbatim}
(?P<year>\d{4})?(?:-(?P<month>\d{1,2}))?(?:-(?P<day>\d{1,2}))?
(?:[T ](?P<hour>\d{1,2}))?(?::(?P<minute>\d{1,2}))?
(?::(?P<second>\d{1,2}))?(?P<suffix>.*)
\end{verbatim}

\subsection{Yes/No  Regex Extraction Patterns}
For \emph{Event Planning}, a case-insensitive regex matches all occurrences of \textit{Yes} or \textit{No} followed by a word boundary.

\begin{verbatim}
(yes|no)\b
\end{verbatim}

The leading word boundary was omitted after manual inspection of model outputs revealed cases where the answer is appended directly after a non-whitespace character. For example, Claude-4.1-opus produced output ending in:

\begin{verbatim}
...March 25, 2037 is after February 28, 2037No
\end{verbatim}

where \texttt{No} follows a digit with no preceding space or punctuation. A leading \verb|\b| fails to match here, as the transition from a digit to a letter does not constitute a word boundary.

\paragraph{ISO~8601 capture property-based test.}
Property-based tests verify that the regex function as intended. For example, we must ensure that valid ISO~8601 strings are captured regardless of zero-padding, separator choice, or preceding text.
{\scriptsize
    \begin{verbatim}
    @given(
            year=st.integers(min_value=1000, max_value=9999),
            month=st.integers(min_value=1, max_value=12),
            day=st.integers(min_value=1, max_value=28),
            hour=st.integers(min_value=0, max_value=23),
            minute=st.integers(min_value=0, max_value=59),
            second=st.integers(min_value=0, max_value=59),
            sep=st.sampled_from(['T', ' ']),
            prefix=st.text(min_size=0, max_size=100),
        )
        def test_iso8601_regex_n_capture_fuzz_valid_no_padding(self, year, month, day, hour, minute, second, sep, prefix):
            """Valid ISO-8601 strings event with no padding should always be captured."""

            self.iso8601_regex_n_comparator.debug = False
            
            iso = f"{year}-{month}-{day}{sep}{hour}:{minute}:{second}"
            text = prefix + iso
            result = self.iso8601_regex_n_comparator(iso, text)
            self.assertTrue(result, f"Failed to capture non-padded ISO-8601: {iso!r} in {text!r}")
    \end{verbatim}
}
 
  \clearpage

  \section{Models evaluated in the benchmark} 
  \label{models}
  
This section provides configuration details and hyperparamaters of the models used throughout the paper. All experiments were carried out between 9th July 2025 and 15th April 2026. 

\subsection{Proprietary Models}
\begin{itemize}
  \item Temperature is set to 0.0 across all models and experimentsTemperature is set to zero for all models where supported\footnote{some APIs do not accept the temperature parameter, e.g.\ the OpenAI o-series}.
  \item Maximum number of generated tokens (\emph{max\_tokens}) is set to 3000 to leave room for \emph{scratchpad} output (Section~\ref{tasks}).
  \item All other settings retain API defaults.
  \item Batch APIs are used for OpenAI, Anthropic, and Gemini, reverting to single calls during transient infrastructure errors (service unavailable, timeouts).
  \item Grok is not evaluated due to prohibitive API costs and no response to a request for academic access.
  \item DeepSeek was not evaluated due to API access constraints.
\end{itemize}

\begin{tabular}{ll}
\textbf{Model} & \textbf{Configuration} \\
\hline
\texttt{Anthropic claude-4.0-sonnet} & claude-sonnet-4-20250514 \\
\texttt{Anthropic claude-4.0-opus} & claude-opus-4-20250514 \\
\texttt{Anthropic claude-4.1-opus} & claude-opus-4-1-20250805 \\
\texttt{OpenAI gpt-3.5-turbo} & gpt-3.5-turbo-0125 \\
\texttt{OpenAI gpt-4.0-turbo} & gpt-4-turbo-2024-04-09 \\
\texttt{OpenAI gpt-4o-mini} & gpt-4o-mini-2024-07-18 \\
\texttt{OpenAI gpt-5} & gpt-5-2025-08-07 \\
\texttt{OpenAI gpt-5-mini} & gpt-5-mini-2025-08-07 \\
\texttt{OpenAI gpt-5-nano} & gpt-5-nano-2025-08-07 \\
\texttt{Gemini gemini-2.5-flash-lite} & gemini-2.5-flash-lite \\
\texttt{Gemini gemini-2.5-flash} & gemini-2.5-flash \\
\texttt{Gemini gemini-2.5-pro} & gemini-2.5-pro \\
\end{tabular}

\subsection{Open-weight Models}
\label{inference}
Computational resources and setup for evaluating the open-weight models are detailed in Section~\ref{compute_resources}.

\begin{itemize}
  \item Inference uses vLLM, versions 0.15.1 and 0.16.0.
  \item \texttt{max\_model\_len} is set to 3584 tokens, comprising the 3000-token output budget plus 584 tokens for the input (system prompt, user template, and rendered datetime).
  \item Decoding is greedy (\texttt{temperature} = 0.0).
  \item Server seed and request seed are both set to 42; under greedy decoding the request seed is functionally irrelevant (argmax is deterministic) and is included for completeness.
  \item Gemma 3 models run with their native bfloat16 weights; all other models use float16.
  \item Inference is performed against vLLM's OpenAI-compatible chat completions endpoint (\texttt{/v1/chat/completions}).
  \item For models that emit reasoning in dedicated thinking tokens (e.g. \texttt{<think>...</think>} blocks), the prediction is taken as the content after the closing \texttt{</think>} tag; the thinking trace itself is discarded before evaluation.
\end{itemize}

\begin{tabular}{ll}
\textbf{Model} & \textbf{Configuration} \\
\hline
\texttt{Llama 3.2 1B Greedy} & meta-llama/Llama-3.2-1B-Instruct \\
\texttt{Llama 3.2 3B Greedy} & meta-llama/Llama-3.2-3B-Instruct \\
\texttt{Gemma 3 1B Greedy} & google/gemma-3-1b-it \\
\texttt{Gemma 3 4B Greedy} & google/gemma-3-4b-it \\
\texttt{Phi 4 Mini Greedy} & microsoft/Phi-4-mini-instruct \\
\texttt{SmolLM2 1.7B Greedy} & HuggingFaceTB/SmolLM2-1.7B-Instruct \\
\texttt{Qwen3 0.6B Greedy} & Qwen/Qwen3-0.6B \\
\texttt{Qwen3 1.7B Greedy} & Qwen/Qwen3-1.7B \\
\texttt{Qwen3 4B Greedy} & Qwen/Qwen3-4B \\
\texttt{OLMo2 1B Greedy} & allenai/OLMo-2-0425-1B-Instruct \\
\texttt{Falcon3 3B Greedy} & tiiuae/Falcon3-3B-Instruct \\
\end{tabular}

\subsection{Fine-Tuned Models}
\label{fine_tuning}

\paragraph{Models.} All fine-tuned models are Qwen~2.5 \citep{qwen2025qwen25technicalreport} variants from 0.5B to 14B parameters, in both Base and Instruct versions, fine-tuned from Unsloth's pre-quantized 4-bit releases\footnote{\url{https://www.unsloth.ai/license}}. Models are available on HuggingFace at \texttt{https://huggingface.co/unsloth/Qwen2.5-\{size\}B} and the corresponding Instruct variants with suffix \texttt{-Instruct} in the URL. Computational resources and setup are detailed in Section~\ref{compute_resources}.

\paragraph{Fine-tuning prompt.} A single prompt template per task is used for both fine-tuning and inference of the fine-tuned models; the limitation of evaluating fine-tuned models under a single prompt is discussed in Section~\ref{limitations}. The system prompt, user prompt template and a rendered example are provided in Appendix~\ref{datetime_translation_unsloth_repro} for \emph{Translation}, Appendix~\ref{datetime_computation_add_250_unsloth_repro} for \emph{Add-250}, and Appendix~\ref{datetime_natural_context_unsloth_repro} for \emph{Event Planning}, respectively.

\paragraph{Fine-tuning configuration.} We used Unsloth version 2026.3.4 with default settings except where noted below. 

\begin{itemize}
  \item \texttt{max\_seq\_length = 128}
  \item \texttt{dtype = bfloat16}
  \item \texttt{load\_in\_4bit = True}
  \item Number of epochs: 1
  \item \texttt{per\_device\_train\_batch\_size}: 32 (0.5B) decreasing to 8 (14B)
  \item \texttt{gradient\_accumulation\_steps}: 1 (0.5B) increasing to 8 (14B)
  \item \texttt{gradient\_checkpointing = "unsloth"}
  \item \texttt{warmup\_steps = 5}
  \item \texttt{learning\_rate = 2e-4}
  \item \texttt{lr\_scheduler\_type = "linear"}
  \item \texttt{weight\_decay = 0.01}
  \item \texttt{optim = "adamw\_8bit"}
  \item \texttt{bf16 = True}
  \item \texttt{seed = 3407}
  \item LoRA default values; \texttt{rank = 16}; \texttt{lora\_alpha = 16}; \texttt{lora\_dropout = 0}; \texttt{bias = "none"}; \texttt{use\_rslora = False}; \texttt{loftq\_config = None}
  \item LoRA default target modules: \texttt{q\_proj}, \texttt{k\_proj}, \texttt{v\_proj}, \texttt{o\_proj}, \texttt{gate\_proj}, \texttt{up\_proj}, \texttt{down\_proj}
  \item Number of training observations varies per experiment and is reported in the relevant column of each results table
\end{itemize}

\paragraph{Software environment.} CUDA 12.4 to 12.8; \texttt{torch==2.10.0+cu126}; \texttt{transformers==5.2.0}; \texttt{trl==0.24.0}; \texttt{datasets==4.3.0}; \texttt{peft==0.18.1}; \texttt{accelerate==1.13.0}.

\paragraph{Model IDs.} 
\begin{tabular}{ll}
\textbf{Model} & \textbf{Configuration} \\
\hline
\texttt{unsloth/Qwen2.5-0.5B} & unsloth/Qwen2.5-0.5B \\
\texttt{unsloth/Qwen2.5-1.5B} & unsloth/Qwen2.5-1.5B \\
\texttt{unsloth/Qwen2.5-1.5B Instruct} & unsloth/Qwen2.5-1.5B-Instruct \\
\texttt{unsloth/Qwen2.5-3B} & unsloth/Qwen2.5-3B \\
\texttt{unsloth/Qwen2.5-3B Instruct} & unsloth/Qwen2.5-3B-Instruct \\
\texttt{unsloth/Qwen2.5-7B Instruct} & unsloth/Qwen2.5-7B-Instruct \\
\texttt{unsloth/Qwen2.5-14B Instruct} & unsloth/Qwen2.5-14B-Instruct \\
\end{tabular}

\TODO{Lora Config}

  \clearpage

\section{Experiments Compute Resources} 
  \label{compute_resources}
  This section provides details on the computational resources and setup for evaluating the open-weight models, fine-tuning and running long-lasting batch jobs for the proprietary APIs. The closed-source models were evaluated using each vendor's API, for which computational information is not available.

\subsection{GPU Server 1}
\begin{itemize}
    \item Operating System : Ubuntu 22.04.5 LTS
    \item Main RAM : 256GB 
    \item Main CPU :AMD EPYC 7313P 16-Core Processor
    \item 1 GPUs : NVIDIA RTX 6000 Ada Generation, 49140 MiB
    \item CUDA Version: 12.4
    \item Disk storage : 1TB
\end{itemize}

\subsection{GPU Server 2}
\begin{itemize}
    \item Operating System : Ubuntu 24.04.4 LTS
    \item Main RAM : 256GB 
    \item Main CPU : 24-core AMD EPYC 7402P 24-Core Processor
    \item 4 GPUs : NVIDIA GeForce RTX 3090, each with 24GB VRAM 
    \item CUDA Version: 12.8
    \item Disk storage : 100GB
\end{itemize}

\subsection{GPU Server 3}
\begin{itemize}
    \item Operating System : Ubuntu 24.04.4 LTS
    \item Main RAM : 192GB 
    \item Main CPU : 32-core Xeon® E5-2682 v4
    \item 4 GPUs : NVIDIA GeForce RTX 3060, each with 12GB VRAM 
    \item CUDA Version: 12.8
    \item Disk storage : 100GB
\end{itemize}

\subsection{GPU Server 4}
\begin{itemize}
    \item Operating System : Ubuntu 24.04.4 LTS
    \item Main RAM : 128GB 
    \item Main CPU : 16-core AMD EPYC 7302P
    \item 4 GPUs : NVIDIA GeForce RTX 3070, each with 8GB VRAM 
    \item CUDA Version: 12.8
    \item Disk storage : 100GB
\end{itemize}

\subsection{AUX Server 1}
Used for running long-lasting batch jobs for the proprietary APIs.
\begin{itemize}
    \item Operating System : Ubuntu 22.04.5 LTS
    \item Main RAM : 64GB 
    \item Main CPU : 32-core Intel(R) Xeon(R) CPU E5-2620 v4 @ 2.10GHz
    \item Disk storage : 439G
\end{itemize}

\subsection{Aux Server 2}
Used for LLM-as-Judge for the failure analysis (see Appendix\ref{failure_analysis}).
\begin{itemize}
    \item Operating System : Ubuntu 24.04.2 LTS
    \item Main RAM : 192GB 
    \item Main CPU : 64-core Intel(R) Xeon(R) Gold 6242 CPU @ 2.80GHz
    \item 4 GPUs : 3x NVIDIA TITAN V and 1x NVIDIA TITAN Xp, each with 12GB VRAM
    \item CUDA Version: 12.8
    \item Disk storage : 920GB
\end{itemize}

  \clearpage

\section{Examples of datetime tasks in existing benchmarks} \label{appendix_existing_datetime_tasks}
  
Verbatim examples of date and datetime related tasks found in the literature.

\subsection{Date understanding in BIG-Bench Hard \cite{suzgun2022challenging}}
Date Understanding already exists to a limited in the BIG-Bench Hard tasks.

\begin{lstlisting}
    Today is Christmas Eve of 1937. What is the date tomorrow in MM/DD/YYYY? Options: 
    (A) 12/11/1937 
    (B) 12/25/1937 
    (C) 01/04/1938 
    (D) 12/04/1937 
    (E) 12/25/2006 
    (F) 07/25/1937
    
    Answer: (B)

\end{lstlisting}

\begin{lstlisting}
The concert was scheduled to be on 06/01/1943, but was delayed by one day to today. What is the date
yesterday in MM/DD/YYYY?
\end{lstlisting}

\subsection{Example from QuAC \citep{choi2018quacquestionanswering}}
\begin{lstlisting}
    Context: Isaac Asimov, Biography: Asimov was born on an unknown day between October 4, 1919 and January 2, 1920, inclusive. Asimov himself celebrated it on January 2. Asimov's parents were Anna Rachel (nee Berman) and Judah Asimov, a family of Jewish millers. He was named Isaac after his mother's father, Isaac Berman. When he was born, his family lived in Petrovichi near Klimovichi, which was then Gomel Governorate in the Russian Soviet Federative Socialist Republic (now Smolensk Oblast, Russia). Asimov wrote of his father, "My father, for all his education as an Orthodox Jew, was not Orthodox in his heart", noting that "he didn't recite the myriad prayers prescribed for every action, and he never made any attempt to teach them to me". In 1921, Asimov and 16 other children in Petrovichi caught double pneumonia. Only Asimov survived. He later had two younger siblings: a sister, Marcia (born Manya, June 17, 1922 - April 2, 2011), and a brother, Stanley (July 25, 1929 - August 16, 1995), who was vice-president of New York Newsday. His family emigrated to the United States when he was three years old. Since his parents always spoke Yiddish and English with him, he never learned Russian, but he remained fluent in Yiddish as well as English. Growing up in Brooklyn, New York, Asimov taught himself to read at the age of five, and his mother got him into first grade a year early by claiming he was born on September 7, 1919. In third grade he learned about the "error" and insisted on an official correction of the date to January 2. After becoming established in the U.S., his parents owned a succession of candy stores, in which everyone in the family was expected to work. The candy stores sold newspapers and magazines, a fact that Asimov credited as a major influence in his lifelong love of the written word, as it presented him with an unending supply of new reading material as a child that he could not have otherwise afforded. He became a naturalized U.S. citizen in 1928 at the age of eight.

    Question: What is the closest date?

    Answer: October 4, 1919 and January 2, 1920,
\end{lstlisting}

\subsection{Example from WNLI \citep{Levesque2012TheWS} (part of SuperGLUE \citep{wang2020superglue})}
\begin{lstlisting}
    Thomson visited Cooper's grave in 1765. At that date he had been dead for five years.   Cooper had been dead for five years.
\end{lstlisting}

\subsection{Example from BoolQ \citep{clark2019boolqexploringsurprisingdifficulty} (part of SuperGLUE \citep{wang2020superglue})}
\begin{lstlisting}
    "question": "can you still use old 5 euro note", "passage": "5 euro note -- The changeover period during which the former currencies' notes and coins were exchanged for those of the euro lasted about two months, going from 1 January 2002 until 28 February 2002. The official date on which the national currencies ceased to be legal tender varied from member state to member state. The earliest date was in Germany, where the mark officially ceased to be legal tender on 31 December 2001, though the exchange period lasted for two months more. Even after the old currencies ceased to be legal tender, they continued to be accepted by national central banks for periods ranging from ten years to forever.", "idx": 1125, "label": false
\end{lstlisting}

\subsection{Example from MultiRC \citep{khashabi2016questionansweringintegerprogramming} (part of SuperGLUE \citep{wang2020superglue})}
\begin{lstlisting}
    The Romans: Legend says Rome was founded by Romulus, sired with twin brother Remus by Mars of a Vestal Virgin and abandoned on the Palatine Hill to be suckled by a she-wolf. Historians agree with the mythmakers that the site and traditional founding date of 753 b.c. are just about right. Under Etruscan domination, Rome had been a monarchy until a revolt in 510 b.c. established a patrician republic, which lasted five centuries. In contrast to other Italian cities weakened by internal rivalries and unstable government, Rome drew strength from a solid aristocracy of consuls and senate ruling over plebeians proud of their Roman citizenship and only rarely rebellious. Recovering quickly from the Gallic invasion of 390 b.c. , the Romans took effective control of the peninsula by a military conquest reinforced by a network of roads with names that exist to this day: Via Appia, Flaminia, Aurelia. All roads did indeed lead to\u2002\u2014\u2002and from\u2002\u2014\u2002Rome. By 250 b.c. , the city's population had grown to an impressive 100,000. Roman power extended throughout the Mediterranean with a victory in the Punic Wars against Carthage (now Tunisia) and conquests in Macedonia, Asia Minor, Spain, and southern France. The rest of Italy participated only by tax contributions to the war effort and minor involvement in commerce and colonization. Resentment surfaced when former Etruscan or Greek cities such as Capua, Syracuse, and Taranto supported Hannibal's invasion in 218 b.c. Rome followed up defeat of the Carthaginians with large-scale massacres and enslavement of their Italian supporters. The Third and final Punic War ended in 149 b.c. , though national solidarity was still a long way off. Under Julius Caesar, elected in 59 b.c. ", 
    "question": "According to myth in what year was Rome founded and on what site?"
\end{lstlisting}

\subsection{Example from Recognizing Textual Entailment (RTE) (part of SuperGLUE \citep{wang2020superglue})}
\begin{lstlisting}
{"premise": "Though the exact date is debated, dogs are thought to have been domesticated by humans approximately 12,000 years ago.", "hypothesis": "Humans existed 10,000 years ago.", "label": "entailment", "idx": 146}
\end{lstlisting}

\subsection{Date reasoning in Natural language embedded programs (NLEP)}
The novel NLEP approach \citep{zhang2024naturallanguageembeddedprograms} uses a LLM to generate python code which is subsequently compiled and executed to solve datetime computation tasks.
Whilst no benchmark is provided, the challenge related to understandig and processing datetimes is explicit.
\begin{lstlisting}
us_presidents = {
    "Dwight D. Eisenhower": {"birth_date": "1890-10-14", "term_start": "1953-01-20"},
    "John F. Kennedy": {"birth_date": "1917-05-29", "term_start": "1961-01-20"},
    "Lyndon B. Johnson": {"birth_date": "1908-08-27", "term_start": "1963-11-22"},
    "Richard Nixon": {"birth_date": "1913-01-09", "term_start": "1969-01-20"},
    "Gerald Ford": {"birth_date": "1913-07-14", "term_start": "1974-08-09"},
    Structured
    knowledge
    "Jimmy Carter": {"birth_date": "1924-10-01", "term_start": "1977-01-20"},
    "Ronald Reagan": {"birth_date": "1911-02-06", "term_start": "1981-01-20"},
    "George H. W. Bush": {"birth_date": "1924-06-12", "term_start": "1989-01-20"},
    "Bill Clinton": {"birth_date": "1946-08-19", "term_start": "1993-01-20"},
    "George W. Bush": {"birth_date": "1946-07-06", "term_start": "2001-01-20"},
    "Barack Obama": {"birth_date": "1961-08-04", "term_start": "2009-01-20"},
    "Donald Trump": {"birth_date": "1946-06-14", "term_start": "2017-01-20"},
    "Joe Biden": {"birth_date": "1942-11-20", "term_start": "2021-01-20"},
}

The US presidents who were elected after 1950 and born on Mondays are:

\end{lstlisting}

  \clearpage

\section{Classification of LLM tasks} \label{appendix_existing_task_classification}
  
The following classification scheme for existing LLM benchmarks was used throughout Appendix \ref{appendix_benchmark_review}. 

Surprisingly, the review of \numbenchmarks frequently used LLM benchmarks reveals no granular classification of the tasks. The taxonomies used in the 16 core scenarios of HELM \citep{liang2023holisticevaluationlanguagemodels} or the three categories in MT-Bench \citep{zheng2023judgingllmasajudgemtbenchchatbot} are too high-level for determining if datetime tasks have already been researched.

Only text-to-text tasks in English were considered, i.e., not considering multi-modal input domains such as images or files. Programming tasks were also out of scope.

\begin{itemize}
 \item Question answering
 \item Pattern / Sequence prediction
 \item Verifiable instructions, instruction following
 \item Chat conversation (e.g. multi-turn)
 \item Textual entailment or contradiction
 \item Reading comprehension (NLU)
 \item Sentence (di-)similarity
 \item Linguistic acceptability
 \item Sentiment analysis
 \item Entailment
 \item Completion (word prediction, sentence completion, ending)
 \item Word sense disambiguation
 \item Part-of-speech tagging (POS)
 \item Named-entity Recognition
 \item Speaker commitment
 \item Translation
 \item Social biases
 \item Reasoning and algorithmic
 \item World Knowledge
 \item Expert Knowledge
 \item Date understanding and reasoning
 \item Programming / Coding
 \item Question generation (creation of questions)
 \item Commonsense reasoning (NLI, what likely happens next, physical, social)
 \item Maths, Physics (High school)
 \item Medical Questions and Answers
 \item Human-centric standardized exams
 \item Grammatical acceptability
 \item Data analysis
\end{itemize}

  \clearpage

\section{Detailed benchmark review and classification} \label{appendix_benchmark_review}
  Classification of \numbenchmarks frequently used LLM benchmarks in recent literature, with a high-level description of the tasks covered in each benchmark, using a granular classification schema specified in Appendix \ref{appendix_existing_task_classification}.

\begin{itemize}
    \item Not using HELM \citep{liang2023holisticevaluationlanguagemodels} taxonomy as too high level for determining if datetimes already researched
    \item Only considering text-to-text tasks (i.e. not considering multi-modal input domains such as images, files)
    \item Not considering programming tasks
    \item Only considering English tasks (C-Eval \citep{huang2023ceval} and CMMLU \citep{li2024cmmlumeasuringmassivemultitask} were not examined)
\end{itemize}

\begin{longtable}{| p{.30\textwidth} | p{.70\textwidth} |} 
\toprule
Benchmark & Task(s) \\
\midrule

GLUE \citep{wang2019glue} 
& 
\begin{itemize} 
    \item Linguistic acceptability (CoLA)
    \item Sentiment analysis (SST-2)
    \item Sentence similarity (MRPC, STS-B)
    \item Question answering (QQP, QNLI)
    \item Textual entailment (MNLI, RTE)
    \item Reading comprehension (WNLI)
\end{itemize}
\\

SuperGLUE \citep{wang2020superglue} 
& 
\begin{itemize} 
    \item Question answering (BoolQ \citep{clark2019boolqexploringsurprisingdifficulty})
    \item Speaker commitment (CB)
    \item Question answering (COPA, MultiRC, ReCoRD)
    \item Textual entailment (RTE)
    \item Word sense disambiguation (WiC)
    \item Reading comprehension (WSC)    
\end{itemize}
\\

XTREME \citep{hu2020xtreme} 
& 
\begin{itemize} 
 \item Textual entailment (XNLI)
 \item Sentence similarity (PAWS-X, BUCC, Tatoeba)
 \item Part-of-speech tagging (POS)
 \item Named-entity Recognition (NER)
 \item Question answering (XQuAD, MLQA, TyDiQA-GoldP)
\end{itemize}
\\

Open LLM Leaderboard v1 \footnote{\url{https://huggingface.co/docs/leaderboards/open_llm_leaderboard/archive}}
& 
\begin{itemize} 
 \item AI2 Reasoning Challenge (ARC) \citep{clark2018thinksolvedquestionanswering}
 \item HellaSwag \citep{zellers2019hellaswagmachinereallyfinish}
 \item MMLU \citep{hendrycks2021measuringmassivemultitasklanguage}
 \item TruthfulQA \citep{lin2022truthfulqameasuringmodelsmimic}
 \item Winogrande \citep{sakaguchi2019winograndeadversarialwinogradschema}
 
\end{itemize}
\\

Open LLM Leaderboard v2 \footnote{\url{https://huggingface.co/docs/leaderboards/open_llm_leaderboard/about}}
& 
\begin{itemize} 
 \item IFEval \citep{zhou2023instructionfollowingevaluationlargelanguage}
 \item BBH (Big Bench Hard) \citep{suzgun2022challenging}
 \item MATH \citep{hendrycks2021measuringmathematicalproblemsolving}
 \item GPQA \citep{rein2023gpqagraduatelevelgoogleproofqa}
 \item MuSR (Multistep Soft Reasoning) \citep{sprague2024musrtestinglimitschainofthought}
 \item MMLU-PRO \citep{wang2024mmluprorobustchallengingmultitask}
\end{itemize}
\\

OpenCompass \citep{2023opencompass}
& 
\begin{itemize} 
 \item This is a framework of 45 benchmarks
 \item MATH \citep{hendrycks2021measuringmathematicalproblemsolving}
 \item GSM8k \citep{cobbe2021trainingverifierssolvemath}
 \item AGIEval \citep{zhong2023agievalhumancentricbenchmarkevaluating}
 \item MMLU \citep{hendrycks2021measuringmassivemultitasklanguage}
 \item AI2 Reasoning Challenge (ARC) \citep{clark2018thinksolvedquestionanswering}
 \item C-Eval \citep{huang2023ceval}
 \item Xiezhi \citep{gu2024xiezhieverupdatingbenchmarkholistic}
 \item And many more

\end{itemize}
\\

LiveBench \citep{white2024livebenchchallengingcontaminationfreellm}
& 
\begin{itemize} 
 \item Consists of 18 tasks across 6 categories
 \item Maths (High-school)
 \item Programming / Coding
 \item Reasoning and algorithmic
 \item Verifiable instructions
 \item Data analysis

\end{itemize}
\\

ARC-AGI \citep{chollet2019measureintelligence}
& 
\begin{itemize} 
 \item Pattern / Sequence prediction
\end{itemize}
\\

TyDi QA \citep{clark2020tydi} 
& 
\begin{itemize} 
 \item Question answering (Passage Selection Task, Minimal Answer Span Task)
\end{itemize}
\\

CoNLL-2002 \citep{sang2002introduction} 
& 
\begin{itemize} 
 \item Named-entity Recognition (Shared Task)
\end{itemize}
\\

CoNLL-2003 \citep{adelani2021masakhaner} 
& 
\begin{itemize} 
 \item Named-entity Recognition (Shared Task)
\end{itemize}
\\

MMLU \citep{hendrycks2021measuringmassivemultitasklanguage}
& 
\begin{itemize} 
 \item Humanities
 \item Social Science, e.g microeconmics
 \item STEM, e.g College maths and physics
\end{itemize}
\\

BBQ \citep{parrish2022bbqhandbuiltbiasbenchmark}
& 
\begin{itemize} 
 \item Social biases, e.g age, disability, gender
\end{itemize}
\\

BIG-Bench  \citep{srivastava2023imitationgamequantifyingextrapolating}
& 
\begin{itemize} 
    \item Many tasks
    \end{itemize}
\\

BIG-Bench Hard  \citep{suzgun2022challenging}
& 
\begin{itemize} 
    \item 204 tasks
    \end{itemize}
\\

HumanEval  \citep{chen2021evaluatinglargelanguagemodels}
& 
\begin{itemize} 
    \item Programming
    \end{itemize}
\\

NATURAL INSTRUCTIONS  \citep{mishra2022crosstaskgeneralizationnaturallanguage}
& 
\begin{itemize} 
    \item Question generation
    \item Classification
    \end{itemize}
\\

HellaSwag  \citep{zellers2019hellaswagmachinereallyfinish}
& 
\begin{itemize} 
    \item Commonsense NLI (what likely happens next)
    \end{itemize}
\\

GSM8K  \citep{cobbe2021trainingverifierssolvemath}
& 
\begin{itemize} 
    \item Maths (High-school)
    \end{itemize}
\\

ANLI  \citep{nie2020adversarialnlinewbenchmark}
& 
\begin{itemize} 
    \item Commonsense NLI
    \end{itemize}
\\

MATH, MATH-hard  \citep{hendrycks2021measuringmathematicalproblemsolving}
& 
\begin{itemize} 
    \item Maths (High-school)
    \end{itemize}
\\

MedQA  \citep{jin2020diseasedoespatienthave}
& 
\begin{itemize} 
    \item Medical Questions and Answers
    \end{itemize}
\\

AGIEval  \citep{zhong2023agievalhumancentricbenchmarkevaluating}
& 
\begin{itemize} 
    \item Human-centric standardized exams
    \item Contains Gaokao (China College Entrance Exam)
    \end{itemize}
\\

TriviaQA  \citep{joshi2017triviaqalargescaledistantly}
& 
\begin{itemize} 
    \item Reading comprehension (NLU)
    \end{itemize}
\\

ARC-C, ARC-E  \citep{clark2018thinksolvedquestionanswering}
& 
\begin{itemize} 
    \item World Knowledge
    \item Question answering
    \item Reasoning and algorithmic
    \end{itemize}
\\

PIQA  \citep{bisk2019piqareasoningphysicalcommonsense}
& 
\begin{itemize} 
    \item Physical commonsense
    \end{itemize}
\\

SocialIQA  \citep{bisk2019piqareasoningphysicalcommonsense}
& 
\begin{itemize} 
    \item Social situation commonsense
    \end{itemize}
\\

GSM-8K  \citep{clark2018thinksolvedquestionanswering}
& 
\begin{itemize} 
    \item Maths (High-school)
    \end{itemize}
\\

WinoGrande  \citep{sakaguchi2019winograndeadversarialwinogradschema}
& 
\begin{itemize} 
    \item Commonsense (pronoun resolution)
    \end{itemize}
\\

OpenBookQA  \citep{mihaylov2018suitarmorconductelectricity}
& 
\begin{itemize} 
    \item Commonsense (science facts)
    \end{itemize}
\\

BoolQ  \citep{clark2019boolqexploringsurprisingdifficulty}
& 
\begin{itemize} 
    \item Commonsense (reading comprehension with yes/no questions)
    \end{itemize}
\\

CommonSenseQA  \citep{talmor2019commonsenseqaquestionansweringchallenge}
& 
\begin{itemize} 
    \item Commonsense (question answering with prior knowledge)
    \end{itemize}
\\

TruthfulQA  \citep{lin2022truthfulqameasuringmodelsmimic}
& 
\begin{itemize} 
    \item Commonsense
    \item World Knowledge
    \end{itemize}
\\

MBPP  \citep{austin2021programsynthesislargelanguage}
& 
\begin{itemize} 
    \item Programming (Python)
    \end{itemize}
\\

GPQA  \citep{rein2023gpqagraduatelevelgoogleproofqa}
& 
\begin{itemize} 
    \item Expert Knowledge (chemistry, biology, genetics)
    \end{itemize}
\\

MT Bench  \citep{zheng2023judgingllmasajudgemtbenchchatbot}
& 
\begin{itemize} 
    \item Verifiable instructions, instruction following
    \item Chat conversation
    \item Expert Knowledge (chemistry, biology, genetics)
    \end{itemize}
\\

IFEval  \citep{zhou2023instructionfollowingevaluationlargelanguage}
& 
\begin{itemize} 
    \item Verifiable instructions
    \end{itemize}
\\

Musr  \citep{sprague2024musrtestinglimitschainofthought}
& 
\begin{itemize} 
    \item Commonsense reasoning (Multistep Soft Reasoning)
    \end{itemize}
\\

SVAMP  \citep{patel-etal-2021-nlp}
& 
\begin{itemize} 
    \item Maths (English math word problems)
    \end{itemize}
\\

ASDiv  \citep{miao2021diversecorpusevaluatingdeveloping}
& 
\begin{itemize} 
    \item Maths (English math word problems)
    \end{itemize}
\\

Natural Questions (NQ)  \citep{kwiatkowski-etal-2019-natural}
& 
\begin{itemize} 
    \item Question answering
    \end{itemize}
\\

MathVista  \citep{lu2024mathvistaevaluatingmathematicalreasoning}
& 
\begin{itemize} 
    \item Maths
    \item Multimodal (IQTest, FunctionQA, and PaperQA)
    \end{itemize}
\\

Geometry3K  \citep{lu2021intergpsinterpretablegeometryproblem}
& 
\begin{itemize} 
    \item Maths (Multimodal)
    \end{itemize}
\\

MGSM, MGSM8KInstruct  \citep{chen2024breakinglanguagebarriersmultilingual}
& 
\begin{itemize} 
    \item Maths (multilingual math reasoning instruction)
    \end{itemize}
\\

SQuAD  \citep{rajpurkar2016squad100000questionsmachine}
& 
\begin{itemize} 
    \item Reading comprehension (NLU)
    \item World Knowledge
    \end{itemize}
\\

SQuADv2  \citep{rajpurkar2018knowdontknowunanswerable}
& 
\begin{itemize} 
    \item Reading comprehension (NLU)
    \item World Knowledge
    \item Reasoning (question is unanswerable)
    \end{itemize}
\\

Penn Tree Bank (PTB)  \citep{marcus-etal-1994-penn}
& 
\begin{itemize} 
    \item Part-of-speech tagging (POS)
    \end{itemize}
\\

CoQA  \citep{reddy2019coqaconversationalquestionanswering}
& 
\begin{itemize} 
    \item Question answering
    \item Commonsense reasoning
    \end{itemize}
\\

LAMBADA  \citep{paperno2016lambadadatasetwordprediction}
& 
\begin{itemize} 
    \item Completion (Next word prediction)
    \end{itemize}
\\

StoryCloze  \citep{mostafazadeh2016corpusevaluationframeworkdeeper}
& 
\begin{itemize} 
    \item Reading comprehension (NLU)
    \item Completion (story ending)
    \end{itemize}
\\

Winograd  \citep{Levesque2012TheWS}
& 
\begin{itemize} 
    \item Reading comprehension (NLU)
    \item Sentence disimilarity
    \end{itemize}
\\

QuAC  \citep{choi2018quacquestionanswering}
& 
\begin{itemize} 
    \item Question answering
    \item World Knowledge
    \end{itemize}
\\

RACE  \citep{lai2017racelargescalereadingcomprehension}
& 
\begin{itemize} 
    \item Question answering (multiple choice)
    \item Reading comprehension
    \end{itemize}
\\

DROP  \citep{dua2019dropreadingcomprehensionbenchmark}
& 
\begin{itemize} 
    \item Question answering (multiple choice)
    \item Reading comprehension
    \end{itemize}
\\

COPA  \citep{dua2019dropreadingcomprehensionbenchmark}
& 
\begin{itemize} 
    \item Question answering (multiple choice)
    \item Reading comprehension
    \end{itemize}
\\

CoLA  \citep{warstadt2019neuralnetworkacceptabilityjudgments}
& 
\begin{itemize} 
    \item Grammatical acceptability
    \end{itemize}
\\

RTE  \citep{wang2019glue}
& 
\begin{itemize} 
    \item Textual entailment
    \end{itemize}
\\

QNLI  \citep{wang2019glue}
& 
\begin{itemize} 
    \item Reading comprehension (NLU)
    \end{itemize}
\\

MNLI  \citep{williams2018broadcoveragechallengecorpussentence}
& 
\begin{itemize} 
    \item Textual entailment or contradiction
    \end{itemize}
\\

STS-B  \citep{Cer_2017}
& 
\begin{itemize} 
    \item Sentence (di-)similarity
    \end{itemize}
\\

SST-2  \citep{socher-etal-2013-recursive}
& 
\begin{itemize} 
    \item Sentiment analysis
    \end{itemize}
\\

MRPC  \citep{dolan-brockett-2005-automatically}
& 
\begin{itemize} 
    \item Sentence similarity
    \end{itemize}
\\

QQP \citep{qqp}
& 
\begin{itemize} 
    \item Sentence similarity
    \end{itemize}
\\

MultiRC \citep{qqp}
\begin{itemize} 
    \item Reading comprehension (NLU)
    \end{itemize}
\\

NLEP \citep{zhang2024naturallanguageembeddedprograms} 
& 
\begin{itemize}
 \item Date understanding and reasoning 
\end{itemize}
\\

OlympiadBench \citep{he2024olympiadbenchchallengingbenchmarkpromoting} 
& 
\begin{itemize}
 \item Maths, Physics
\end{itemize}
\\

OrcaMath \citep{mitra2024orcamathunlockingpotentialslms} 
& 
\begin{itemize}
 \item Maths
\end{itemize}
\\

Minerva \citep{lewkowycz2022solvingquantitativereasoningproblems} 
& 
\begin{itemize}
 \item Maths
\end{itemize}
\\

\bottomrule
\end{longtable}

  \clearpage

\section{TaskFigures} 
  \label{task_figures}
  \TODO{this appendix may not be necessry}
\TODO{check task figures represent new tasks}

\begin{figure}[H]
    \includegraphics[trim={0 11cm 0 0}, clip, width=1.5\textwidth]{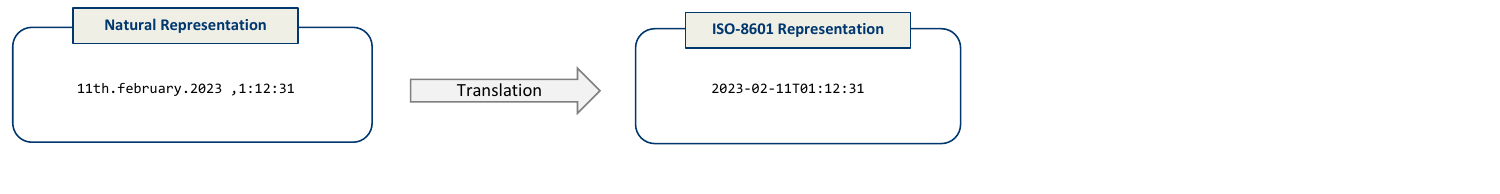}
    \caption{The \emph{Translation ISO-8601} task requires the translation of a datetime from it's natural representation to it's ISO-8601 representation.}
    \label{fig:natural-and-iso8601-example}
\end{figure}

\begin{figure}[H]
    \includegraphics[trim={0 11cm 0 0}, clip, width=1.5\textwidth]{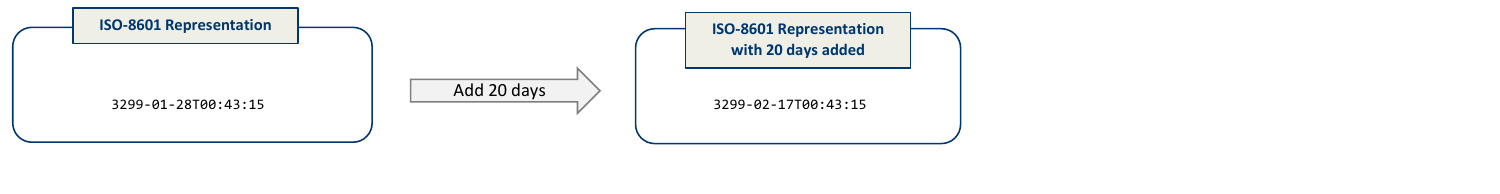}
    \caption{The \emph{Computation Add-20} task requires adding 20 days to a datetime provided in ISO-8601 representation, and producing a new ISO-8601 datetime with the result.}
    \label{fig:add-20-simple-example}
\end{figure}

Figures \ref{fig:natural-and-iso8601-example} and \ref{fig:add-20-simple-example} illustrate the \emph{Translation} and \emph{Computation} tasks. The \emph{Mixed} tasks are essentialy a combination of translation and computation, requiring both capabilities jointly, and are covered in Appendix \label{datetime_mixed}.

  \clearpage

\section{Glossary}
  
\section{Scenarios}
\begin{itemize}
    \item Scenario \citep{liang2023holisticevaluationlanguagemodels}. A 5-tuple comprised of a (task, what, who, when and language)
\end{itemize}

\section{User facing tasks}
Tasks using user-facing tasks taxonomy from HELM \citep{liang2023holisticevaluationlanguagemodels}. 
\begin{itemize}
    \item question answering
    \item information retrieval
    \item summarization
    \item sentiment analysis
    \item toxicity detection
\end{itemize}

NLEP: Natural language embedded programs

  \clearpage

\end{document}